\documentclass{article}

\usepackage[nonatbib, preprint]{neurips_2024}

\usepackage[sorting=nyt,style=authoryear,bibencoding=utf8,backend=biber,natbib=true,uniquename=false,uniquelist=false,maxbibnames=99,maxcitenames=1,mincitenames=1]{biblatex}

\usepackage[utf8]{inputenc} 
\usepackage[T1]{fontenc}    
\usepackage{hyperref}       
\usepackage{url}            
\usepackage{booktabs}       
\usepackage{amsfonts}       
\usepackage{nicefrac}       
\usepackage{microtype}      
\usepackage{xcolor}         
\usepackage{multirow}
\usepackage{amsmath}
\usepackage{graphicx}
\usepackage{array}
\usepackage{listings}
\usepackage{caption}
\usepackage{algorithm2e}
\usepackage{subcaption}

\usepackage{tikz}
\usetikzlibrary{positioning, shapes.geometric, arrows.meta, bending, decorations.markings}

\definecolor{lightblue}{rgb}{0.68, 0.85, 0.9}  

\hypersetup{
    colorlinks,
    linkcolor={red!50!black},
    citecolor={blue!50!black},
    urlcolor={blue!80!black},
    pdftitle={Sparse Autoencoders Enable Scalable and Reliable Circuit Identification in Language Models},
}

\title{Sparse Autoencoders Enable Scalable and Reliable Circuit Identification in Language Models}

\author{%
  Charles O'Neill \\
  School of Computing\\
  The Australian National University\\
  Canberra, ACT, 2601 \\
  \texttt{charles.oneill@anu.edu.au} \\
  \And 
  Thang Bui
  \\
School of Computing\\
  The Australian National University\\
  Canberra, ACT, 2601 \\
  \texttt{thang.bui@anu.edu.au} \\
}

\begin{document}

\maketitle

\begin{abstract}
This paper introduces an efficient and robust method for discovering interpretable circuits in large language models using discrete sparse autoencoders. Our approach addresses key limitations of existing techniques, namely computational complexity and sensitivity to hyperparameters. We propose training sparse autoencoders on carefully designed positive and negative examples, where the model can only correctly predict the next token for the positive examples. We hypothesise that learned representations of attention head outputs will signal when a head is engaged in specific computations. By discretising the learned representations into integer codes and measuring the overlap between codes unique to positive examples for each head, we enable direct identification of attention heads involved in circuits without the need for expensive ablations or architectural modifications. On three well-studied tasks - indirect object identification, greater-than comparisons, and docstring completion - the proposed method achieves higher precision and recall in recovering ground-truth circuits compared to state-of-the-art baselines, while reducing runtime from hours to seconds. Notably, we require only 5-10 text examples for each task to learn robust representations. Our findings highlight the promise of discrete sparse autoencoders for scalable and efficient mechanistic interpretability, offering a new direction for analysing the inner workings of large language models.
\end{abstract}

\section{Introduction}

The rapid advancement of large language models (LLMs) based on transformers \citep{vaswani2017attention} has spurred interest in mechanistic interpretability \citep{olah2018building}, which aims to break down model components into human-understandable \textit{circuits}. Circuits are defined as subgraphs of a model's computation graph that implement a task-specific behaviour \citep{olah2020zoom}. While progress has been made in automating the isolation of certain circuits \citep{conmy2024towards, cao2021low, michel2019sixteen}, automatic circuit discovery remains too brittle and complex to replace manual inspection. As model sizes grow \citep{kaplan2020scaling, lieberum2023does}, manual inspection becomes increasingly impractical, hence the need for more robust and efficient methods. 

Automated circuit identification algorithms suffer from several drawbacks, such as sensitivity to the choice of metric, the type of intervention used to identify important components, and computational intensity \citep{conmy2024towards, olsson2022context, wang2022interpretability, geiger2021causal}. Whilst faster variants of automated algorithms have been leveraged with some success \citep{syed2023attribution, hanna2024have}, they retain many of the same failure modes and performance is heavily dependent on metric choice. All variants of automated circuit-identification have been shown to perform poorly at recovering ground-truth circuits in specific situations \citep{conmy2024towards}, meaning researchers cannot know \textit{a priori} whether the circuit found is accurate or not. Without simpler and more robust algorithms, researchers will be limited to using painstakingly slow manual circuit identification \citep{rauker2023toward}.

Our main contribution is the introduction of a highly performant yet remarkably simple circuit-identification method based on the presence of features in sparse autoencoders (SAEs) trained on transformer attention head outputs. SAEs have been shown to learn interpretable compressed features of the model's internal states \citep{cunningham2023sparse, sharkey2022taking, bricken2023towards}. We hypothesise that these representations of attention head outputs should contain signal about when a head is engaged in a particular type of computation as part of a circuit. The key insight behind our approach is that by training SAEs on carefully designed examples of a task that requires the language model to use a specific circuit (and examples where it doesn't), the learned representations should capture circuit-specific behaviour. 

We demonstrate that by simply looking for the codes unique to positive examples within \textit{as few as 5-10 text examples of a task}, we can directly identify the attention heads in the ground-truth circuit with better or equal precision and recall than existing methods, while being significantly faster and less complex. Specifically, our method allows us to do away with choosing a metric to measure the importance of a model component, which we see as a fundamental advantage of our method over previous approaches. We evaluate the proposed method on three well-studied circuits and demonstrate its robustness to hyperparameter choice. Our findings highlight the potential of using discrete sparse autoencoders for efficient and effective circuit identification in large language models.


\begin{figure}
    \centering
    \includegraphics[width=\textwidth, trim=180pt 0pt 10pt 0pt, clip]{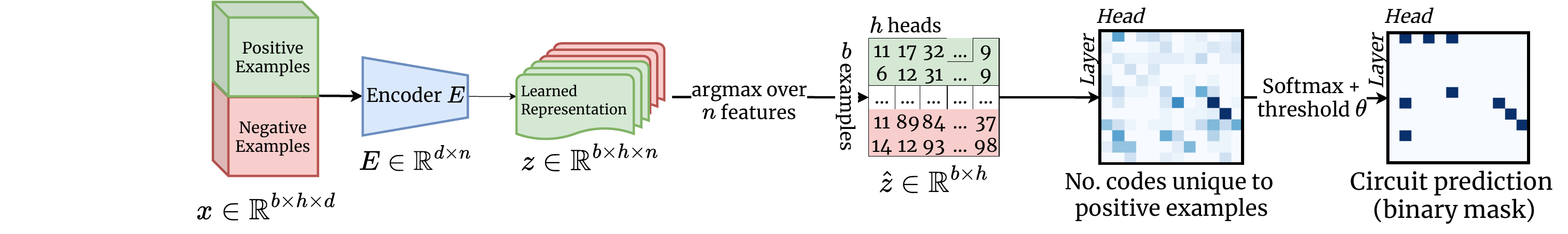}
    \caption{After training the sparse autoencoder, we obtain discrete representations $\mathbf{z}$ by passing tensor $\mathbf{x}$ and taking the argmax over the feature dimension, obtaining an integer code for each head in each example in $\hat{\mathbf{z}}$. $b$ is the number of examples, $h$ is the number of heads, $d$ is the transformer hidden dimension and $n$ is the number of learned features. For \textbf{node-level circuit identification}, shown here, we compute the number of codes unique to positive examples per head, normalise with softmax, choose a threshold $\theta$, and identify a head as being in the circuit if it surpasses the threshold. For \textbf{edge-level circuit identification}, shown in Figure \ref{fig:edge_method}, we count the number of co-occurrences of codes between heads for the top-$k$ co-occurrences, and then again take the softmax and thresholding with $\theta$.}
    \label{fig:method}
\end{figure}

\section{Background}

\subsection{Attention Heads and Circuits in Autoregressive Transformers}

Autoregressive, decoder-only transformers rely on self-attention to weigh the importance of different parts of the input sequence \citep{vaswani2017attention}, with a goal to predict the next token. In the multi-head attention mechanism, each attention head operates on a unique set of query, key, and value matrices, allowing the model to capture diverse relationships \textit{between} elements of the input. The output of the $i$-th attention head can be formally described as:
\begin{align*}
\mathbf{h}_i = \text{softmax}\left(\frac{(XW_Q^i)(XW_K^i)^T}{\sqrt{d_k}}\right)(XW_V^i)
\end{align*}
where $X$ is the input sequence, $W_Q^i, W_K^i, W_V^i$ are the query, key, and value matrices for the $i$-th head, and $d_k$ is the dimensionality of the key vectors. The outputs of the individual attention heads are then concatenated and linearly transformed to produce the overall output of the multi-head attention layer:
\begin{align*}
\text{MultiHead}_{\ell}(Q_{\ell}, K_{\ell}, V_{\ell}) &= \text{Concat}(\mathbf{h}_{\ell 1}, \ldots, \mathbf{h}_{\ell J}) W_O^{\ell}
\end{align*}
where $W_O^{\ell} \in \mathbb{R}^{Jd_v \times d}$ is the projection matrix. Each attention head output $\mathbf{h}_{\ell j}$ resides in the residual stream space $\mathbb{R}^d$, contributing independently to total attention at that layer \citep{elhage2021mathematical}.

The \textit{residual stream} refers to the sequence of token embeddings, with each layer's output being added back into this stream. Attention heads read from the residual stream by extracting information from specific tokens and write back their outputs, modifying the embeddings for subsequent layers. This additive form allows us to analyse each head's contribution to the model's behaviour by examining them independently. By tracing the flow of information across layers, we can identify computational circuits composed of multiple attention heads.

\subsection{Learning Sparse Representations of Attention Heads with Autoencoders}

Sparse autoencoders provide a promising approach to learn useful representations of attention head outputs that are amenable to circuit analysis. Given a set of attention head outputs $\{\mathbf{h}_i\}_{i=1}^{n_\text{heads}}$, where $\mathbf{h}_i \in \mathbb{R}^{d_\text{model}}$, we train an autoencoder with a single hidden layer and tied weights for the encoder $E$ and decoder $D$. The autoencoder learns a dictionary of basis vectors $\mathbf{v}_j \in \mathbb{R}^{d_\text{model}}$ such that each $\mathbf{h}_i$ can be approximated as a sparse linear combination of the dictionary elements: $\mathbf{h}_i \approx \sum_{j=1}^{d_\text{bottleneck}} z_{i,j} \mathbf{v}_j,
$ where $z_{i,j}$ are the sparse activations and $d_\text{bottleneck}$ is the dimensionality of the bottleneck layer. The autoencoder is trained to minimise a loss function that includes a reconstruction term and a sparsity penalty, controlled by the hyperparameter $\lambda$:
$$
\mathcal{L} = \sum_{i=1}^{n_\text{heads}} \left\lVert \mathbf{h}_i - \sum_{j=1}^{d_\text{bottleneck}} z_{i,j} \mathbf{v}_j \right\rVert_2^2 + \lambda \sum_{i=1}^{n_\text{heads}} \sum_{j=1}^{d_\text{bottleneck}} |z_{i,j}|.
$$
The dimensionality of the bottleneck layer $d_\text{bottleneck}$ can be either larger (projecting up) or smaller (projecting down) than the input dimensionality $d_\text{model}$. While projecting up allows for an overcomplete representation and can capture more nuanced features, projecting down can also be effective in learning a compressed representation that captures the most essential aspects of the attention head outputs \citep{olshausen1997sparse, lee2006efficient, wright2022high}. We propose a subtle but significant shift in perspective by treating sparse autoencoding as a compression problem rather than a problem of learning higher-dimensional sparse bases in the context of transformers. We hypothesise that compression is likely a key mechanism in identifying features which represent circuit-related computation, and in contrast which computation is shared between positive and negative examples.

To further simplify the representation and facilitate the identification of distinct behaviours within the attention heads, we discretise the sparse activations obtained from the autoencoder using an argmax operation over the feature dimension, $c_i = \text{argmax}_{j} z_{i,j}$, where $c_i \in {1, \ldots, d_\text{bottleneck}}$ is the discrete code assigned to the $i$-th attention head output. This yields a discrete bottleneck representation analogous to vector quantization \citep{van2017neural}. We will next discuss how to leverage the resulting discrete representations to identify important task-specific circuits in the transformer.

\section{Methodology}

Our approach centers on training a sparse autoencoder with carefully designed positive and negative examples, where the model only successfully predicts the next token for positive examples. The critical insight is that the compressed representation must capture the difference between the two sets of examples to achieve a low reconstruction loss. This differentiation enables us to isolate circuit-specific behaviours and identify the attention heads involved in the circuit of interest.

\subsection{Datasets}

We compile datasets of 250 ``positive'' and 250 ``negative'' examples for each task. Positive examples are text sequences where the model must use the circuit of interest to correctly predict the next token. In contrast, negative examples are semantically similar to positive examples but corrupted such that there is no correct ``next token.'' This dataset design ensures that the learned representations are common between positive and negative examples for attention heads processing semantic similarities but different for heads involved in circuit-specific computations. Table \ref{tab:examples} shows task examples, and Appendix \ref{app:circuit_visualisation} contains details of each dataset.

\begin{table}[htbp]
\centering
\footnotesize
\begin{tabular}{@{}p{1.7cm}p{4.6cm}p{4.8cm}p{1.7cm}@{}}
\toprule
Task       & Positive Example                                              & Negative Example  & Answer \\ \midrule
\textit{IOI}        & ``When \textcolor{lightblue}{\textbf{Elon}} and \textcolor{lightblue}{\textbf{Sam}} finished their meeting, \textcolor{lightblue}{\textbf{Elon}} gave the model to '' & ``When \textcolor{lightblue}{\textbf{Elon}} and \textcolor{lightblue}{\textbf{Sam}} finished their meeting, \textcolor{lightblue}{\textbf{Andrej}} gave the model to '' & ``Sam''    \\
\addlinespace 
\textit{Greater-than} & ``The AI war lasted from \textcolor{lightblue}{\textbf{2024}} to \textcolor{lightblue}{\textbf{20}}'' & ``The AI war lasted from \textcolor{lightblue}{\textbf{2024}} to \textcolor{lightblue}{\textbf{19}}'' & Any two digit number $>24$      \\
\vspace{2.5mm} \textit{Docstring}  & 
\begin{lstlisting}
def old(self, page, names, size):
    """sector gap"""
    
    :param page: message tree
    :param names: detail mine
    :param 
\end{lstlisting} &  
\begin{lstlisting}
def old(self, page, names, size):
    """sector gap"""
    
    :param image: message tree
    :param update: detail mine
    :param 
\end{lstlisting} 
    &   \vspace{2.5mm} \texttt{size}   \\ \bottomrule
\end{tabular}
\caption{Task-specific positive and negative examples. Positive examples are designed to elicit the behaviour being studied when the model conducts next token prediction on the example. Negative examples are designed to be semantically similar to the positive examples but with minor corruptions that mean there is now no obviously correct answer.}
\label{tab:examples}
\end{table}

The \textit{Indirect Object Identification} (IOI) task involves sentences such as ``When Elon and Sam finished their meeting, Elon gave the model to'' with the aim being to predict ``Sam'', the indirect object \citep{wang2022interpretability}. Negative examples introduce a third name, eliminating any bias towards completing either of the two original names.

The \textit{Greater-than} task involves sentences of the form ``The \texttt{<noun>} lasted from \texttt{XXYY} to \text{XX}'', where the aim is give all non-zero probability to years > \texttt{YY} \citep{hanna2024does}. Negative examples consist of impossible completions, with the ending year preceding the starting century.

The \textit{Docstring} task assesses the model's ability to predict argument names in Python docstrings based on the function's argument list \citep{heimersheim2023circuit}. Docstrings follow a format with \texttt{:param} followed by an argument name. The model predicts the next argument name after the \texttt{:param} tag. Negative examples employ random argument names.

\subsection{Model and circuit identification with learned features}

Our methodology consists of two stages: training the sparse autoencoder to conduct dictionary learning on the cached model activations, and using these learned representations to identify model components involved in the circuit (Figure \ref{fig:method}).

\paragraph{Training the autoencoder to get learned features}
We first take all positive and negative input prompts for a dataset and tokenize them. Since each prompt is curated to have the same number of tokens for all positive and negative examples across all datasets, we concatenate the prompts into a single tensor $\mathbf{x} = \left[ \{x_i\}_{i=1}^n ; \{x_i^\prime\}_{i=1}^n \right] \in \mathbb{R}^{n_\text{examples} \times \text{sequence length}}$. \textit{We found that using only 10 examples (with an equal number of positive and negative examples) led to the most robust representations learned by the SAE} (see Figure \ref{fig:roc_auc_vs_examples}). The remaining examples are used as an evaluation set.

\paragraph{Node-Level and Edge-Level Circuit Identification}

\textit{Node-level} circuit discovery predicts model components (i.e., attention heads) as part of the circuit based on individual outputs in isolation. In contrast, \textit{edge-level} circuit discovery predicts whether the information flow (i.e., the edge) is important by considering how certain components act together, specifically the frequency of co-activation of specific codes in different heads. For a full discussion of the details and semantics of node-level and edge-level discovery, see Appendix \ref{app:acdc}.

After training the SAE, we perform a forward pass of all examples $\mathbf{x}$ through the encoder $E$ to obtain the learned activations $\mathbf{z} \in \mathbb{R}^{n_\text{examples}\times n_\text{heads}\times d_\text{bottleneck}}$. We then apply an argmax operation across the feature (bottleneck) dimension of $\mathbf{z}$, which yields a matrix of discrete codes $\mathbf{z}_\text{discrete} = \text{argmax}_{d}(\mathbf{z})$, where each code represents the most activated feature for a particular attention head.

\textbf{Node-level identification:} Let $\mathbf{p} \in \mathbb{R}^{n_\text{heads} \times d_\text{bottleneck}}$ be a matrix of one-hot vectors indicating which codes are activated for each head in the positive examples, and let $\mathbf{n} \in \mathbb{R}^{n_\text{heads} \times d_\text{bottleneck}}$ be the corresponding matrix for the negative examples.

We next compute a vector $\mathbf{u} \in \mathbb{R}^{n_\text{heads}}$, where each element $\mathbf{u}_i$ represents the number of unique codes that appear only in the positive examples, optionally normalised by the total number of codes across all examples, for the $i$-th attention head: $\mathbf{u}_i = |\mathbf{p}_i \setminus \mathbf{n}_i| / |\mathbf{p}_i \cup \mathbf{n}_i|$.  Intuitively, a high value of $\mathbf{u}_i$ indicates that the $i$-th head activates a large proportion of codes that are unique to positive examples. We then apply a softmax function to $\mathbf{u}$ and select a threshold $\theta$ to determine if a head is part of the ground-truth circuit (Figure \ref{fig:method}). Whilst we vary $\theta$ to construct analyses such as ROC curves, in practice a single $\theta$ should be selected to predict a circuit.

\textbf{Edge-level identification:} Let us construct co-occurrence matrices $\mathbf{C}^+$ and $\mathbf{C}^-$ for the positive and negative examples, respectively. Each entry $\mathbf{C}_{h1,h2,c1,c2}$ represents the frequency of co-occurrence between codes $c1$ and $c2$ in heads $h1$ and $h2$:	
\begin{align*}
\mathbf{C}_{h1,h2,c1,c2}^+ &= \Big|(c1, c2) \text{ co-occurrences in positive examples between } h1 \text{ and } h2\Big| \\
\mathbf{C}_{h1,h2,c1,c2}^- &= \Big|(c1, c2) \text{ co-occurrences in negative examples between } h1 \text{ and } h2\Big|
\end{align*}
We then compute a matrix $\mathbf{U}$, where each entry $\mathbf{U}_{h1,h2}$ represents the number of co-occurrences that appear in the positive examples but not in the negative examples for the head pair $(h1, h2)$: $\mathbf{U}_{h1,h2} = \sum_{i,j} \mathbf{C}^+_{h1,h2,i,j}$ where $\mathbf{C}^+_{h1,h2,:,:} > 0 \land \mathbf{C}^-_{h1,h2,:,:} = 0$. Once the head pairs $(h1, h2)$ are sorted in descending order of their corresponding values in $\mathbf{U}$, we introduce a hyperparameter $k$ to determine the number of top-ranked head pairs to include in the predicted circuit. We set $k$ to be half the total number of head pairs for all analyses, and show that this is a robust choice in Appendix \ref{app:k_effect}.

The next step is to initialise $\mathbf{u} \in \mathbb{R}^{n_{\text{heads}}}$ as a zero vector. For each of the top $k$ head pairs $(h1, h2)$, the corresponding entries in $\mathbf{u}$ are incremented: $\mathbf{u}[h1] += 1$ and $\mathbf{u}[h2] += 1$. We then apply softmax across $\mathbf{u}$ and choose a threshold $\theta$ to predict whether a particular head is part of the circuit (Figure \ref{fig:edge_method}).

\section{Results}
\label{sec:results}

We tokenise positive and negative input prompts with the GPT-2 tokeniser \citep{radford2019language}, pass them through GPT-2 Small, and cache the outputs of each attention head. We concatenate all prompts into a single tensor $\mathbf{x} \in \mathbb{R}^{n_\text{examples}\times n_\text{heads} \times d_\text{model}}$, aggregating across the position dimension by taking the mean. We train the SAE on 10 examples from this tensor and use the rest for validation. The SAE is trained until convergence on the evaluation set, using a combination of reconstruction and sparsity losses, optimised with the Adam algorithm. For the main results, we set the number of learned features to be 200 and $\lambda$ to be 0.02 across all datasets. For edge-level identification, we choose $k$ to be half of the total number of co-occurrences. 

We compare our method to the three state-of-the-art approaches to circuit discovery in language models: automatic circuit discovery (ACDC) \citep{conmy2024towards}, head importance score pruning (HISP) \citep{michel2019sixteen}, and subnetwork probing (SP) \citep{cao2021low}. Appendix \ref{app:acdc_hisp_sp} contains details of these algorithms. Additionally, we provide an unsupervised evaluation comparison of our method with edge attribution patching \citep{syed2023attribution}, which uses linear approximations to the patches performed in ACDC above (see \ref{sec:related_work}). This makes EAP comparable to our method in terms of speed and efficiency; see Appendix \ref{app:eap}.

The whole process, from training the SAE to sweeping over all thresholds $\theta$ in a circuit, typically takes less than 15 seconds on GPT2-Small, and less than a minute on GPT2-XL. See Figure \ref{fig:wall_time} for a detailed indication of wall-times for our method at various scales. 


\begin{figure}
    \centering
    \includegraphics[width=0.9\textwidth]{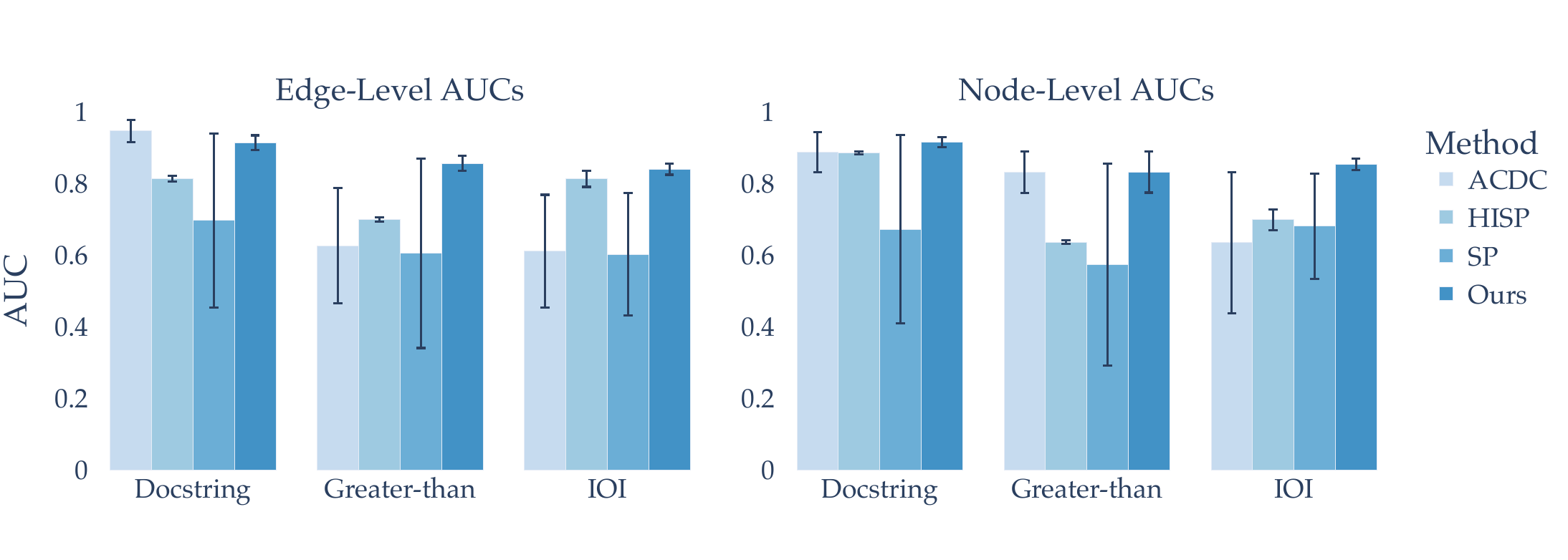}
    \caption{Comparison of our method's performance against state-of-the-art circuit identification techniques (ACDC, HISP, and SP) on three well-studied transformer circuits: Docstring, Greater-than, and Indirect Object Identification (IOI). The bar plots show the average AUC (Area Under the ROC Curve) scores for each method, averaged across KL divergence and loss metrics, for both edge-level and node-level circuit identification. Error bars for \textit{Ours} represent the standard deviation of AUC scores across 5 runs. Our method consistently outperforms or matches the performance of existing techniques across all tasks.}
    \label{fig:edge_node_auc}
\end{figure}

\subsection{Discrete SAE features excel in node-level and edge-level circuit discovery}

Our circuit-identification method outperforms ACDC, HISP, and SP in terms of ROC AUC across all datasets, regardless of ablation type used for these methods (Figure \ref{fig:edge_node_auc}), with the exception of ACDC, which achieves a higher ROC AUC on edge-level circuit identification on the docstring task. We find a strong correlation between the number of unique positive codes per head and the presence of that head in the ground-truth circuit (Figure \ref{fig:comparison_unique_to_positive_vs_ground_truth}, Appendix \ref{app:circuit_visualisation}). ROC curves are constructed by sweeping over thresholds (Figure \ref{fig:combined_roc_curves}): for our method, we sweep over the threshold $\theta$ required for a softmaxed head's number of unique positive codes to be included in the circuit, while for ACDC, we sweep over the threshold $\tau$, which determines the difference in the chosen metric between an ablated and clean model required to remove a node from the circuit. Notably, we identify negative name-mover heads in the IOI circuit \citep{wang2022interpretability} (heads that calibrate probability by downweighting the logit of the IO name), which other algorithms struggled to do \citep{conmy2024towards} (see Appendix \ref{app:ioi}). 


\subsection{Performance is robust to hyperparameter choice}

\begin{figure}[ht]
    \centering
    \begin{minipage}{0.48\textwidth}
        \centering
        \includegraphics[width=\textwidth]{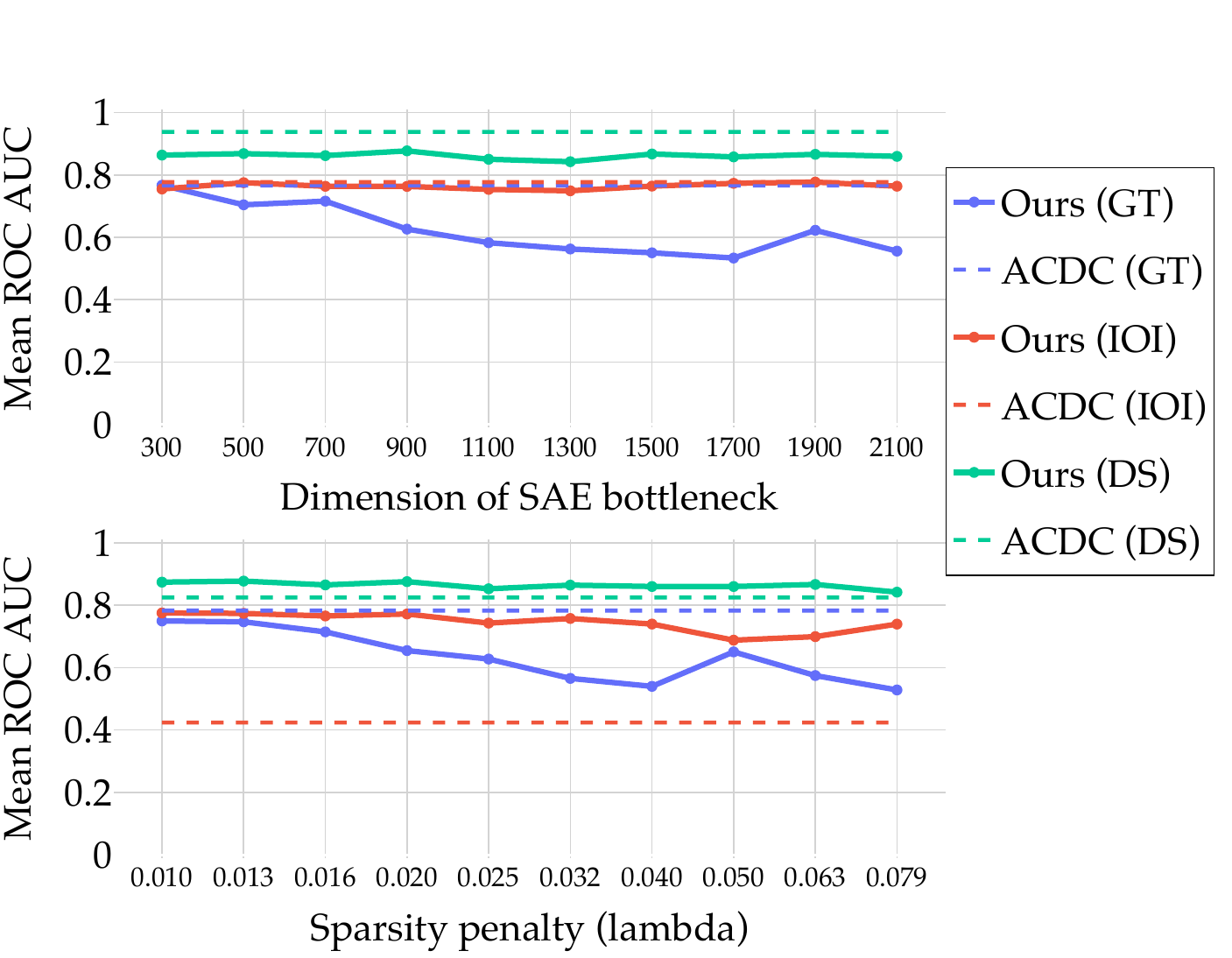}
        \caption{Mean ROC AUC scores across different values of the number of SAE features and sparsity penalty $\lambda$.}
        \label{fig:roc_auc_vs_params}
    \end{minipage}\hfill
    \begin{minipage}{0.48\textwidth}
        \centering
        \includegraphics[width=0.95\textwidth, trim=0 0 0 50pt, clip]{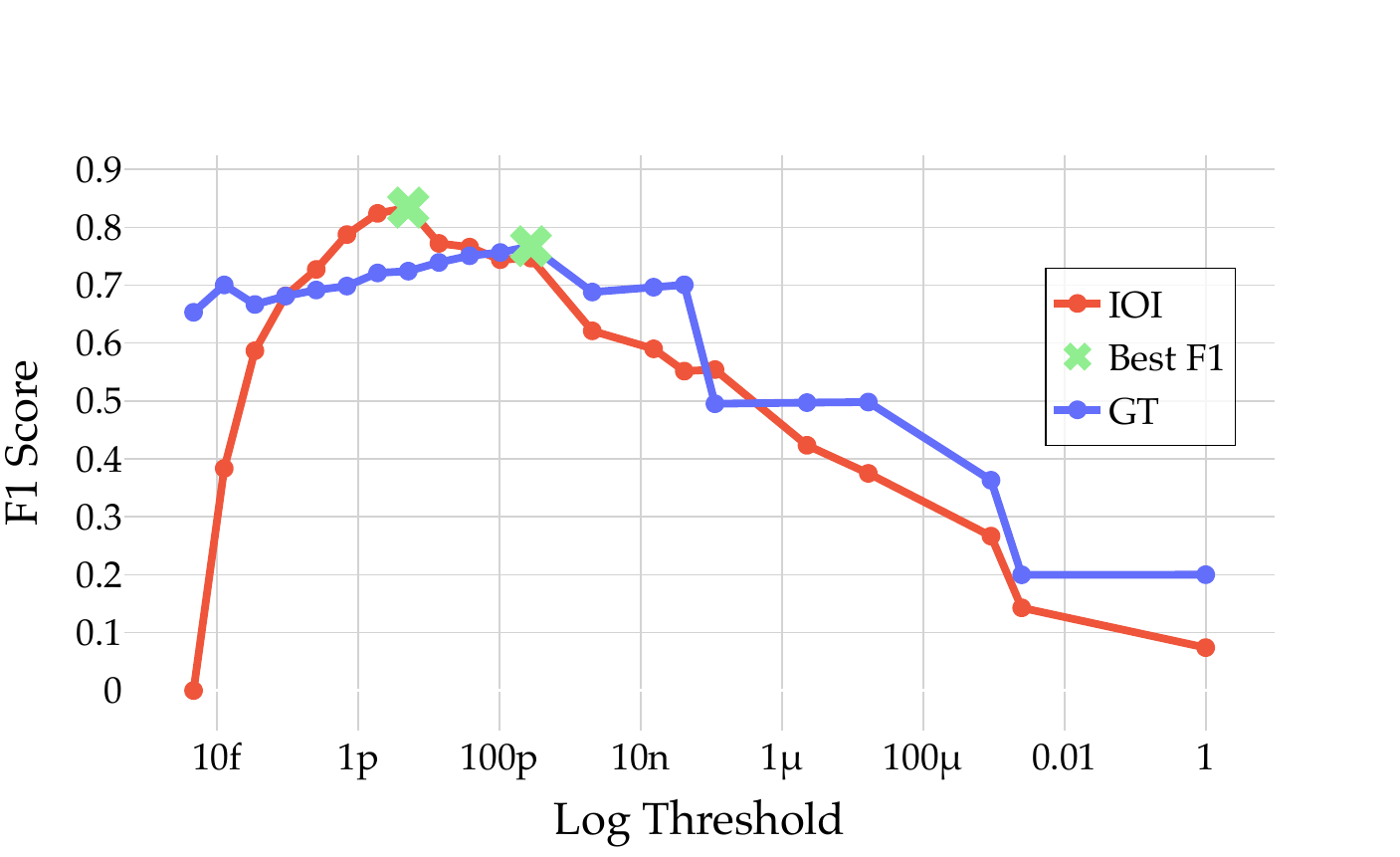}
        \caption{F1 score (node-level) for each dataset given a threshold $\theta$ for selecting a head's importance (after softmax). The optimal threshold is approximately the same for both IOI and Greater-than tasks.}
        \label{fig:f1_vs_threshold}
    \end{minipage}
\end{figure}

To demonstrate the robustness of our method to its hyperparameters, we consider two distinct groups: (1) those controlling the capacity and expressiveness of the sparse autoencoder (SAE), namely the size of the bottleneck and the sparsity penalty $\lambda$, and (2) the threshold $\theta$ for selecting a head after softmax. We trained 100 autoencoders with varying numbers of features in the hidden layer and different values of $\lambda$. We observed no significant drop-off in ROC AUC for IOI and Docstring tasks, and a slight drop-off for Greater-than, as we increase both hyperparameters (Figure \ref{fig:roc_auc_vs_params}). Finally, we examine the robustness of the value of $\theta$ on the pointwise F1 score (node-level) for both the IOI and GT datasets (Figure \ref{fig:f1_vs_threshold}). The optimal threshold is approximately the same for both tasks, suggesting we may be able to set this threshold for any arbitrary circuit. For edge-level discovery, we also find that performance is robust to $k$ (Appendix \ref{app:k_effect}).

We note that a more abstract hyperparameter is the construction of negative examples. We present an examination of this in Figure \ref{fig:easy_negative_types} in Appendix \ref{app:alt_negative_type}, and find that our choice of semantically similar yet corrupted examples yields the best performance.

\subsection{Identified circuits outperform or match the full model}

\paragraph{Standalone metrics of circuit performance}

We evaluate the effectiveness of our circuit relative to the full GPT-2 model, a fully corrupted counterpart, and a random complement circuit of equivalent size across two distinct tasks. The corrupted activations are created by caching activations on corrupted prompts, similar to our negative examples (see Appendix \ref{app:circuit_visualisation}). To measure a given circuit, we replace the activations of all attention heads not in the circuit with their corrupted activation. We use metrics specifically designed for each task, and our circuit is chosen by using the maximum F1 score across thresholds.

For the IOI task, the primary metric is \textit{logit difference}, calculated as the difference in logits between the indirect object's name and the subject's name. Our circuit achieves a logit difference of 3.62, surpassing the full GPT-2 model's average of 3.55, indicating that the correct name is approximately $e^{3.62}\approx 37.48$ times more likely than the incorrect name. However, our circuit performs slightly worse than the ground-truth circuit identified by \citet{wang2022interpretability}; full results are in Table \ref{tab:combined_performance_grouped} (see Appendix \ref{app:ioi} for details).

For the Greater-than task, we focus on \textit{probability difference} (PD) and \textit{cutoff sharpness} (CS), as defined by \citet{hanna2024does}. These metrics evaluate the model's effectiveness in distinguishing years greater than the start year and the sharpness of the transition between valid and invalid years (see Appendix \ref{app:greater_than} for formal details). Despite having fewer attention heads, our circuit achieves a PD of 76.54\% and a CS of 5.76\%, slightly outperforming the ground-truth circuit and significantly surpassing the clean GPT-2 model. The corrupted model and random complements exhibit negative PDs and negligible CS values; see Table \ref{tab:combined_performance_grouped}.

\begin{table}[htbp]
\centering
\scriptsize
\begin{tabular}{@{}lcccccc@{}}
\toprule
\multirow{2}{*}{\textit{Model/Circuit}} & \multirow{2}{*}{\textit{No. Heads}} & \multicolumn{2}{c}{\textit{\textbf{IOI}}} & \multicolumn{2}{c}{\textit{\textbf{Greater-than}}} \\ \cmidrule(lr){3-4} \cmidrule(lr){5-6}
& & \textit{Probability mult.} & \textit{Logit Diff.} & \textit{Probability Diff.} & \textit{Sharpness} \\ 
\midrule
GPT-2 (Clean) & 144 & 34.88 & 3.55 & 76.96\% & 5.57\% \\
GPT-2 (Corrupt) & 144 & 0.03 & -3.55 & -40.32\% & -0.06\% \\
Ground-truth & 26 & 61.14 & 4.11 & 71.30\% & 5.50\% \\
Ours & 40 & 37.48 & 3.62 & 76.54\% & 5.76\% \\
Random comp. & 40 & 0.23 & -2.23 & -37.91\% & -0.04\% \\
\bottomrule
\end{tabular}
\caption{Different standalone metrics of circuit performance for the IOI and Greater-than (GT) tasks, using a clean model, corrupted model, ground-truth circuit and random circuit.}
\label{tab:combined_performance_grouped}
\end{table}

\paragraph{Faithfulness of IOI and Greater-than circuits}

In the absence of a ground-truth circuit, evaluating whether our learned circuit reflects the true circuit used by the model is challenging. To this end, we employ the concept of \textit{faithfulness} introduced by \citet{marks2024sparse}. Faithfulness represents the proportion of the model's performance that our identified circuit explains, relative to the baseline performance when no specific input information is provided. We measure faithfulness by selecting a threshold $\theta$ to determine which heads to include in the circuit and ablating all other heads by replacing them with corrupted activations. Faithfulness is computed as $\frac{m(C)-m(\emptyset)}{m(M)-m(\emptyset)}$, where $m(C)$, $m(\emptyset)$, and $m(M)$ are the average performance metrics over the dataset $\mathcal{D}$ for the identified circuit, all heads ablated, and the full model, respectively. By sweeping over all $\theta$, we track performance improvement as we add circuit components. For comparison, we randomly select $n$ heads to use clean ablations for at each $\theta$, repeat this sampling 10 times, and average the metrics.



\begin{figure}[htbp]
    \centering
    \begin{subfigure}{0.49\textwidth}
        \includegraphics[width=\textwidth]{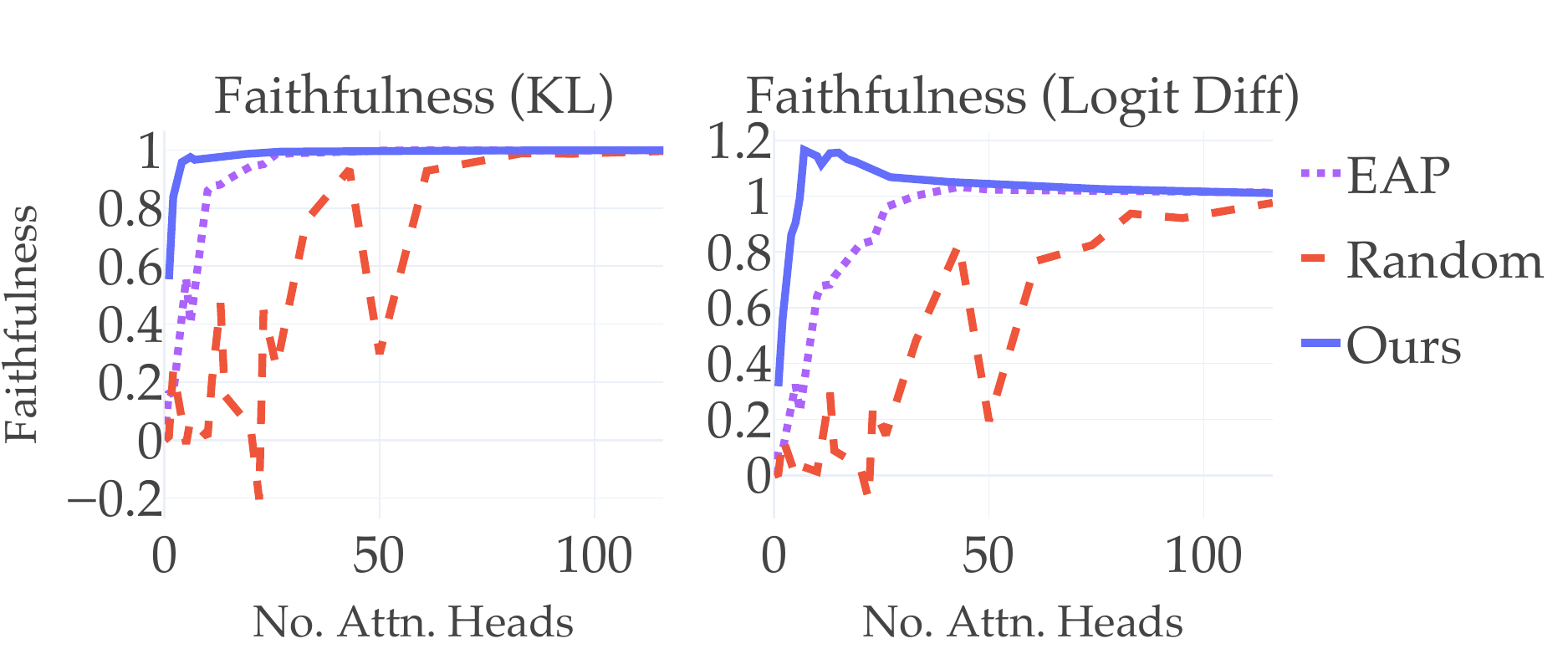}
        \caption{IOI}
        \label{fig:faithfulness_ioi}
    \end{subfigure}
    \hfill 
    \begin{subfigure}{0.49\textwidth} 
        \includegraphics[width=\textwidth]{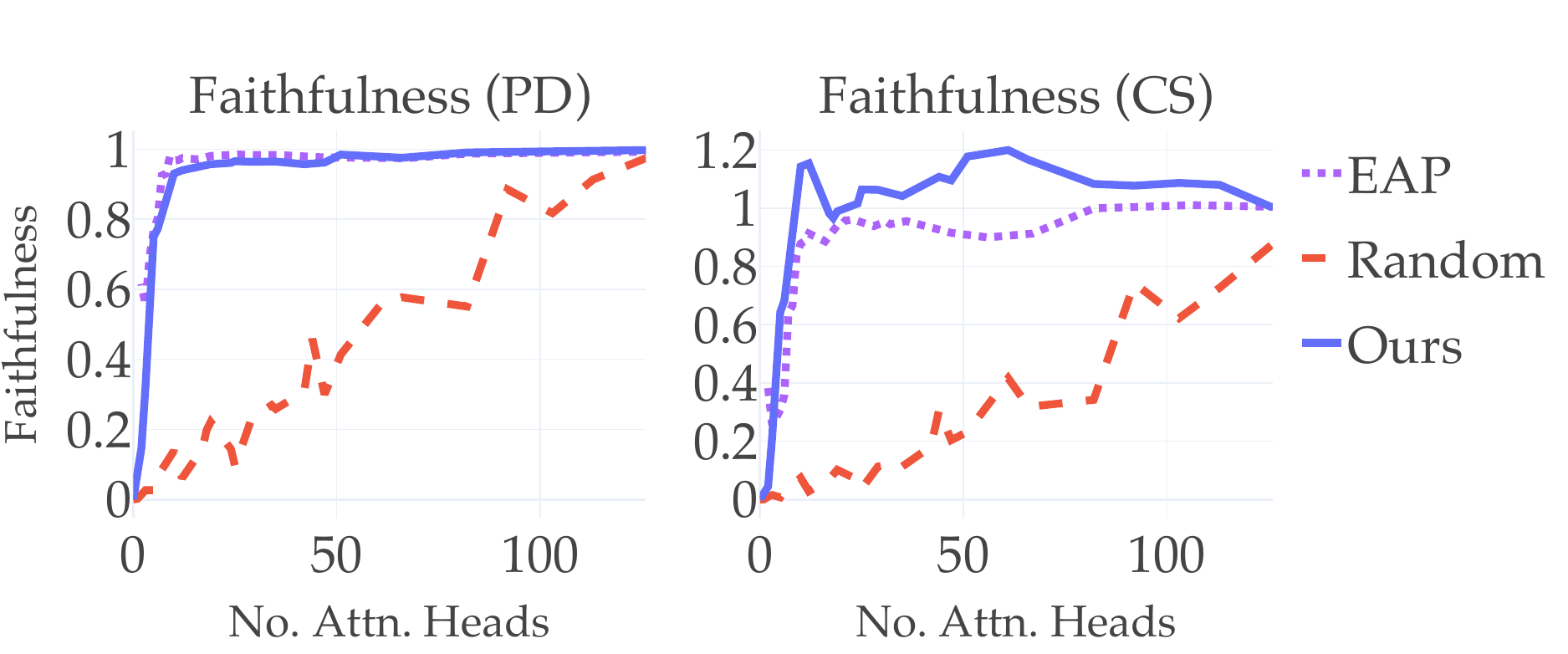}
        \caption{Greater-than}
        \label{fig:faithfulness_gt}
    \end{subfigure}
    \caption{Faithfulness of our learned circuits, circuits from edge attribution patching (EAP), and randomly selected circuits of equivalent size for the (a) IOI and (b) Greater-than tasks. Our circuits quickly approach or surpass the full model's performance as attention heads are added in order of importance. We outperform or match the performance of EAP at all thresholds for all metrics. Faithfulness of 1 indicates complete agreement with the unablated model.} 
    \label{fig:faithfulness}
\end{figure}

Our results are shown in Figure \ref{fig:faithfulness}. We also show the same faithfulness and metric curves applied to \textit{edge attribution patching} (EAP) \citep{syed2023attribution}. As we add attention heads from our circuit in order of threshold, performance quickly approaches that of the full model across all metrics and, in some cases, even outperforms the full model with considerably fewer heads. Importantly, our predicted circuit performs better or equal in all metrics than EAP.

\section{Related Work}
\label{sec:related_work}

\subsection{Sparse and Discrete Representations for Circuit Discovery}

Sparse and discrete representations of transformer activations have gained attention for their potential to enhance model interpretability. \citet{sharkey2022taking} and \citet{bricken2023towards} are generally considered the first groups to explore sparse dictionary learning to untangle features conflated by superposition, where multiple features are distributed across fewer neurons in transformer MLPs. Their work highlighted the utility of sparse representations but does not fully address the identification of computational circuits. \citet{kissane2024sparse} were the first to show that SAEs also learn useful representations when applied to attention heads rather than MLPs, and scaled this to GPT-2 \citep{kissane2attention, radford2019language}.

\citet{tamkin2023codebook} integrated a vector-quantized codebook into the transformer architecture. This technique demonstrates that incorporating discrete, interpretable features incurs only modest performance degradation and facilitates causal interventions. However, it necessitates architectural modifications, rendering it redundant for interpreting existing large-scale language models. \citet{cunningham2023sparse} used recursive analysis to trace the activation lineage of target dictionary features across layers. While this offers insights into layer-wise contributions, it falls short of mapping these activations to specific model components or elucidating their role within the residual stream. 

Most closely related to our work and conducted in parallel is that of \citet{marks2024sparse}, who employed a large SAE trained on diverse components, defining a framework for explicitly finding circuits. Their method relies on attribution patching (see below), which introduces practical difficulties at scale and again relies on a choice of metric. Additionally, their approach requires an SAE trained on millions of activations with significant upward projection to the dictionary, making it impractical for identifying specific circuits. Similarly, \citet{he2024dictionary} used SAE-learned features to map attention head contributions to identified circuits. However, their approach still uses a form of attribution patching and suggests a tendency for identified features to be polysemantic.

\subsection{Ablation and Attribution-Based Circuit Discovery Methods}

Ablation-based methods are fundamental in identifying critical components within models. \citet{conmy2024towards} introduced the ACDC algorithm, which automatically determines a component's importance by looking at the model's performance on a chosen metric with and without that component ablated. ACDC explores different ablation methods, such as replacing activations with zeros \citep{olsson2022context}, using mean activations across a dataset \citep{wang2022interpretability}, or activations from another data point \citep{geiger2021causal}. Despite its effectiveness, ACDC is computationally intensive and sensitive to the choice of metric and type of intervention. The method often fails to identify certain critical model components even when minimising KL divergence between the subgraph and full model.

Subnetwork Probing (SP) and Head Importance Score for Pruning (HISP) are similar methods. SP identifies important components by learning a mask over internal components, optimising an objective that balances accuracy and sparsity \citep{cao2021low}. HISP ranks attention heads based on the expected sensitivity of the model to changes in each head's output, using gradient-based importance scores \citep{michel2019sixteen}. Both methods, however, are computationally expensive and sensitive to hyperparameters.

Recent advancements have addressed limitations of traditional circuit discovery methods. \citet{syed2023attribution} introduced Edge Attribution Patching (EAP), using linear approximations to estimate the importance of altering an edge in the computational graph from normal to corrupted states \citep{neelnandaAttributionPatching}, reducing the need for extensive ablations. However, EAP's reliance on linear approximations can lead to overestimation of edge importance and weak correlation with true causal effects. Additionally, EAP fails when the gradient of the metric is zero, necessitating task-specific metrics for each new circuit. \citet{hanna2024have} recently proposed Edge Attribution Patching with Integrated Gradients (EAP-IG) to address these issues, evaluating gradients at multiple points along the path from corrupted to clean activations for more accurate attribution scores. Future work will benchmark our method against EAP and EAP-IG to understand the tradeoffs of each.


\section{Discussion}

The alignment of SAE-produced representations with language model circuits has significant implications for the scalability and interpretability of circuit discovery methods. If the level of granularity required for feature components in the circuit is coarser than the original head output dimension, it suggests that SAEs can efficiently project down rather than up, corresponding to a low level of feature-splitting and a high level of abstraction in the terminology of \citet{bricken2023towards}. This finding is promising for the scalability of SAEs as circuit finders, especially when dealing with small datasets where the SAE is trained directly on positive/negative examples, eliminating the need for expensive training on millions of activations across all layers, heads, and components. The fact that we can learn sufficient representations by training the SAE on only 5-10 examples speaks to the scalability of our method. We will release the code upon acceptance.

Moreover, using SAEs for circuit discovery also eliminates the need for ablation, which all prior approaches rely on \citep{vig2020investigating, finlayson2021causal} to assess a component's indirect effect on performance as a proxy for importance \citep{pearl2022direct}. By directly examining features, we bypass the computational complexities and difficulties in choosing a metric for each different circuit. Further, using features themselves as circuit components makes them inherently interpretable, opening up the possibility of applying auto-interpretability techniques to features in circuits \citep{bills2023language}. The combination of automatic circuit identification and interpretable by-products represents a significant step towards the ultimate goal of mechanistic interpretability: the automatic \textit{identification} and \textit{interpretation} of circuits, at scale, in language models.

\subsection{Limitations}
\label{sec:limitations}

Our method has several limitations that will be addressed in future work. First, although we learn discrete representations of attention head outputs, the interpretability of these learned codes may still be limited. Further work is needed to map these codes to human-interpretable concepts and behaviours. Second, we require the generation of a dataset of positive and negative examples for a circuit. This means we cannot do unsupervised circuit discovery and must carefully craft negative examples that are semantically similar to the positive ones, but are still corrupted enough to switch off the target circuit. To address this limitation, we plan to apply techniques such as quanta discovery from gradients \citep{michaud2024quantization} to automatically curate our positive and negative token inputs.

In addition to these method-specific limitations, any circuit discovery method faces the fundamental limitation of relying on human-annotated ground truth. The circuits found by previous researchers through manual inspection may be incomplete \citep{wang2022interpretability} or include edges that are correlated with model behaviour but not causally active \citep{zhang2024best}. Further, SAEs have been shown to make pathological errors \citep{lesswrongReconstructionErrors}; until these are resolved, we may need to include these errors in the circuit discovery process itself (much like \cite{marks2024sparse}).

\subsection{Future Directions}

One promising direction for future exploration is investigating the compositionality of the identified circuits and how they interact to give rise to complex model behaviours. Developing methods to analyse the hierarchical organisation of circuits and their joint contributions to various tasks could provide a more comprehensive understanding of the inner workings of large language models. A key aspect of this research could involve applying autointerpretability methods \citep{bills2023language} to our learned features in discovered circuits.

Finally, extending our approach to other model components, such as feedforward layers and embeddings, could offer a more complete picture of the computational mechanisms underlying transformer-based models. By combining insights from different levels of abstraction, we can work towards developing a more unified and coherent framework for mechanistic interpretability, thus advancing our understanding of how transformer models process and generate language.



\printbibliography

\newpage

\tableofcontents

\appendix

\section{Methodology details}
\label{app:methodology_details}

\subsection{Sparse autoencoder architecture}

Much like \citet{bricken2023towards}, our sparse autoencoders (SAEs) consist of a single hidden layer with tied weights for the encoder \( E \) and decoder \( D \). The SAE learns a dictionary of basis vectors \( \mathbf{v}_j \in \mathbb{R}^{d_\text{model}} \) such that each attention head output \( \mathbf{h}_i \in \mathbb{R}^{d_\text{model}} \) can be approximated as a sparse linear combination of the dictionary elements:

\[ \mathbf{h}_i \approx \sum_{j=1}^{d_\text{bottleneck}} z_{i,j} \mathbf{v}_j, \]

where \( z_{i,j} \) are the sparse activations and \( d_\text{bottleneck} \) is the dimensionality of the bottleneck layer. Our SAEs use the following parameters:

\[ W_E \in \mathbb{R}^{d_\text{bottleneck} \times d_\text{model}}, W_D \in \mathbb{R}^{d_\text{model} \times d_\text{bottleneck}}, \quad b \in \mathbb{R}^{d_\text{model}} \]

The columns of \( W_D \) are constrained to be unit vectors, representing the dictionary elements \( \mathbf{v}_j \). Given an input attention head output \( \mathbf{h}_i \), the activations of the bottleneck layer are computed as:

\[ \mathbf{z}_i = \text{ReLU}(W_E(\mathbf{h}_i - \mathbf{b}) + \mathbf{b}_E) \]

and the reconstructed output is obtained via:

\[ \hat{\mathbf{h}}_i = W_D \mathbf{z}_i + \mathbf{b}_D, \]

where the tied bias \( \mathbf{b} \) is subtracted before encoding and added after decoding. The dimensionality of the bottleneck layer \( d_\text{bottleneck} \) can be either larger (projecting up) or smaller (projecting down) than the input dimensionality \( d_\text{model} \). Hyperparameter sweeps found that projecting down (using fewer dimensions in the bottleneck layer than the dimension of the input) worked best for circuit identification. Additionally, we use a custom backward hook to ensure the dictionary vectors maintain unit norm by removing the gradient information parallel to these vectors before applying the gradient step.

\subsection{Training the sparse autoencoder}

We collate 250 and 250 positive examples for each dataset. We randomly sample 10 examples for training the SAE from this collection of examples, unless otherwise specified. We tokenise these examples and stack them into a single tensor of shape $n_\text{examples} \times n_\text{heads} \times d_\text{model}$, which can easily be passed through the SAE in a single forward pass. We then cache all attention head results in all layers - these are the inputs to our sparse autoencoder.

The SAE is trained to minimise a loss function that includes a reconstruction term and a sparsity penalty, controlled by the hyperparameter $\lambda$:

$$
\mathcal{L} = \sum_{i=1}^{n_\text{heads}} \left\lVert \mathbf{h}_i - \sum_{j=1}^{d_\text{bottleneck}} z_{i,j} \mathbf{v}_j \right\rVert_2^2 + \lambda \sum_{i=1}^{n_\text{heads}} \sum_{j=1}^{d_\text{bottleneck}} |z_{i,j}|.
$$

where $\lambda$ is typically about 0.01, our learning rate is 1e-3, and we train for 500 epochs using the Adam optimiser. We use a single NVIDIA Tesla V100 Tensor Core with 32GB of VRAM for all experiments.

\subsection{Counting the unique positive occurrences and co-occurrences}

\paragraph{Edge-level: co-occurrences}

Edge-level circuit identification aims to predict which attention heads are part of a circuit by analysing patterns in how the heads perform similar computation in tandem, rather than in isolation.

The first step is to construct two co-occurrence matrices, $\mathbf{C}^+$ and $\mathbf{C}^-$, which capture how often different token codes co-occur between each pair of heads in the positive and negative examples, respectively. For instance, $C^+_{h1,h2,c1,c2}$ counts the number of times code c1 in head h1 occurs together with code $c2$ in head $h2$ across the positive examples. $\mathbf{C}^-$ does the same for the negative examples.

Next, we compute a matrix $\mathbf{U}$ that identifies code co-occurrences that are unique to the positive examples for each head pair. An entry $U_{h1,h2}$ sums up the number of positive-only co-occurrences between heads $h1$ and $h2$ - that is, cases where a code pair has a positive count in $\mathbf{C}^+$ but a zero count in $\mathbf{C}^-$ for that head pair.

Intuitively, $\mathbf{U}$ captures which head pairs tend to jointly attend to particular token patterns more often in positive examples compared to negative examples. Head pairs with high values in $\mathbf{U}$ are stronger candidates for being part of the relevant circuit.

The head pairs are then sorted in descending order by their $\mathbf{U}$ values. To build the predicted circuit, we take the top $k$ head pairs from this sorted list, where $k$ is a hyperparameter. For each of these top $k$ pairs $(h1, h2)$, we increment the entries for $h1$ and $h2$ in a vector $\mathbf{u}$. This vector keeps track of how many times each head appears in the top head pairs. The reason for using only the top $k$ pairs is that including all pairs would make each head co-occur with every other head once, leading to a uniform $\mathbf{u}$ that would not distinguish between heads. 

\begin{figure}
    \centering
    \includegraphics[width=\textwidth, trim=465pt 0 50pt 0, clip]{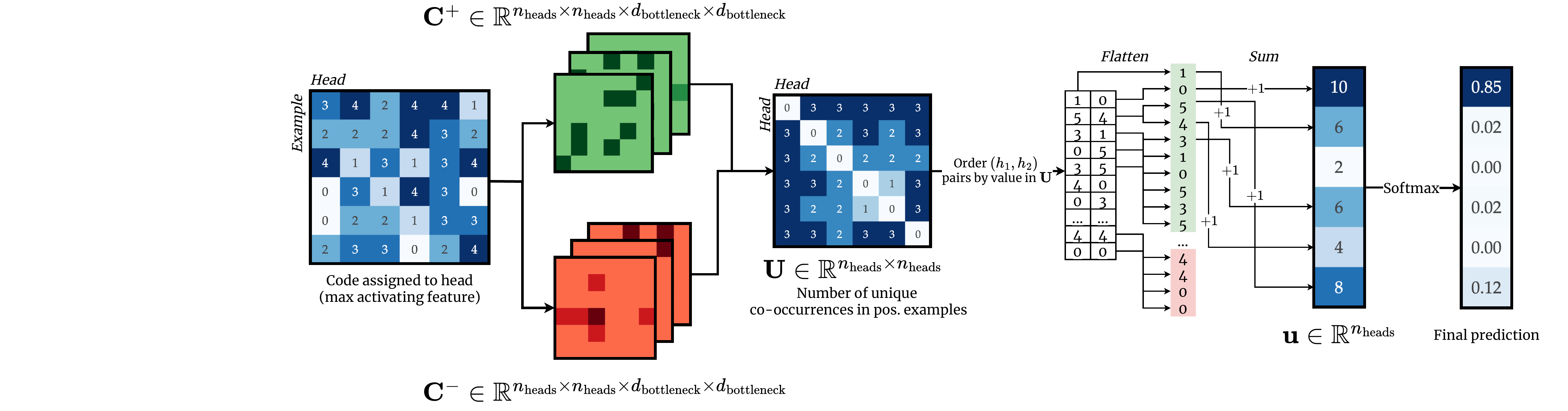}
    \caption{Edge-level circuit identification process for a 1-layer transformer with 6 attention heads, and 6 examples (first 3 rows are positive examples, last 3 are negative). Co-occurrence tensors $\mathbf{C}^+$ and $\mathbf{C}^-$ are constructed from positive and negative examples, counting specific code co-occurrences between each pair of heads. Matrix $\mathbf{U}$ tabulates the number of unique co-occurrences in the positive examples for each head pair $(h1, h2)$. The top $k$ head pairs (shown in green) by U value are used to build the predicted circuit, incrementing the corresponding entries in vector $\mathbf{u}$. After softmax normalisation, heads exceeding a threshold $\theta$ in $\mathbf{u}$ are predicted to be part of the circuit.}
    \label{fig:edge_method}
\end{figure}

Applying softmax to $\mathbf{u}$ normalises it into a probability distribution, allowing us to set a threshold $\theta$ to make the final predictions, with $\theta$ being the same scale for any arbitrary circuit (i.e. between 0 and 1). Heads that have a value exceeding $\theta$ are predicted to be part of the circuit. This whole process is outlined in Figure \ref{fig:edge_method}, where we step through an example of the process on a 1-layer transformer with 6 attention heads and using 6 text examples (3 positive, and 3 negative).

\subsection{Indication of wall-time as the underlying language models scale}
\label{app:wall_time}

A key benefit of our method over existing approaches is its efficiency. Whilst ACDC takes upwards of several hours to run on a V100 or A100 GPU for IOI on GPT2-Small \citep{conmy2024towards}, our method completes in under 3 seconds for GPT2-Small, and under 45 seconds for GPT2-XL. In fact, as we previously showed that one may be able to use only 10 examples when training the SAE, if this trend holds across model scales we can reduce time to less than 10 seconds for GPT2-XL.

\begin{figure}
    \centering
    \includegraphics[width=0.8\textwidth]{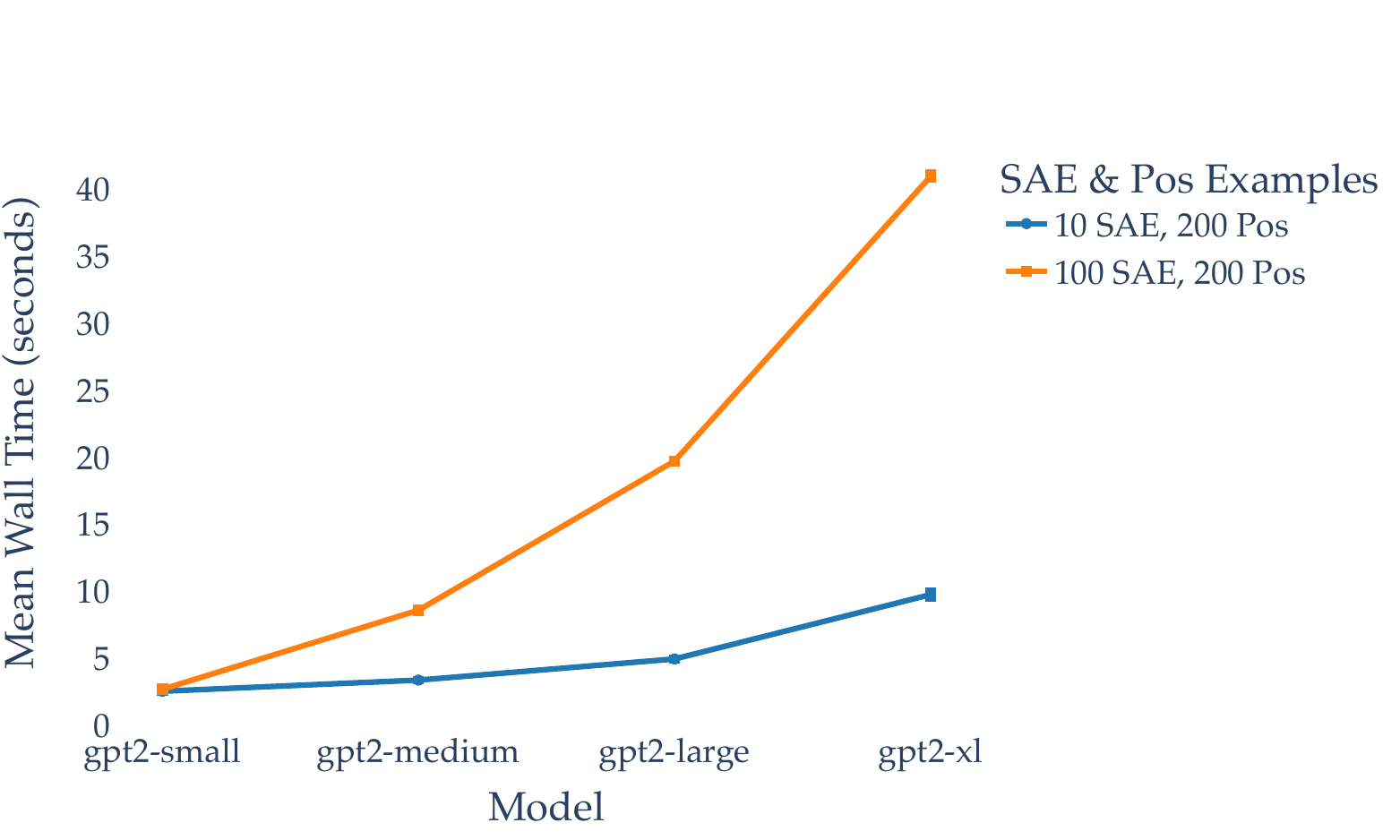}
    \caption{The wall-time in seconds taken to complete our entire circuit identification process, from training the SAE to sweeping over all thresholds $\theta \in [0,1]$ for predicting which heads are in the circuit. We use a single V100 for training and inference with the SAE. We show how this time scales as the model sizes grow. The 10 and 100 SAE refer to using 10 and 100 text examples to train the SAE, respectively. The 200 Pos refers to using 200 examples to count the number of unique positive codes.}
    \label{fig:wall_time}
\end{figure}

We show the specific model specifications in Table \ref{tab:gpt2specs}. If the number of text examples for both the SAE and counting positive codes remains constant, the main contribution to increased time for our method is $n_\text{heads}$ and $d_\text{model}$, as each example is in $\mathbb{R}^{n_\text{heads} \times d_\text{model}}$. Since the counting of unique positive codes involves elementary set operations over only a few hundred arrays of integer codes, it is only training the SAE that takes perceptibly longer as we increase the size of the underlying language model.

\begin{table}[htbp]
\centering
\begin{tabular}{@{}lllllllll@{}}
\toprule
Model & $n_{\text{params}}$ & $n_{\text{layers}}$ & $d_{\text{model}}$ & $n_{\text{heads}}$ & act\_fn & $n_{\text{ctx}}$ & $d_{\text{vocab}}$ & $d_{\text{mlp}}$ \\ \midrule
\textit{GPT-2 Small}  & 85M   & 12 & 768  & 12 & gelu & 1024 & 50257 & 3072 \\
\textit{GPT-2 Medium} & 302M  & 24 & 1024 & 16 & gelu & 1024 & 50257 & 4096 \\
\textit{GPT-2 Large}  & 708M  & 36 & 1280 & 20 & gelu & 1024 & 50257 & 5120 \\
\textit{GPT-2 XL }    & 1.5B  & 48 & 1600 & 25 & gelu & 1024 & 50257 & 6400 \\
\bottomrule
\end{tabular}
\caption{Specifications of GPT-2 Model Variants}
\label{tab:gpt2specs}
\end{table}

\section{Circuit visualisation and analysis}
\label{app:circuit_visualisation}

\begin{figure}[htbp]
    \centering
    \begin{subfigure}{0.32\textwidth} 
        \includegraphics[width=\textwidth]{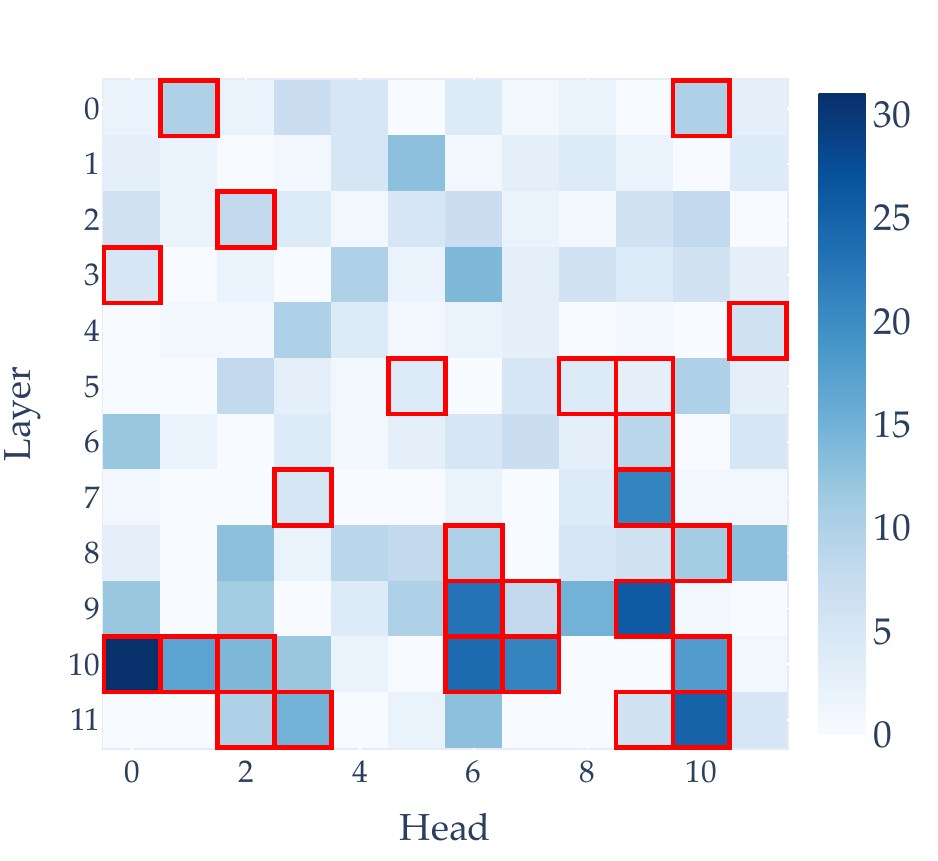}
        \caption{IOI}
        \label{fig:ioi_unique_to_positive}
    \end{subfigure}
    \begin{subfigure}{0.32\textwidth} 
        \includegraphics[width=\textwidth]{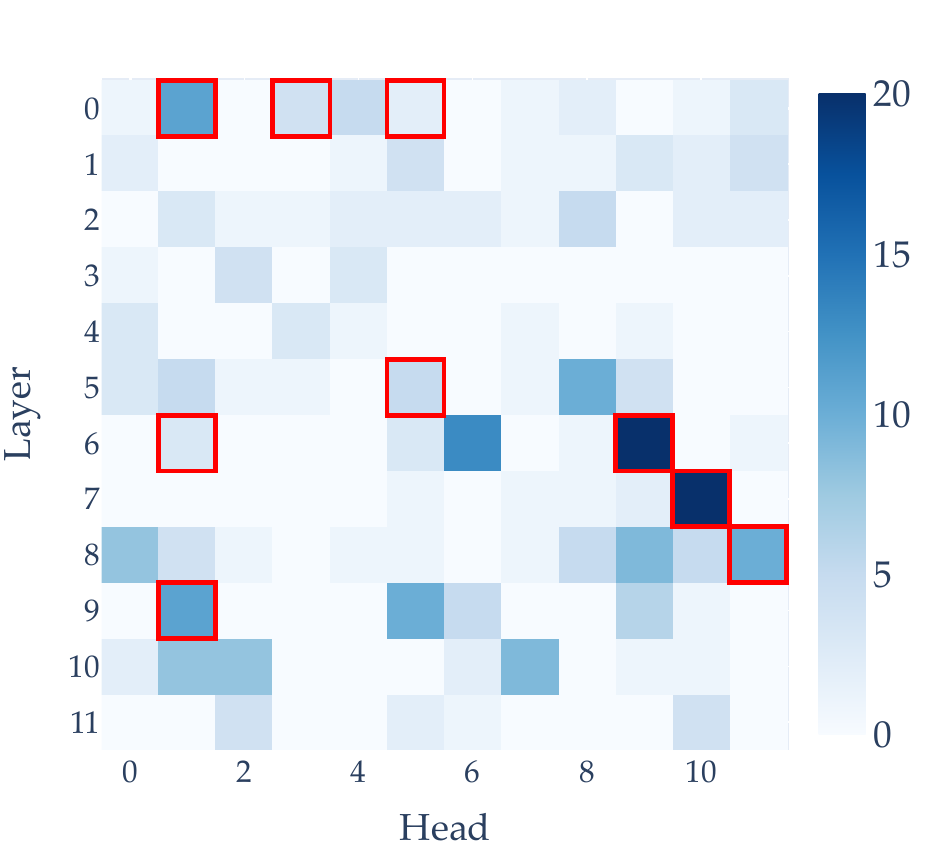}
        \caption{Greater-than}
        \label{fig:gt_unique_to_positive_vs_ground_truth}
    \end{subfigure}
    \begin{subfigure}{0.32\textwidth} 
        \includegraphics[width=\textwidth]{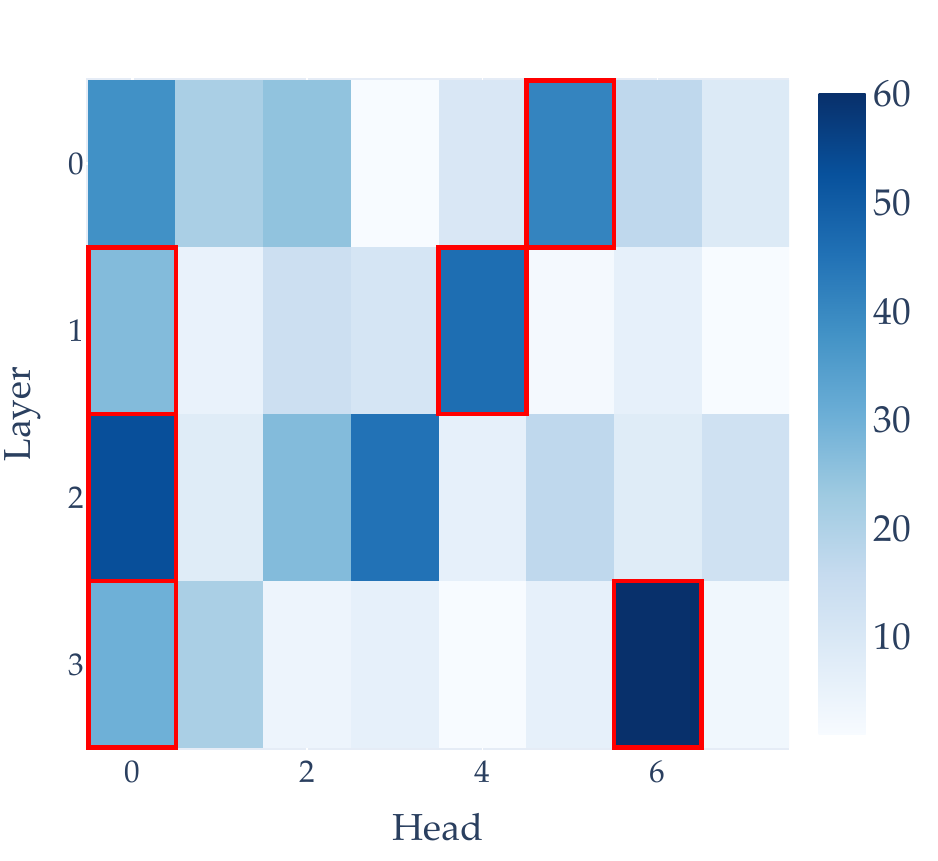}
        \caption{Docstring}
        \label{fig:ds_unique_to_positive_vs_ground_truth}
    \end{subfigure}
    \caption{Comparison of the number of codes unique to the positive examples by individual attention head, compared to a binary mask of whether than individual attention head is in the ground-truth circuit or not. The shade of blue shows how many unique positive codes that head has, and the cell for each head is outlined in red if it is in the canonical ground-truth circuit. Clearly, there is a strong positive correlation between the number of unique codes a head has and its presence in the circuit.}
    \label{fig:comparison_unique_to_positive_vs_ground_truth}
\end{figure}

Clearly, the number of unique positive codes per head is highly \textit{positively} correlated with the presence of that head in the ground-truth circuit, as seen in Figure \ref{fig:comparison_unique_to_positive_vs_ground_truth}. In this section, we provide further details on the predicted circuits and provide some further analysis.

\begin{figure}
    \centering
    \includegraphics[width=\textwidth]{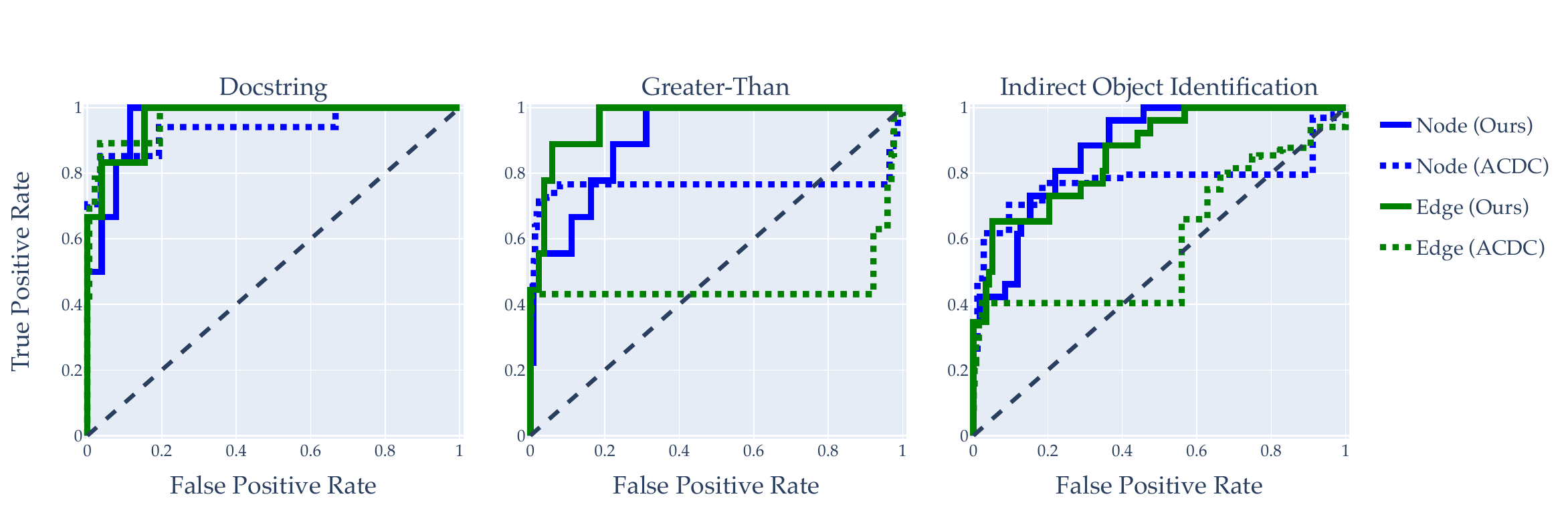}
    \caption{ROC curves for node-level and edge-level circuit identification. ACDC is shown using logit difference as the metric, also on both node-level and edge-level identification.}
    \label{fig:combined_roc_curves}
\end{figure}

\subsection{Indirect Object Identification}
\label{app:ioi}

\paragraph{Predicted circuit}
We compare the performance of our circuit to the full model, a fully corrupted model, and a random complement circuit of the same size. The metric is logit difference: the difference in logit between the indirect object's name and the subject's name. The full model's average logit difference is 3.55, meaning the correct name is $e^{3.55} \approx 34.88$ times more likely than the incorrect name.

To create a corrupted cache of activations, we run the model on the same prompts with the subject's name swapped. Replacing all attention heads' activations with these corrupted activations gives an average logit difference of -3.55. When testing our circuit, we replace activations for heads not in the circuit with their corrupted activations.

Our circuit has a higher logit difference (3.62) than the full GPT-2 model. The ground-truth circuit from \citet{wang2022interpretability} has a logit difference of 4.11. We compare this to the average logit difference (-1.97) of 100 randomly sampled complement circuits with the same number of heads as our circuit. These results are shown in Table \ref{tab:circuit_performance_ioi}.

We also provide the normalised logit difference: the logit difference minus the corrupted logit difference, divided by the signed difference between clean and corrupted logit differences. A value of 0 indicates no change from corrupted, 1 matches clean performance, less than 0 means the circuit performs worse than corrupted, and greater than 1 means the circuit improves on clean performance.

\begin{table}[htbp]
\centering
\begin{tabular}{@{}lcccc@{}}
\toprule
\textit{Model/circuit} & \textit{Attn. heads} & \textit{Logit Diff.} & \textit{Normalised Logit Diff.} & \textit{Probability Diff.} \\ 
\midrule
GPT-2 (Clean) & 144 & 3.55 & 1.0 & 34.88 \\
GPT-2 (Corrupted) & 144 & -3.55 & 0.0 & 0.03 \\
Ground-truth & 26 & 4.11 & 1.08 & 61.14 \\
Ours & 40 & 3.62 & 1.01 & 37.48 \\
Random complement & 40 & -2.23 & 0.19 & 0.23 \\
\bottomrule
\end{tabular}
\caption{Performance comparison of our circuit, the full GPT-2 model, corrupted GPT-2, ground-truth circuit, and random complement circuit. Logit difference measures the difference in logit between the correct and incorrect names. Normalised logit difference is the logit difference minus the corrupted logit difference, divided by the signed difference between clean and corrupted logit differences. Probability difference is the ratio of probabilities for the correct and incorrect names. Our predicted circuit actually improves on the performance of the full model, albeit not as much as the ground-truth circuit.}
\label{tab:circuit_performance_ioi}
\end{table}

\begin{figure}[ht]
    \centering
    \begin{minipage}{0.4\textwidth}
        \centering
        \includegraphics[width=\textwidth]{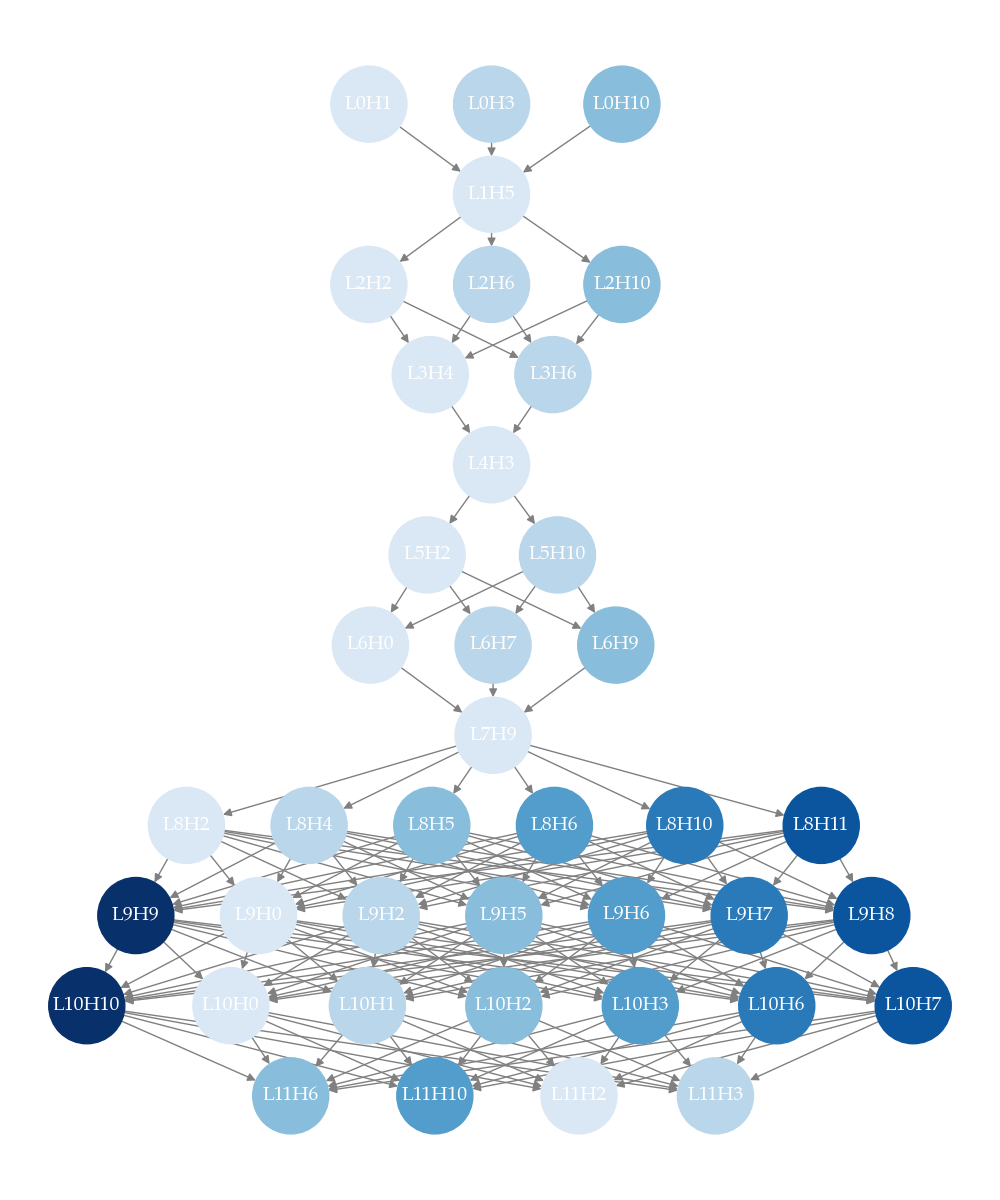}
    \end{minipage}\hfill
    \begin{minipage}{0.59\textwidth}
        \centering
        \includegraphics[width=\textwidth]{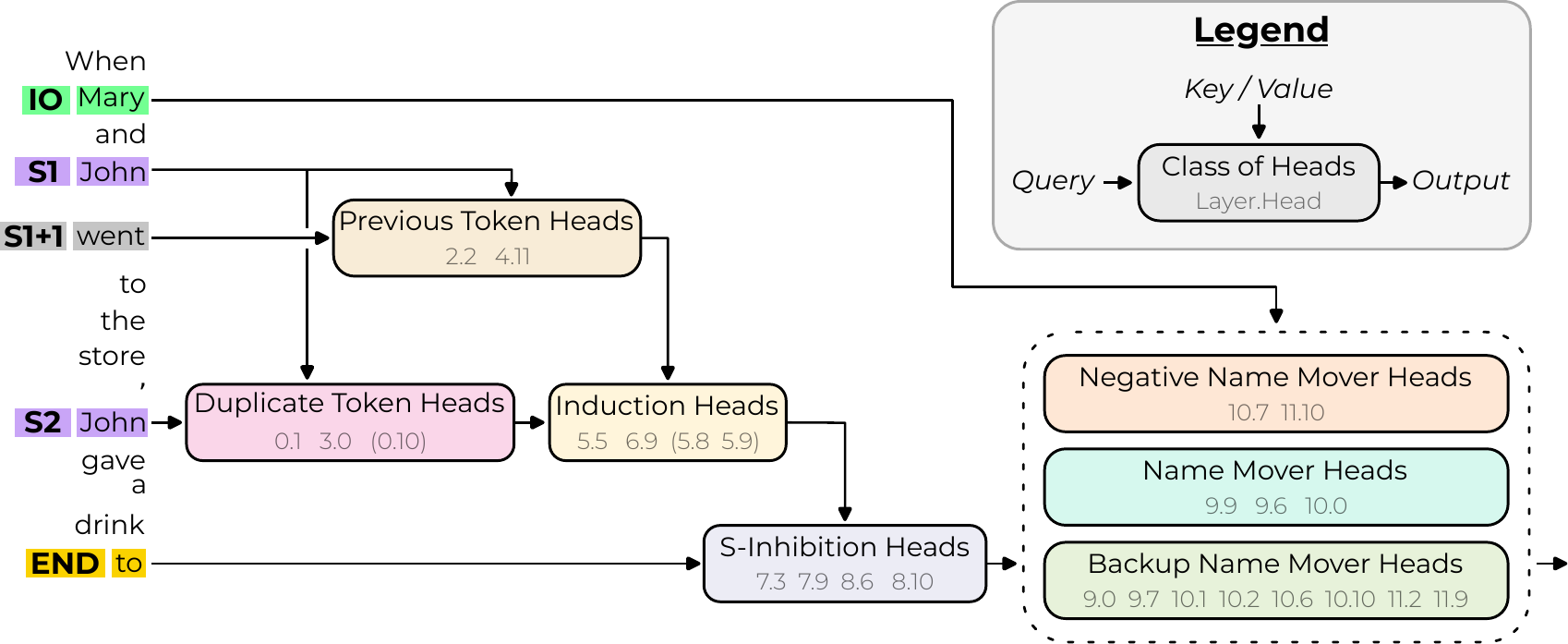}
    \end{minipage}
    \caption{The left figure shows our predicted IOI circuit, and the right is the canonical circuit from \cite{wang2022interpretability}. We by default include directed edges from all heads in a layer to all heads in the subsequent layer. The colour of the heads are darker the higher the softmaxed value of that head. Additionally, our circuit provides no information about position, as we aggregate over the position when caching our residual streams.}
    \label{fig:combined_ioi_figures}
\end{figure}

\paragraph{Negative name mover heads and previous token heads}
What seems to be a key advantage of our method over ACDC is our ability to detect both negative name mover heads and one of the previous token heads. \citet{wang2022interpretability} found that there exist attention heads in GPT-2 that actually write to the residual stream in the \textit{opposite} direction of the heads that output the remaining name in the sentence, called \textit{negative name-mover heads}. These likely ``hedge'' a model's predictions in order to avoid high cross-entropy loss when the sentence has a different structure, like a new person being introduced or a pronoun being used instead of the repeated name \citep{wang2022interpretability}. \textit{Previous token} heads copy information from the second name to the word after and have been found to have a minor role in the circuit.

\citet{conmy2024towards} found that they were unable to identify either of these types of heads as part of the circuit unless using a very low threshold $\tau$, which led to many extraneous heads being included in their prediction. This is despite the fact that negative name mover heads in particular are highly important in calibrating the confidence of name prediction in the circuit \citep{mcdougall2023copy}. The fact that we find both negative name mover heads (L10H7 and L11H10) and one of the previous token heads (L2H2) is highly promising evidence that our the distribution of our SAE activations provide a robust representation of the on-off nature of any given head in a circuit. Being able to identify negative components (that actively decrease confidence in predictions) in circuits is really important, because many circuits involve this general behaviour, known as \textit{copy suppression} \citep{mcdougall2023copy}.

\subsection{Greater-than}
\label{app:greater_than}

\paragraph{Setup details}
The Greater-than task focuses on a simple mathematical operation in the form it appears in text i.e. a sentence of the form ``The \texttt{<noun>} lasted from the year \texttt{XXYY} to the year \texttt{XX}'', where the aim is to give all non-zero probabilities to completions $>$ \texttt{YY}. We use the same setup as \citet{hanna2024does}. We use nouns which imply some form of duration, for example war, found used FrameNet \citep{baker1998berkeley}. The century \texttt{XX} is sampled from $\{12, \ldots, 18\}$ and the start year \texttt{YY} from $\{02, \ldots, 98\}$. Because of GPT-2's byte-pair encoding \citep{sennrich2015neural}, more frequent years are tokenised as single tokens (e.g. ``[1800]'' instead of ``[18][00]'') and so these are removed from the pool. Years ending in ``01'' and ``99'' are removed so as to ensure that there is at least one correct and incorrect valid tokenised answer.\footnote{Code to generate similar datasets can be found at \cite{hanna2024does}'s \href{https://github.com/hannamw/gpt2-greater-than}{Github repository}.}

\begin{figure}[ht]
    \centering
    \begin{minipage}{0.49\textwidth}
        \centering
        \includegraphics[width=\textwidth]{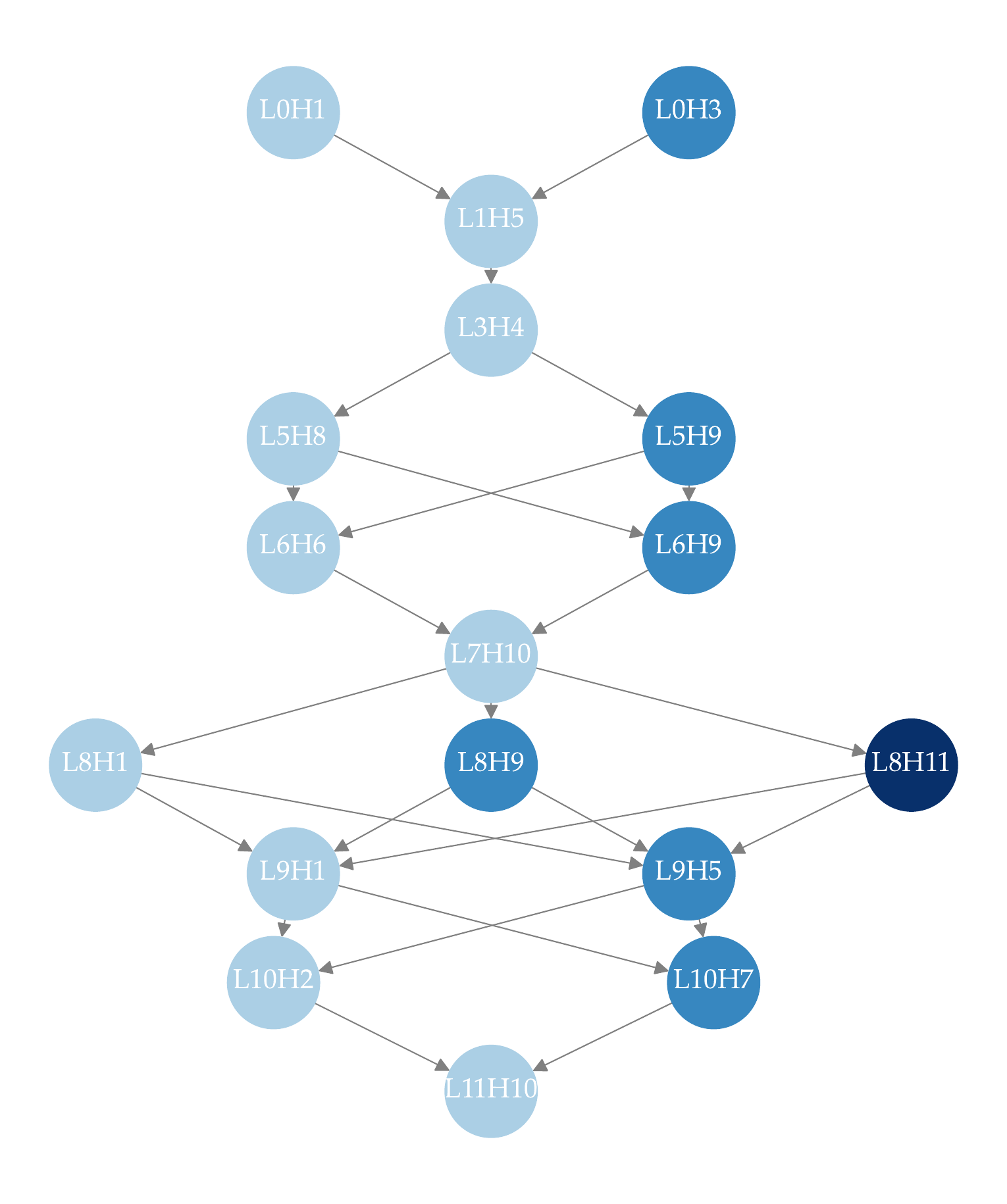}
    \end{minipage}\hfill
    \begin{minipage}{0.49\textwidth}
        \centering
        \includegraphics[width=\textwidth]{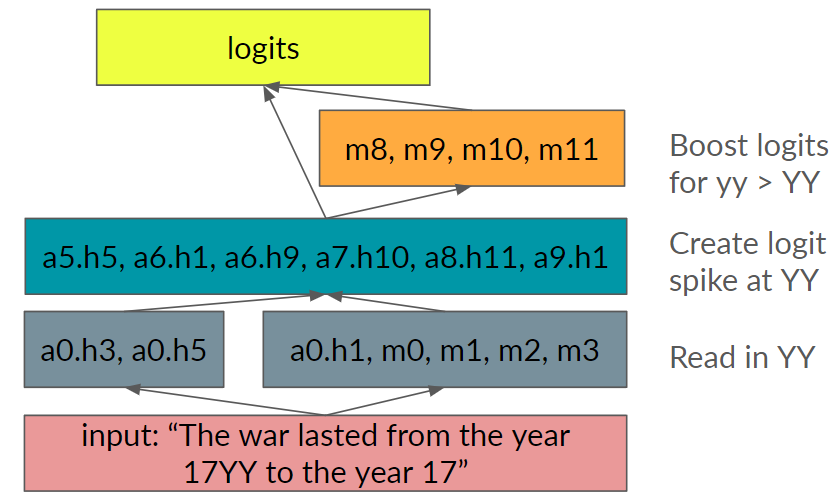}
    \end{minipage}
    \caption{Left is our predicted Greater-than circuit, and the right is the canonical circuit from \cite{hanna2024does}. }
    \label{fig:combined_gt_figures}
\end{figure}

\paragraph{Predicted circuit}
Figure \ref{fig:combined_gt_figures} shows our predicted circuit and the canonical ground-truth circuit from \citet{hanna2024does}. We have a high similarity, although we predict several heads from layers 10 and 11 that \citet{hanna2024does} attribute to MLP layers instead. It is possible that by only looking at attention head outputs and not MLP layers that these later layer heads appear to be doing the role that in actual fact is largely done by MLP layers. An interesting follow-up is examining why these later layer heads appear in our predicted circuit if they're not actually doing any useful computation for the task, or whether they might actually be doing some sort of relevant manipulation of the residual stream.

We again examine the performance of the predicted circuit in the context of the clean model and the ground-truth circuit from \cite{hanna2024does}. We produce a dataset of 100 examples according to the same process outlined above. The corrupted examples are the same prompts but the \texttt{YY} is replaced with ``01''. We define two metrics measuring the performance of the model, introduced by \citet{hanna2024does}.

Let $Y$ be the start year of the sentence, and $p_y$ be the probability assigned by the model to a two-digit output year $y$. The first metric, probability difference ($PD$), measures the extent to which the model assigns higher probability to years greater than the start year. It is calculated as:

\begin{equation}
PD = \sum_{y > Y} p_y - \sum_{y \leq Y} p_y
\end{equation}

Probability difference ranges from -1 to 1, with higher values indicating better performance in reflecting the greater-than operation. A positive value of $PD$ indicates that the model assigns higher probabilities to years greater than the start year, while a negative value suggests the opposite.

The second metric, cutoff sharpness ($CS$), quantifies the model's behaviour of producing a sharp cutoff between valid and invalid years. It is calculated as:

\begin{equation}
CS = p_{Y+1} - p_{Y-1}
\end{equation}

where $p_{Y+1}$ is the probability assigned to the year immediately following the start year, and $p_{Y-1}$ is the probability assigned to the year immediately preceding the start year. Cutoff sharpness also ranges from -1 to 1, with larger values indicating a sharper cutoff. Although not directly related to the greater-than operation, this metric ensures that the model's output depends on the start year and does not produce constant but valid output. A high value of $CS$ suggests that the model exhibits a sharp transition in probabilities between the years adjacent to the start year.

\begin{table}[htbp]
\centering
\begin{tabular}{@{}lccc@{}}
\toprule
\textit{Model/circuit} & \textit{Attention heads} & \textit{Probability Difference} & \textit{Cutoff Sharpness} \\ 
\midrule
GPT-2 (Clean) & 144 & 76.96\% ($\pm$ 26.82\%) & 5.57\% ($\pm$ 8.08\%) \\
GPT-2 (Corrupted) & 144 & -40.32\% ($\pm$ 55.28\%) & -0.06\% ($\pm$ 0.08\%) \\
Ground-truth & 9 & 71.30\% ($\pm$ 28.71\%) & 5.50\% ($\pm$ 6.89\%) \\
Ours & 29 & 76.54\% ($\pm$ 27.51\%) & 5.76\% ($\pm$ 7.42\%) \\
Random complement & 29 & -37.91\% ($\pm$ 55.76\%) & -0.04\% ($\pm$ 0.78\%) \\
\bottomrule
\end{tabular}
\caption{Performance comparison of our predicted circuit, the clean GPT-2 model, the corrupted GPT-2 model, the ground-truth circuit from \cite{hanna2024does}, and a random complement circuit on the greater-than task. The performance is measured using probability difference ($PD$) and cutoff sharpness ($CS$). The values represent the mean and standard deviation across 100 examples. Our predicted circuit achieves comparable performance to the clean GPT-2 model and the ground-truth circuit while being significantly smaller in size.}
\label{tab:circuit_performance_gt}
\end{table}

Table \ref{tab:circuit_performance_gt} presents the performance of our predicted circuit in comparison to the clean GPT-2 model, the corrupted GPT-2 model, the ground-truth circuit from [Hanna et al., 2024], and random complement circuits. The performance is measured using probability difference (PD) and cutoff sharpness (CS). Our predicted circuit, consisting of 29 attention heads, achieves a PD of 76.54\% and a CS of 5.76\%, slightly outperforming the ground-truth circuit (PD: 71.30\%, CS: 5.50\%), albeit having more heads. Notably, our circuit also surpasses the performance of the clean GPT-2 model (144 heads) in CS, and is essentially the same in PD. In contrast, the corrupted GPT-2 model and the average of 100 random complement circuits of the same size as our predicted circuit show negative PD and near-zero CS, indicating poor performance in capturing the greater-than operation and producing a sharp cutoff between valid and invalid years. These results demonstrate that our predicted circuit effectively captures the relevant information for the task while being significantly smaller than the full GPT-2 model.

\paragraph{Exploratory analysis of relationship between codes and year}
We also did some exploratory analysis of whether the learned activations from the encoder, and their corresponding codes, were associated at all with the starting year of the completion. For instance, were there codes that only activated for high year numbers, such as \texttt{<CENTURY>90} and above? If we only use 100 codes, do these codes roughly get distributed to particular years, so there is a soft bijection between codes and two-digit years?

To answer some of these questions, we created a new dataset consisting of prompts of the form ``The war lasted from the year \texttt{<century><year>} to \texttt{<century>}'', and trained a sparse autoencoder on the attention head outputs of GPT2-Small on these prompts, with 100 learned features.

\begin{figure}[htbp]
    \centering
    \begin{subfigure}[b]{0.45\textwidth} 
        \includegraphics[width=\textwidth]{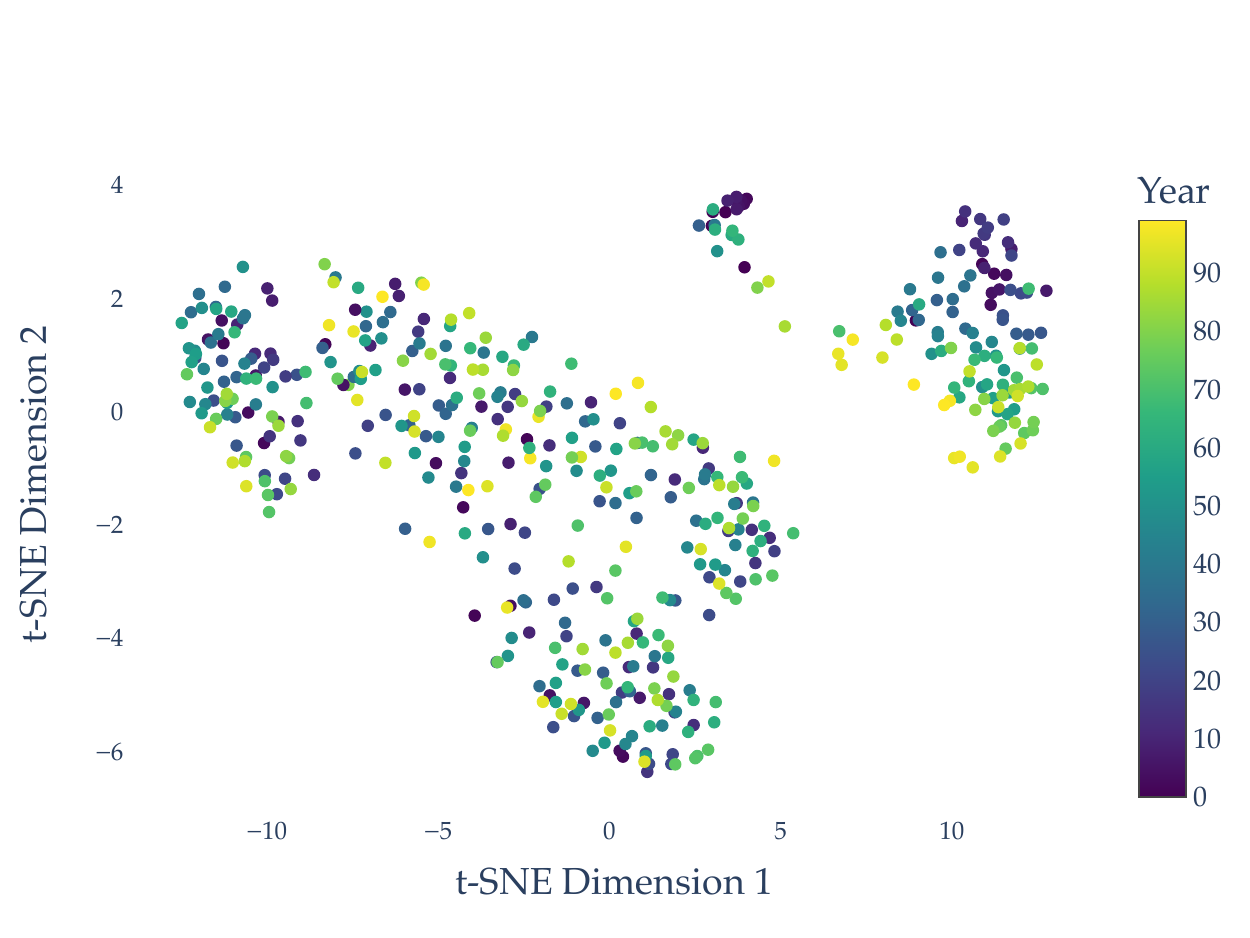}
        \caption{Coloured by year}
        \label{fig:tsne_plot_gt}
    \end{subfigure}
    \hfill 
    \begin{subfigure}[b]{0.45\textwidth} 
        \includegraphics[width=\textwidth]{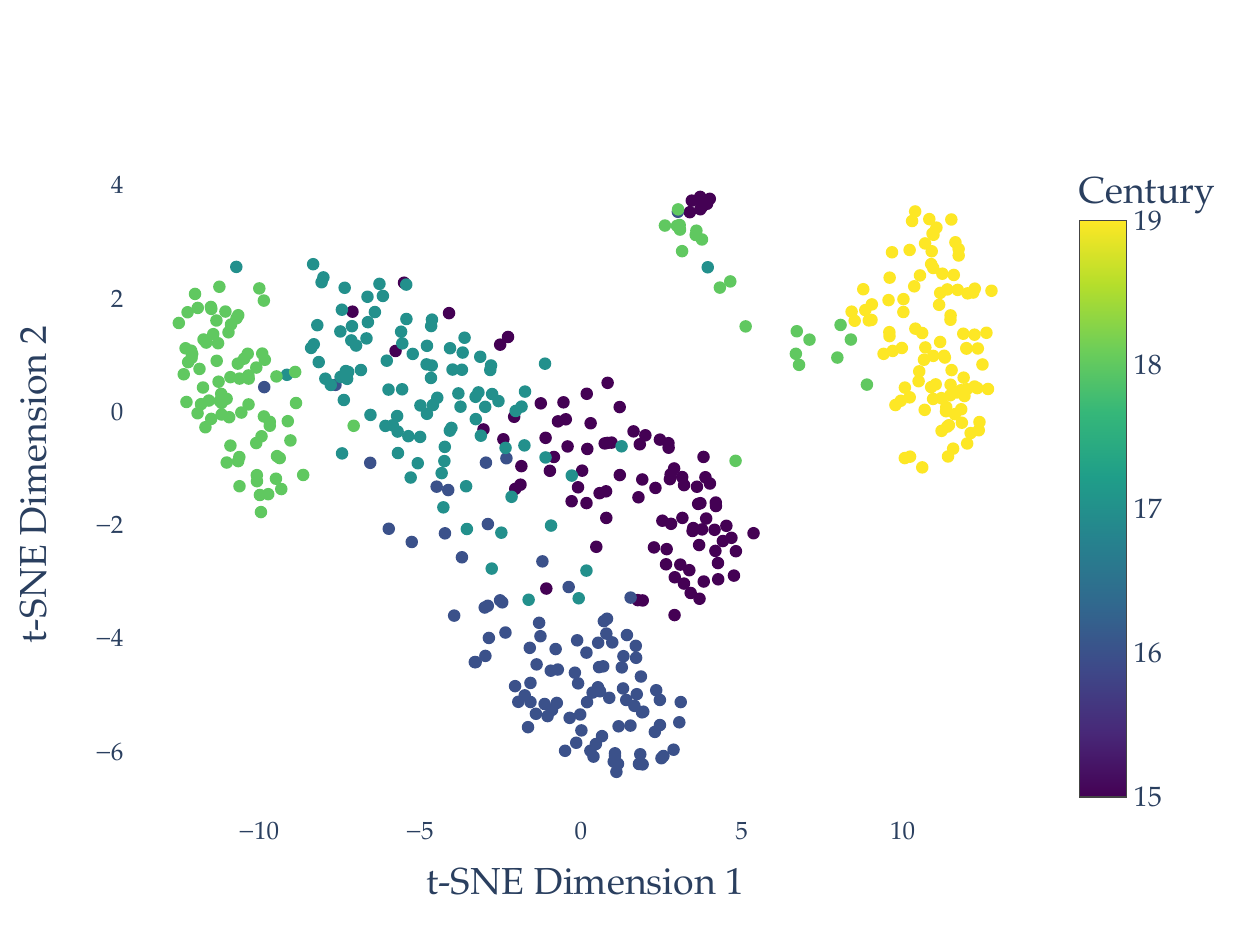}
        \caption{Coloured by century}
        \label{fig:tsne_plot_gt_century}
    \end{subfigure}
    \caption{Dimensionality reduction using t-SNE of the learned activations from the SAE encoder for all Greater-than examples and all heads, coloured by century (i.e. the 18 in ``The war lasted from \textbf{18}07 to 18'') and year (i.e. the 7 in ``The war lasted from 18\textbf{07} to 18''.)}
    \label{fig:dim_reduction_gt}
\end{figure}

We initially examined the t-SNE dimensionality reduction \citep{van2008visualizing} of embeddings for all examples across all heads, shown in Figure \ref{fig:dim_reduction_gt}. We colour the points by the year in the example (e.g. the 14 in ``The war lasted from 19\textbf{14} to 19''). Interestingly, we notice two distinct clusters of activations. The first, on the upper right in Figure \ref{fig:tsne_plot_gt}, seems to have a fairly well-defined transition between examples with low year numbers to examples with high year numbers. However, the other cluster (the lower left in both plots) appears to have no discernible order. This suggests that the SAE may be learning degenerate latent representations for examples that differ only in the century used. 

\begin{figure}
    \centering
    \includegraphics[width=0.5\textwidth]{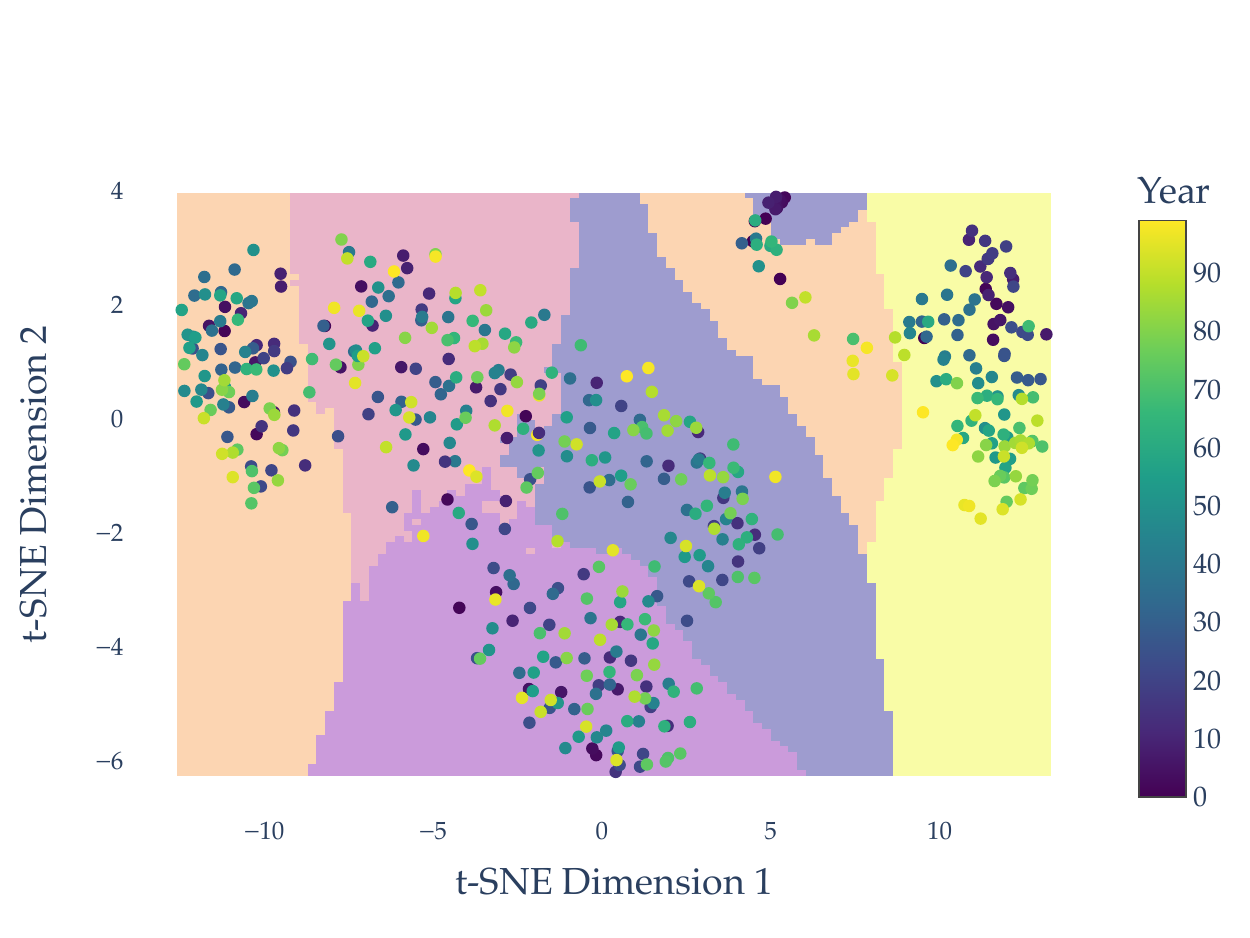}
    \caption{The same PCA as Figure \ref{fig:tsne_plot_gt}, but with the background being coloured by the mjority century of the $k=10$ nearest neighbours. Segmenting the plot in this way seems to make some of the transitions from low years to high years clearer (i.e. in the orange and yellow segments).}
    \label{fig:tsne_plot_gt_background}
\end{figure}

We then show the same plot, except colouring each example by the century of the example (e.g. the 19 in ``The war lasted from \textbf{19}14 to 19''). Incredibly, there is almost perfect linear separation between the classes (where the classes pertain to centuries). If we instead produce the same plot with the background being the century of the 10 nearest neighbouring points, some structure with regards to year of the example begins to emerge (Figure \ref{fig:tsne_plot_gt_background}). There seems to be a stronger gradient within groups, with the year number increasing linearly in a certain direction. However, there is still a significant amount of noise, and future research should examine why the SAE learns representations that focus more on the century than the year, when the year is evidently more important for successful completion of the Greater-than task.

\begin{figure}[htbp]
    \centering
    \begin{subfigure}[b]{0.45\textwidth}
        \includegraphics[width=\textwidth]{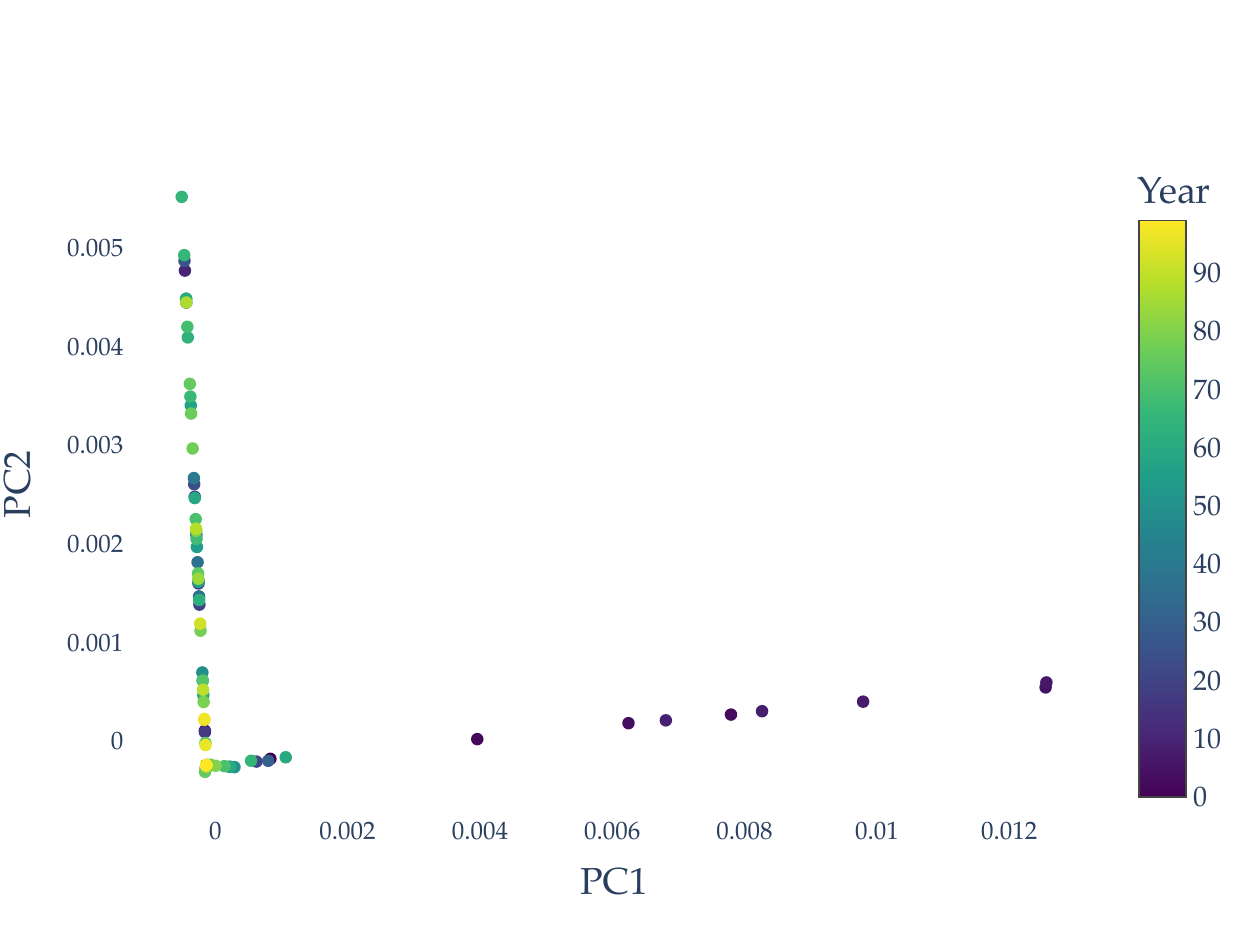}
        \caption{L1H0}
        \label{fig:pca_plot_gt_L1H0}
    \end{subfigure}
    \hfill 
    \begin{subfigure}[b]{0.45\textwidth}
        \includegraphics[width=\textwidth]{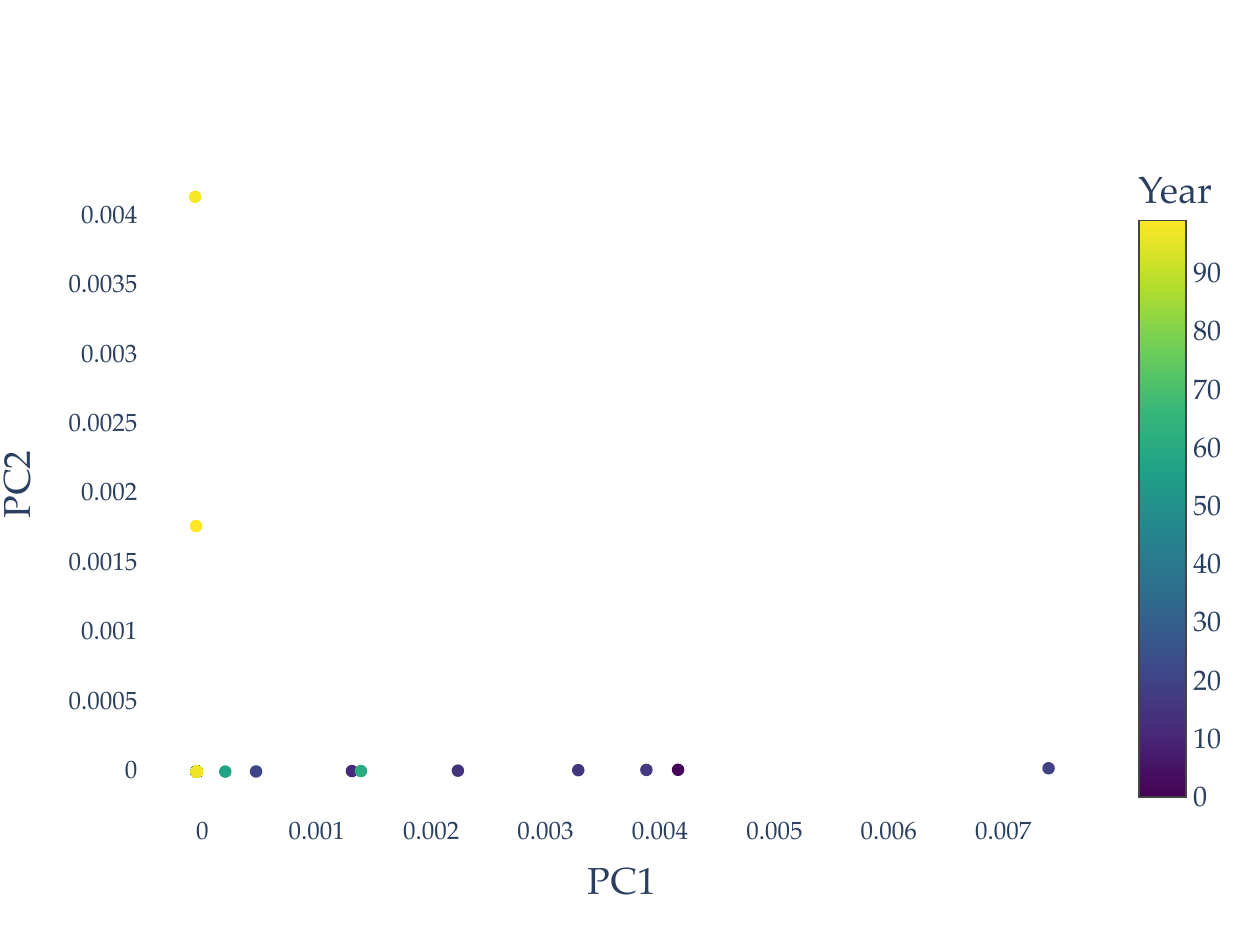}
        \caption{L6H0}
        \label{fig:pca_plot_gt_L6H0}
    \end{subfigure}
    
    
    \begin{subfigure}[b]{0.45\textwidth}
        \includegraphics[width=\textwidth]{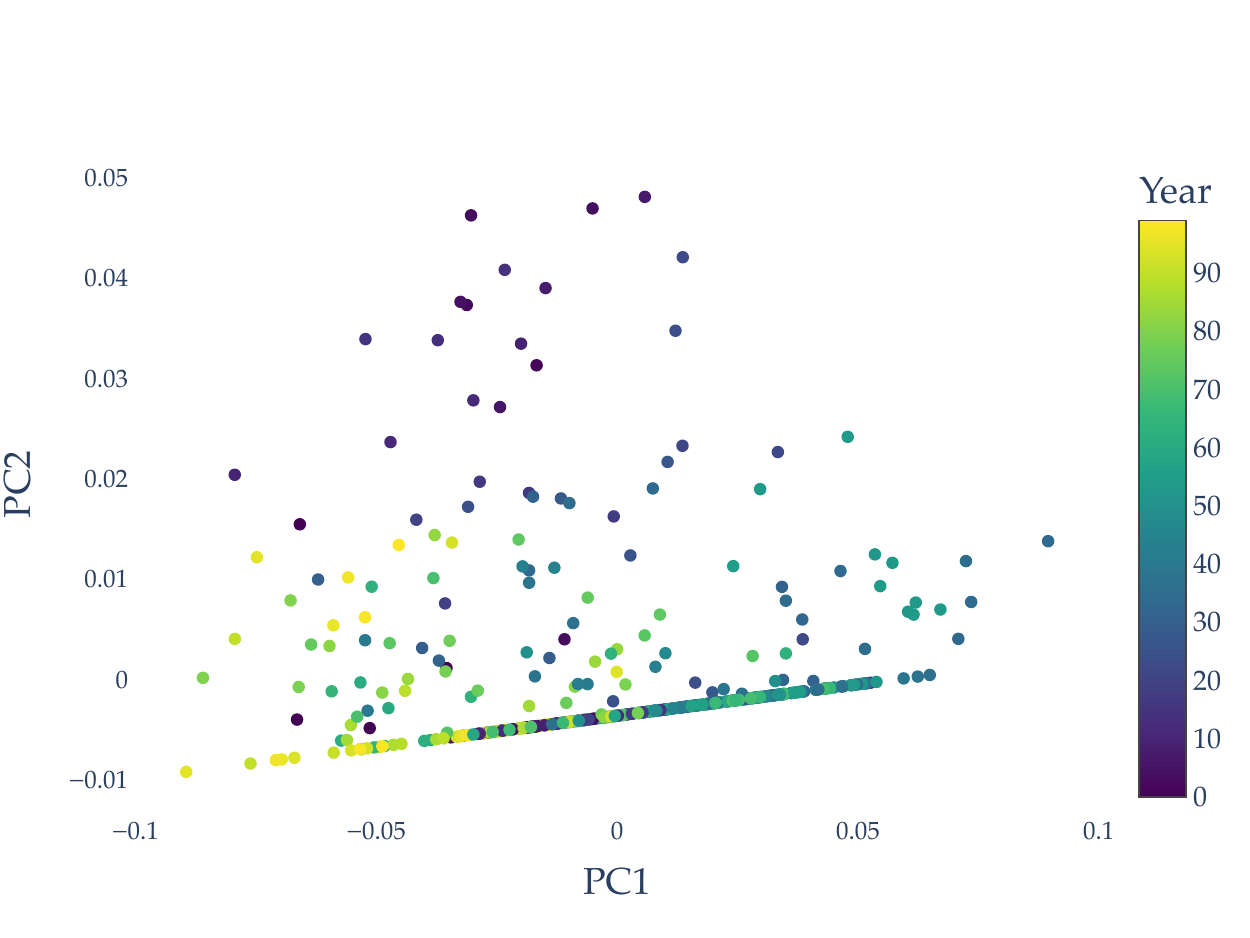}
        \caption{L10H8}
        \label{fig:pca_plot_gt_L10H8}
    \end{subfigure}
    \hfill 
    \begin{subfigure}[b]{0.45\textwidth}
        \includegraphics[width=\textwidth]{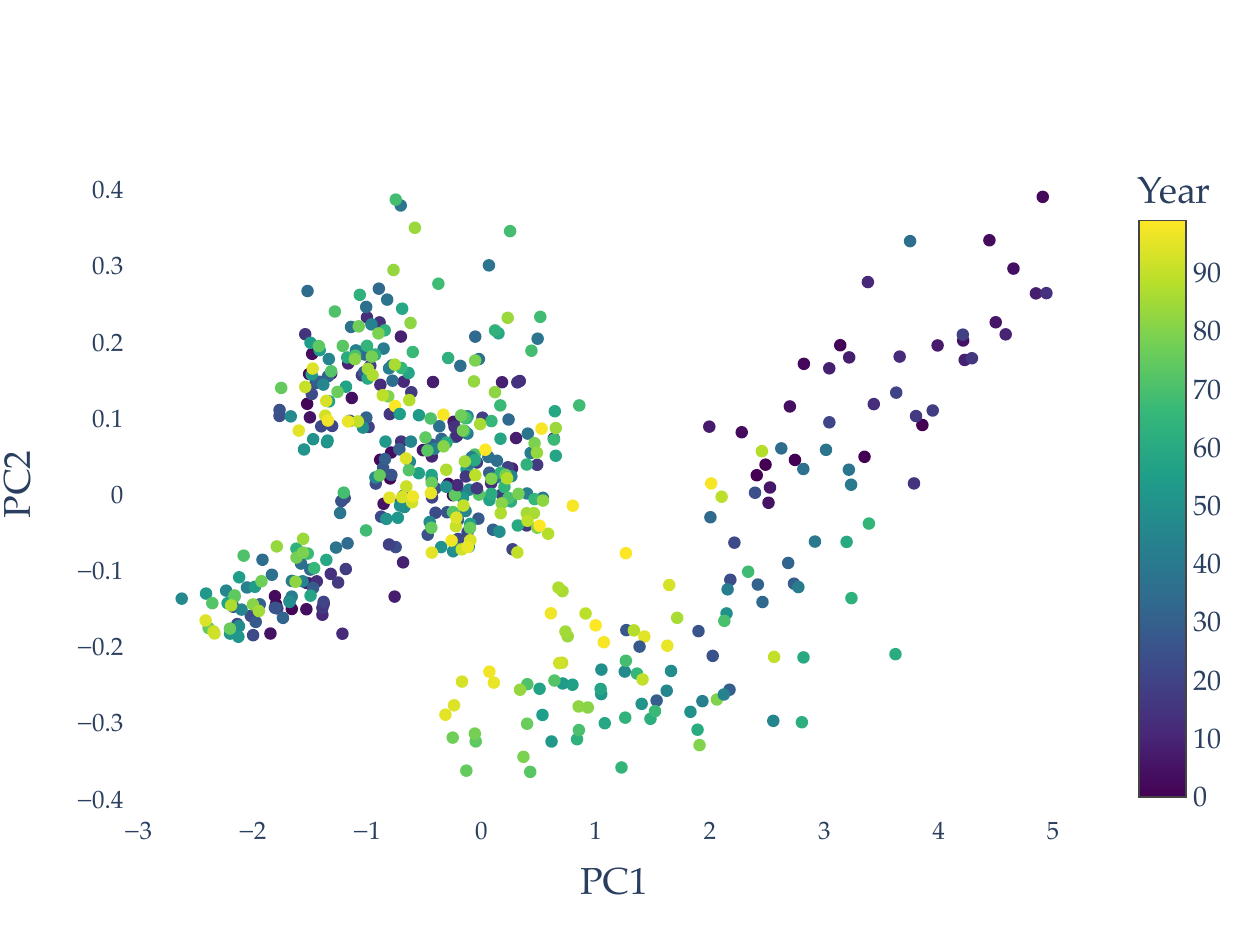}
        \caption{L11H8}
        \label{fig:pca_plot_gt_L11H8}
    \end{subfigure}
    \caption{PCA of learned activations from individual attention heads across examples, coloured by year number in the example.}
    \label{fig:individual_pcas}
\end{figure}

We also examined the top 2 principal components of the encoder activations on individual attention heads across examples to determine if they had some relationship to the year number in the example. This is shown in Figure \ref{fig:individual_pcas}. These four individual heads are selected to show a variety of behaviours. For some, like \ref{fig:pca_plot_gt_L1H0} and \ref{fig:pca_plot_gt_L6H0}, the principal components seem to directly correspond to ``low'' years and ``high'' years, with many examples in the approximately the same decade being mapped to almost exactly the same PCA-reduced point. Other heads, such as \ref{fig:pca_plot_gt_L10H8} and \ref{fig:pca_plot_gt_L11H8}, have significantly more variability, but seem to follow some gradient of transitioning from lower years to higher years as we move across the space.

\subsection{Docstring}

\begin{figure}[ht]
    \centering
    \begin{minipage}{0.39\textwidth}
        \centering
        \includegraphics[width=\textwidth]{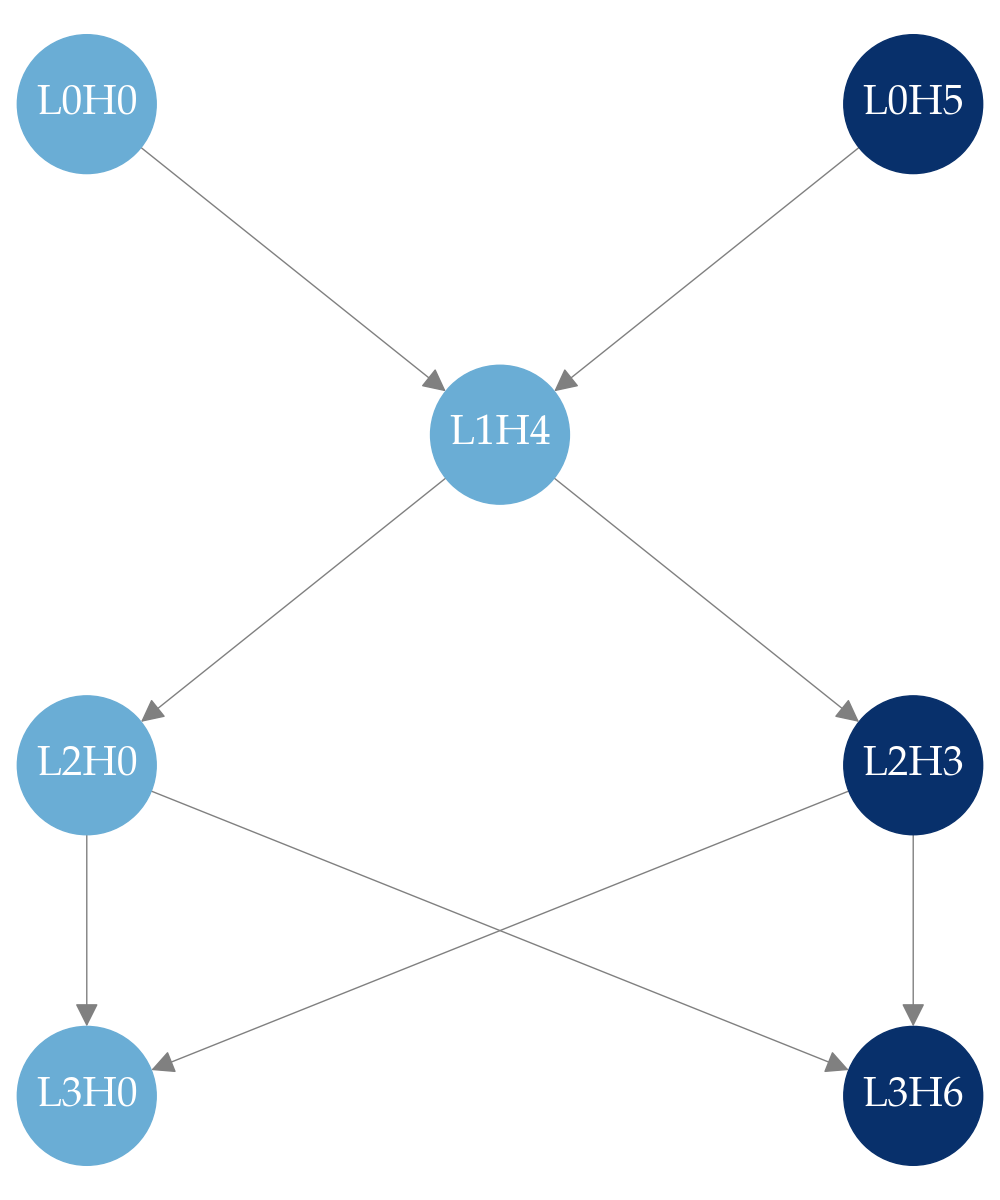}
    \end{minipage}\hfill
    \begin{minipage}{0.59\textwidth}
        \centering
        \includegraphics[width=\textwidth]{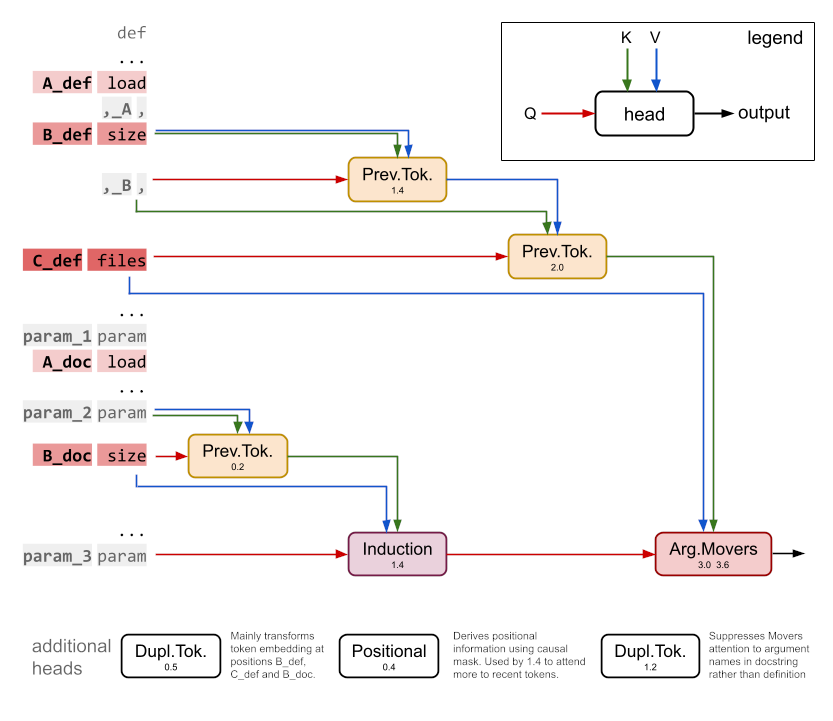}
    \end{minipage}
    \caption{Left is our predicted Docstring circuit, and right is the canonical circuit from \cite{heimersheim2023circuit}. }
    \label{fig:combined_ds_figures}
\end{figure}

Our predicted Docstring circuit is shown in Figure \ref{fig:combined_ds_figures}. Interestingly, our circuit identification method does not predict L0H2 and L0H4 as being part of the circuit, whereas \cite{heimersheim2023circuit} does. However, after running the ACDC algorithm (as well as HISP and HP, and manual interpretation) on the docstring circuit, \cite{conmy2024towards} concluded that these two heads are not relevant under the docstring distribution. The agreement between ACDC and our method with regard to these heads that are manually confirmed to not be part of the circuit is promising for the reliability of our approach.

\section{Detailed comparison to ACDC and other circuit identification methods}
\label{app:acdc}


It is important to clarify the distinctions and similarities between \citet{conmy2024towards}'s ACDC method and our approach. Our work builds upon ACDC, adapting its code, results, and experiments from their MIT-licensed GitHub repository.\footnote{\href{https://github.com/ArthurConmy/Automatic-Circuit-Discovery}{\texttt{https://github.com/ArthurConmy/Automatic-Circuit-Discovery}}} The primary workflow for ACDC begins by specifying the computational graph of the full model for the task or circuit under examination, alongside a threshold for the acceptable difference in a metric between the predicted circuit and the full model. This computational graph, represented using a correspondence class, includes nodes and edges that connect these nodes, typically representing components like attention heads, query/key/value projections, and MLP layers, with edges indicating the connections between these components.

\begin{figure}[htbp]
    \centering
    \begin{subfigure}[b]{0.32\textwidth}
        \centering
        \includegraphics[width=\textwidth]{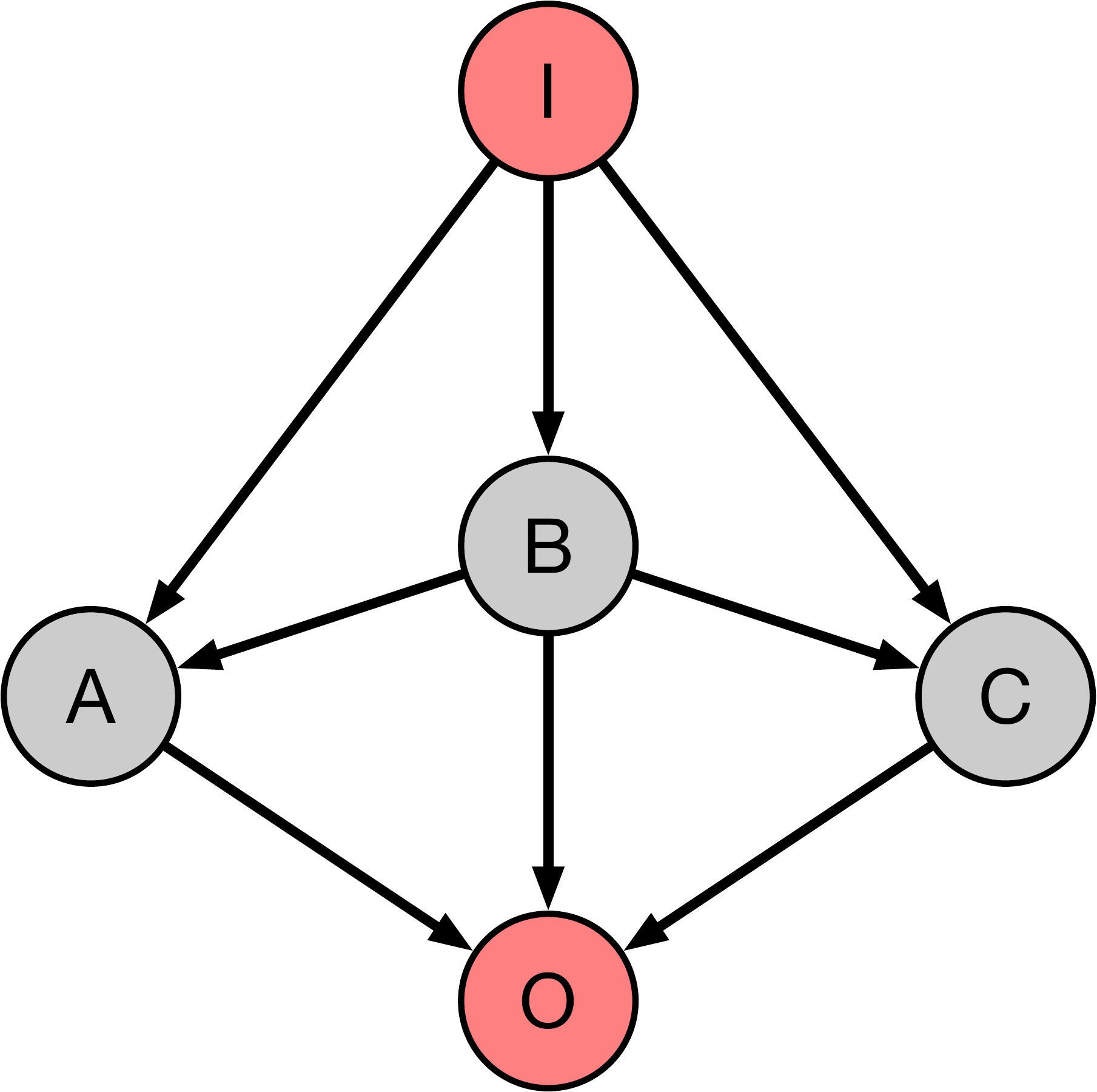}
        \caption{Step 2a: Define the computational graph and threshold $\tau$.}
        \label{fig:step2a}
    \end{subfigure}
    \hfill
    \begin{subfigure}[b]{0.32\textwidth}
        \centering
        \includegraphics[width=\textwidth]{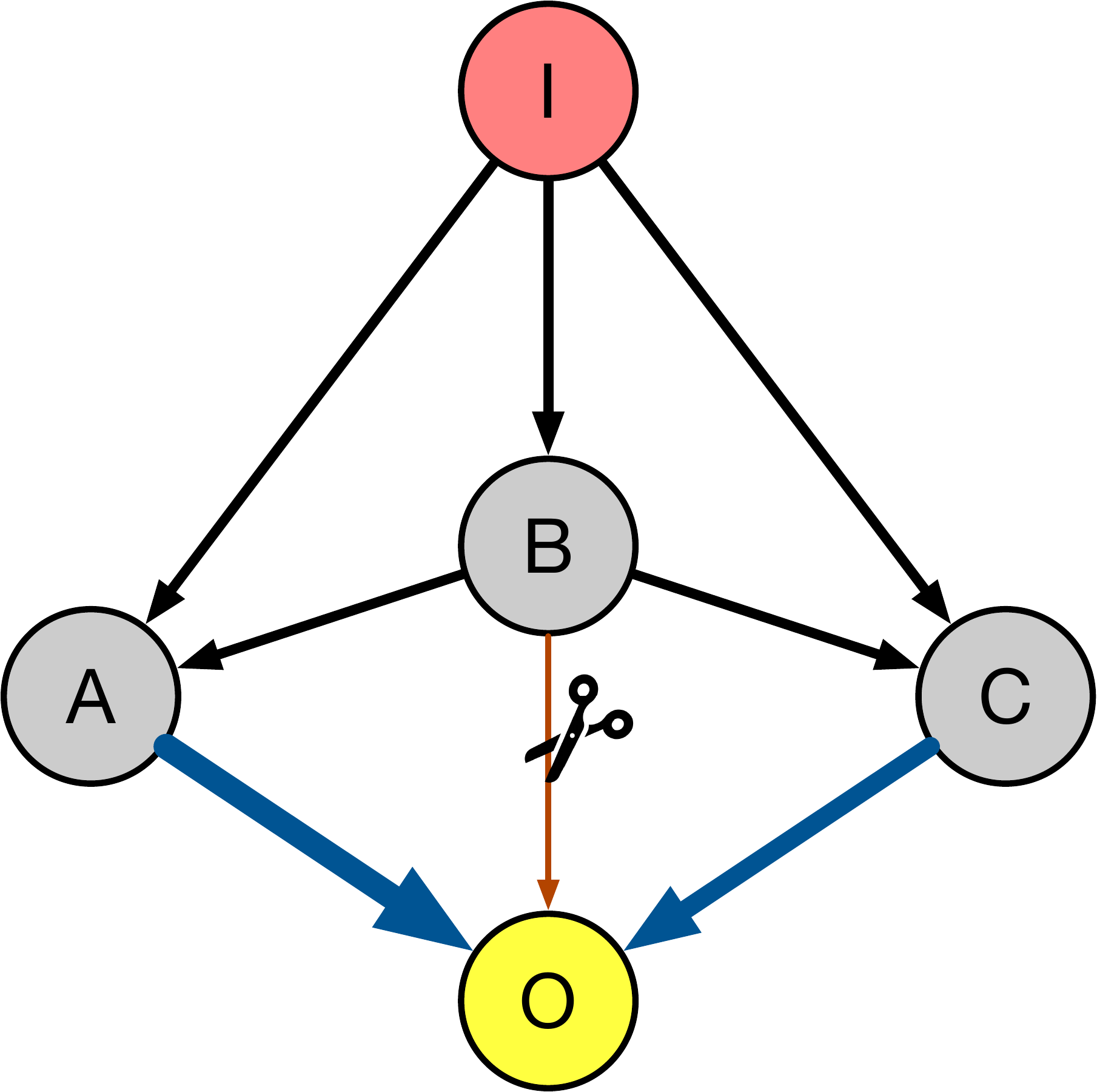}
        \caption{Step 2b: Ablate activations and measure the effect on metric $m$.}
        \label{fig:step2b}
    \end{subfigure}
    \hfill
    \begin{subfigure}[b]{0.32\textwidth}
        \centering
        \includegraphics[width=\textwidth]{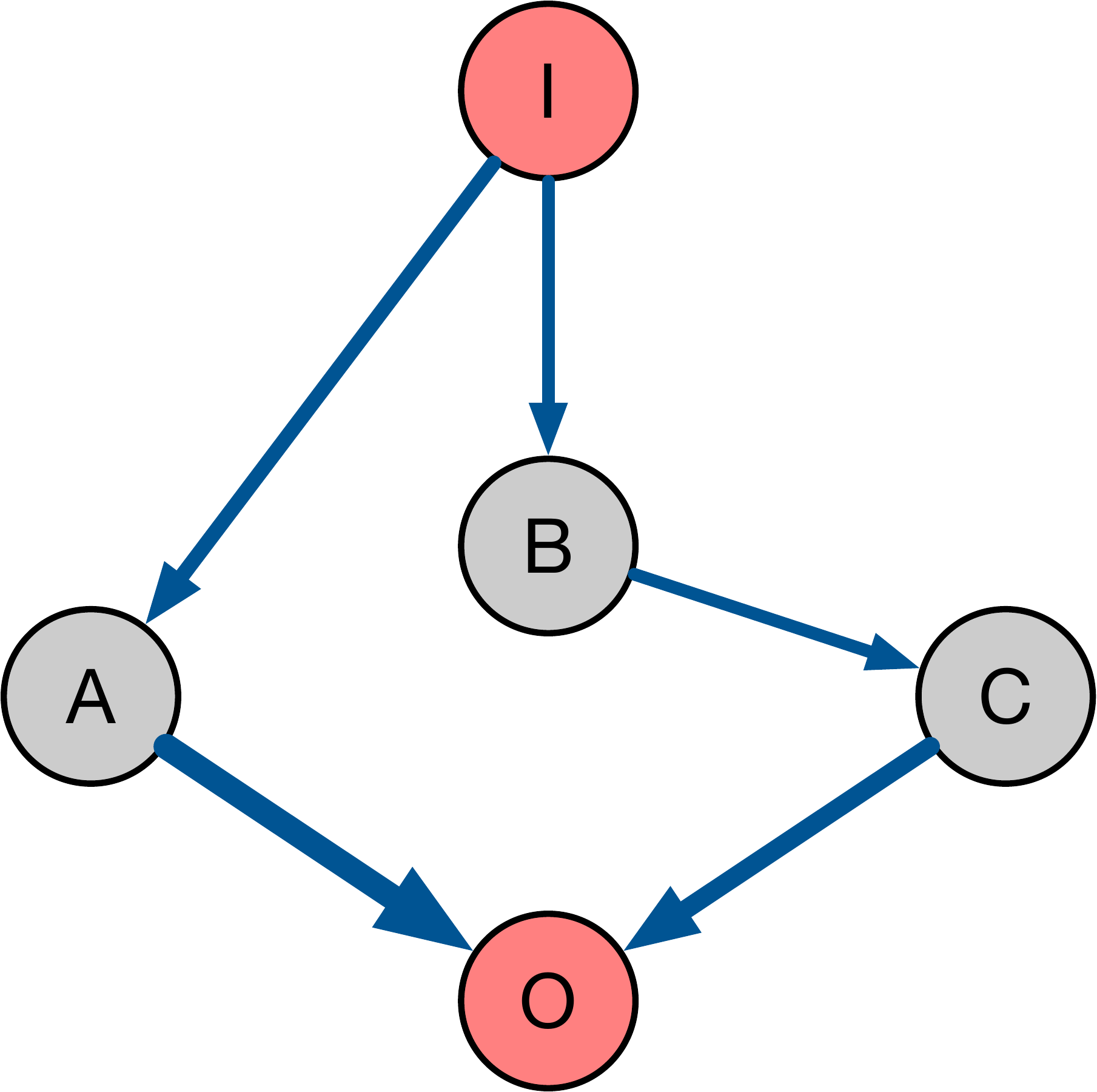}
        \caption{Step 2c: Prune connections and recursively refine the graph.}
        \label{fig:step2c}
    \end{subfigure}
    \caption{Overview of the ACDC method for refining computational graphs. Steps include specifying the graph and pruning threshold (2a), ablation and metric measurement (2b), and recursive pruning and refinement (2c), resulting in a subgraph highlighting critical components for the specified task. Figure taken from \citep{conmy2024towards}.}
    \label{fig:acdc-steps}
\end{figure}

ACDC then iterates backwards over the topologically sorted nodes in the computational graph, starting from the output and moving towards the input. During this process, it ablates activations of connections between a node and its children by replacing the activations with corrupted or zero values, measuring the impact on the output metric. The ablation is performed using a receiver hook function, which modifies the input activations to a node based on the presence or absence of edges connecting it to its parents. If the change in the metric is less than the specified threshold, the connection is pruned, updating the graph structure and altering the parent-child relationships between nodes. This is shown in Figure \ref{fig:acdc-steps}.

This pruning step is recursively applied to the remaining nodes. If a node becomes disconnected from the output node, it is removed from the graph. The resulting subgraph contains the critical components and connections necessary for the given task. An important hyperparameter in ACDC is the order in which the algorithm iterates over the parent nodes. This order, whether reverse, random, or based on their indices, significantly affects the performance in circuit identification.

\subsection{Nodes vs. edges}

ACDC operates primarily on edges rather than nodes, even though the procedure is agnostic to whether we corrupt nodes or edges in the computational graph. The reason for this is that operating on edges allows ACDC to capture the compositional nature of reasoning in transformer models, particularly in how attention heads in subsequent layers build upon the computations of previous layers.

By replacing the activation of the connection between two nodes (e.g., Layer 0 and Layer 1) while maintaining the original activations between other nodes (e.g., Layer 1 and Layer 2), ACDC can distinguish the effect of model components in different layers independently. This is crucial for understanding the role of each component in the compositionality of computation between attention heads in subsequent layers \citep{elhage2021mathematical}.

Although ACDC can split the computational graph into query, key, and value calculations for each attention head, the authors focus primarily on attention heads and MLP layers to complete their circuit identification within a reasonable amount of time. This is similar to the approach taken in our method, where we also focus on attention heads, as the canonical circuits for each task are largely defined in terms of this level of granularity.

\subsection{Final output}

The final output of an ACDC circuit prediction is a subgraph of the original computational graph, which contains the critical nodes and edges for the given task. The nodes in this subgraph represent the components specified in the original computational graph, such as attention heads, query/key/value projections, and MLP layers. The edges represent the connections between these components that are essential for the model's performance on the task.

For most of the circuits examined in the ACDC paper, including the IOI task, the authors focus on attention heads, as these have canonical ground-truths from previous works. This allows for a direct comparison between the ACDC-discovered circuits and the manually identified circuits, providing a way to validate the effectiveness of the ACDC algorithm in recovering known circuits. \textit{This means we can also provide a direct comparison to ACDC on both a node-level and edge-level}. Regardless of the approach in finding the circuit components, the final output of both methods is a predicted circuit of attention heads we can compare to the ground-truth for the relevant task.

\subsection{On why we can compare ACDC to our method}

So why do we believe it makes sense to compare our node-level and edge-level circuit discovery with ACDC's node- and edge-level discovery, when the methods of determining the importance of a node or an edge are fundamentally difference in either case? For instance, we determine the importance of an ``edge'' between two heads by examining the number of unique co-occurrence code-pairs for that pair of heads. ACDC instead ablates the activation of the connection between these two heads. However, we note that the result of both of these methods (that is, a binary classification of a head as being in the circuit or not being in the circuit) is the same. We simply group the edge-level and node-level methods together for comparison because edge-level focuses on the information moving \textit{between} nodes (via the residual stream), whereas node-level looks at the output of an individual head (to the residual stream) \textit{in isolation}.

\subsection{HISP and SP}
\label{app:acdc_hisp_sp}

Subnetwork probing \citep{cao2021low} and head importance score for pruning \citep{michel2019sixteen} are both predecessors of ACDC used to examine which transformer components are important for certain inputs, and thus which components might be part of the circuit for a specific type of task. Whilst they are not the focal comparison of our results, we include the methodology used here largely as a supplement to Figure \ref{fig:edge_node_auc}. We follow the exact same setup as \citep{conmy2024towards}, and direct the reader to the ACDC repository for implementation details \footnote{\url{https://github.com/ArthurConmy/Automatic-Circuit-Discovery}}.

\paragraph{Subnetwork probing (SP)}

To compare our circuit discovery approach with Subnetwork Probing (SP) \citep{cao2021low}, we adopt a similar setup to ACDC \citep{conmy2024towards}. SP learns a mask over the internal model components, such as attention heads and MLP layers, using an objective function that balances accuracy and sparsity. This function includes a regularisation parameter $\lambda$, which we do not refer to in the main text to avoid confusion with the sparsity penalty used in training our sparse autoencoder (SAE). Unlike the original SP method, which trains a linear probe after learning a mask for every component, we omit this step to maintain alignment with ACDC's methodology.

We made three key modifications to the original SP method. First, we adjusted the objective function to match ACDC's, using either KL divergence or a task-specific metric instead of the negative log probability loss originally used by \citet{cao2021low}. Second, we generalised the masking technique to replace activations with both zero and corrupted activations. This change reflects the more common use of corrupted activations in mechanistic interpretability and is achieved by linearly interpolating between a clean activation (when the mask weight is 1) and a corrupted activation (when the mask weight is 0), editing activations rather than model weights. Third, we employed a constant learning rate instead of the learning rate scheduling used in the original SP method.

To determine the number of edges in subgraphs identified by SP, we count the edges between pairs of unmasked nodes. For further implementation details, please refer to the ACDC repository \citep{conmy2024towards}.

\paragraph{Head importance score for pruning (HISP)}
To compare our approach with Head Importance Score for Pruning (HISP) \citep{michel2019sixteen}, we adopt the same setup as ACDC \citep{conmy2024towards}. HISP ranks attention heads based on an importance score and retains only the top $k$ heads to predict the circuit, with $k$ being a hyperparameter used to generate the ROC curve. We made two modifications to the original HISP setup. First, instead of using the derivative of a loss function, we use the derivative of a metric $F$. Second, we account for corrupted activations as well as zero activations by generalizing the interpolation factor $\xi_h$ between the clean head output (when $\xi_h = 1$) and the corrupted head output (when $\xi_h = 0$).

The importance scores for components are computed as follows:
$$
I_C := \frac{1}{n} \sum_{i=1}^n \left|\left(C(x_i) - C(x_i^{\prime})\right)^T \frac{\partial F(x_i)}{\partial C(x_i)}\right|,
$$
where $C(x_i)$ is the output of an internal component $C$ of the transformer. For zero activations, the equation is adjusted to exclude the $-C(x_i^{\prime})$ term. All scores are normalized across different layers as described by \citet{michel2019sixteen}. The number of edges in subgraphs identified by HISP is determined by counting the edges between pairs of unmasked nodes, similar to the approach used in Subnetwork Probing. For more details on the implementation, please refer to the ACDC repository \citep{conmy2024towards}.

\subsection{Edge attribution patching (EAP)}
\label{app:eap}

Edge Attribution Patching (EAP) is designed to efficiently identify relevant model components for solving specific tasks by estimating the importance of each edge in the computational graph using a linear approximation \citep{syed2023attribution}. Implemented in PyTorch, EAP computes attribution scores for all edges using only two forward passes and one backward pass. This method leverages a Taylor series expansion to approximate the change in a task-specific metric, such as logit difference or probability difference, after corrupting an edge. For the IOI and Greater-than tasks, EAP used edge-based attribution patching with absolute value attribution. Both tasks employed the negative absolute metric for computing attributions. EAP pruned nodes using a single iteration, with the pruning mode set to ``edge''. This approach avoids issues with zero gradients in KL divergence by using task-specific metrics, making it a robust and scalable solution for mechanistic interpretability. We adapted the code from \citet{syed2023attribution}'s original paper, available \href{https://github.com/Aaquib111/edge-attribution-patching}{here}.

We noted above that EAP is limited in the metrics we can apply for discovery because the gradient of the metric cannot be zero; we elaborate here. For instance, this means we cannot use the KL divergence metric to find importance components. The KL divergence is equal to 0 when comparing a clean model to a clean model (i.e. without ablations) and is non-negative, so the zero point is a global minimum and all gradients are zero here.

\subsection{Head activation norm difference}
\label{app:head_activation_norm_difference}

The effectiveness of using SAE-learned features for identifying circuit components raises an important question: why is it necessary to project raw head activations into the SAE latent space to distinguish between positive and negative circuit computations? To investigate whether this projection aids in reducing noise or interference, we analysed the mean per-head activation averaged across positive and negative examples and computed the difference. We then calculated the norm of this difference for each head, applied a softmax function over all heads, and evaluated the ROC AUC against the ground-truth circuit. This analysis was conducted using the same number of examples (10) that the SAE was trained on.

\begin{figure}[h]
    \centering
    \includegraphics[width=0.75\textwidth]{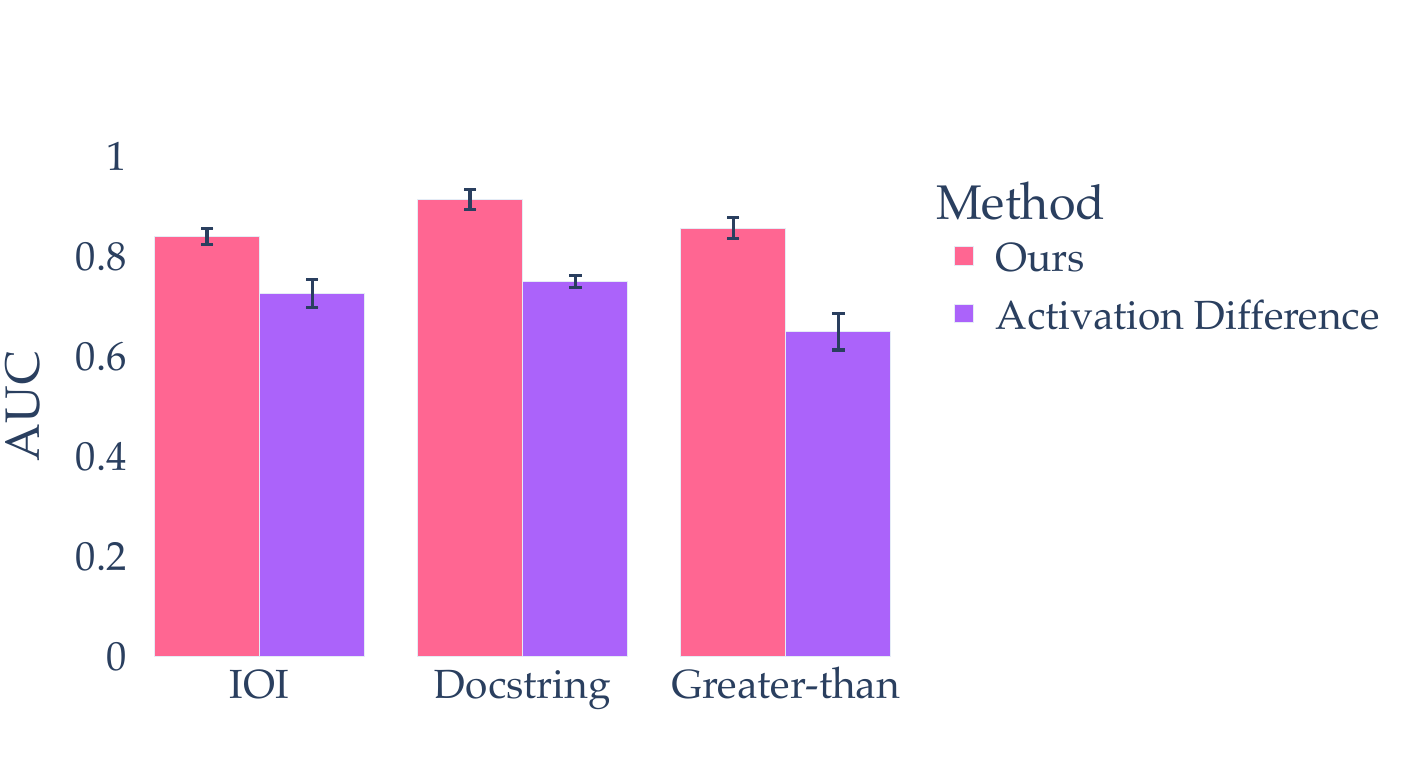}
    \caption{ROC AUC of using the norms of the differences in head activations between 10 positive and negative examples, compared with our method. While there is some signal in the raw head activations, averaging across examples does not capture as much nuance as the SAE features.}
    \label{fig:difference_norm}
\end{figure}

\begin{figure}[h]
    \centering
    \includegraphics[width=0.85\textwidth]{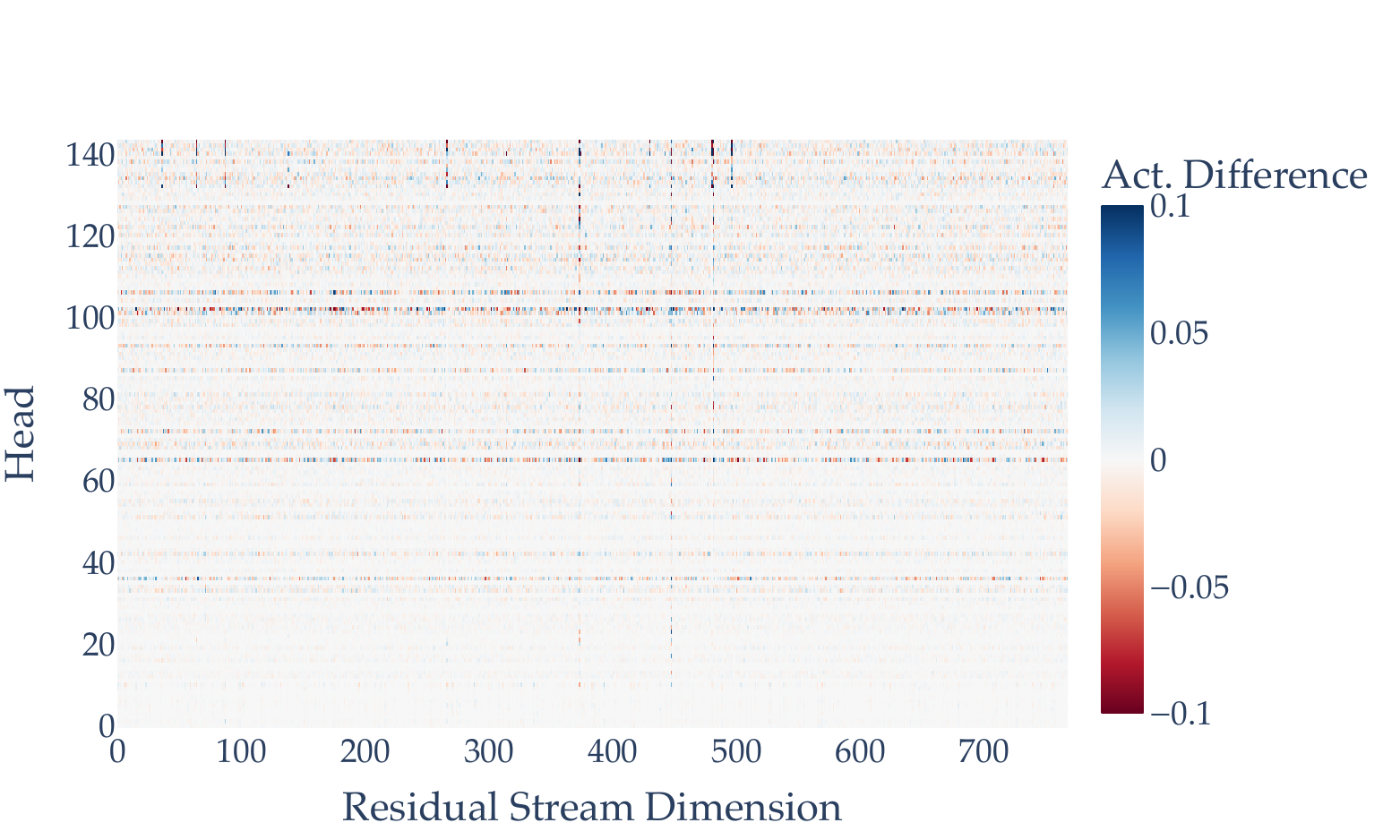}
    \caption{Mean activation difference between positive and negative examples for the IOI task. Some heads exhibit consistently high differences (horizontal stripes), and certain residual stream dimensions show consistently high differences (vertical stripes).}
    \label{fig:ioi_difference_activations}
\end{figure}

As shown in Figure \ref{fig:difference_norm}, head activations do contain some signal regarding the heads involved in circuit-specific computation. However, they are not as effective as our method in distinguishing these computations. This may be due to the variation in particular dimensions within the residual stream across all heads (corresponding to the vertical stripes in Figure \ref{fig:ioi_difference_activations}), which likely requires non-linear computation to disentangle positive and negative examples, a task the SAE likely performs effectively.

We observed that performance for the IOI and Greater-than tasks improved as the number of activations over which we computed the mean difference increased. Specifically, both tasks reached a ROC AUC of approximately 0.80-0.85 around 500 examples. However, for the docstring task, performance actually worsened with an increased number of examples. This suggests that while the mean difference method can serve as an initial sanity check, it lacks robustness for reliable circuit identification.

\section{How does the formulation of positive and negative examples affect performance?}

\subsection{Alternative negative examples for the Greater-than task}
\label{app:alt_negative_type}

The choice of negative examples is a crucial factor in the performance of the circuit identification method. In this study, we selected negative examples that were semantically similar to the positive examples but corrupted enough to prevent the model from using the current circuit to generate a correct answer.

\begin{figure}
    \centering
    \includegraphics[width=0.8\textwidth]{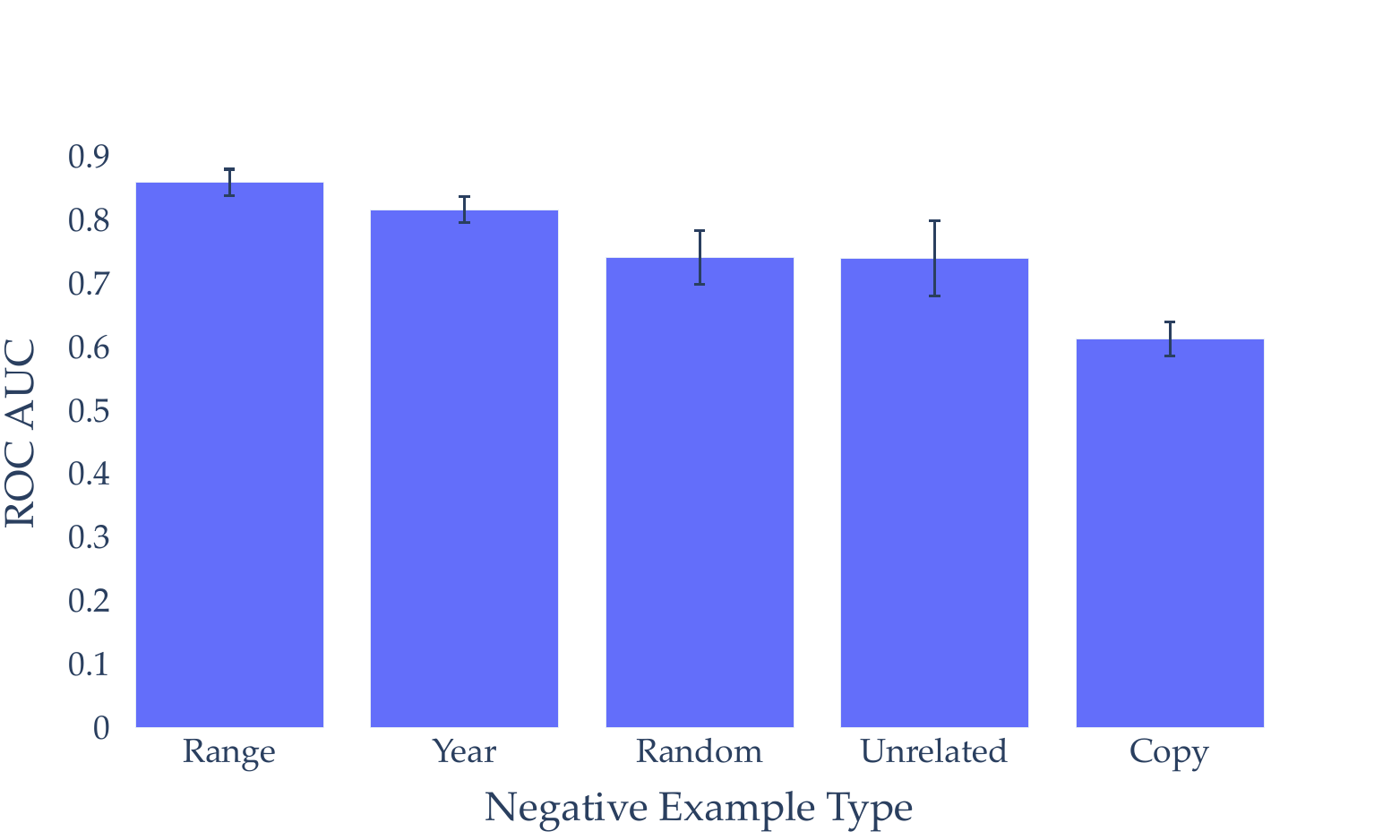}
    \caption{Node-level circuit identification performance with different types of negative examples on the Greater-than task. We train 5 SAEs for each type with different random seeds and record the ROC AUC for each. }
    \label{fig:easy_negative_types}
\end{figure}

To investigate the sensitivity of our method to the choice of negative examples, we conducted experiments with five different types of negative examples for the greater-than task:
\begin{enumerate}
\item \textcolor{lightblue}{\texttt{Range}}: The completion year starts with the preceding century. For example, ``The competition lasted from the year 1523 to the year 14''. These are the negative examples used throughout this paper.
\item \textcolor{lightblue}{\texttt{Year}}: The original negative examples from the previous paper, where the year starts with ``01''. For example, ``The competition lasted from the year 1501 to the year 15''.
\item \textcolor{lightblue}{\texttt{Random}}: The numeric completion of the century is replaced with random uppercase letters. For example, ``The competition lasted from the year 19AB to the year 19''.
\item \textcolor{lightblue}{\texttt{Unrelated}}: Examples unrelated to the task, similar to the easy negatives, in the form of ``I've got a lovely bunch of <NOUN>''.
\item \textcolor{lightblue}{\texttt{Copy}}: Negative examples with the same form as the positive examples but with different randomized years and centuries.
\end{enumerate}
The results of these experiments are presented in Figure \ref{fig:easy_negative_types}. The findings clearly demonstrate that our heuristic of selecting semantically similar examples that switch off the circuit is an effective approach to maximising performance. This is evident from the drop in performance when using \texttt{Year} types compared to \texttt{Range}. When using \texttt{Year} types, the circuit likely remains active when detecting the need to find a two-digit completion greater than ``01". In contrast, the \texttt{Range} type makes the negative examples nonsensical by setting the completion century in the past, which likely switches off the circuit.

Interestingly, \texttt{Unrelated} negative examples lead to a considerable drop in performance, which we explore further below.

\subsection{Including ``easy negatives'' in the training data}
Various studies suggest that hard negative samples, which have different labels from the anchor samples (in this case, our positive examples) but with very similar embedding features, allow contrastive-loss trained autoencoders to learn better representations to distinguish between the two \citep{shen2020simple}. However, in our case, our negative samples are specifically designed to all be hard negatives.

Currently, there is no reason to believe that the most important codes for differentiating between positive and negative examples should capture all the codes in the IOI task. This is because the IOI negative examples are actually \textit{almost} positive examples. For instance, we would expect previous token heads to be exactly the same in both the negative and positive examples (since both involve two names at least). So, we actually need to give the model data such that some of the codes are forced to be assigned to non-IOI related behaviour. This will hopefully make the remaining codes more relevant for finding the right attention heads in the right layer. This suggests that we should include some non-IOI related data, such as samples from the Pile dataset, in the training data.

We experimented with whether inclusion of ``easy negatives'', defined as random pieces of text sampled from the Pile \citep{gao2020pile}, would allow the autoencoder to produce representations that were better for us to pick out the important model components for implementing the task. For example, if the positive samples and hard negative samples shared heads for the IOI task, such as detecting names, we would not identify those heads as important because importance is defined by whether the discrete representation helps distinguish a positive sample from a negative one. Thus, including easy negatives could make those particular heads important.

\begin{figure}
\centering
\includegraphics[width=0.85\textwidth]{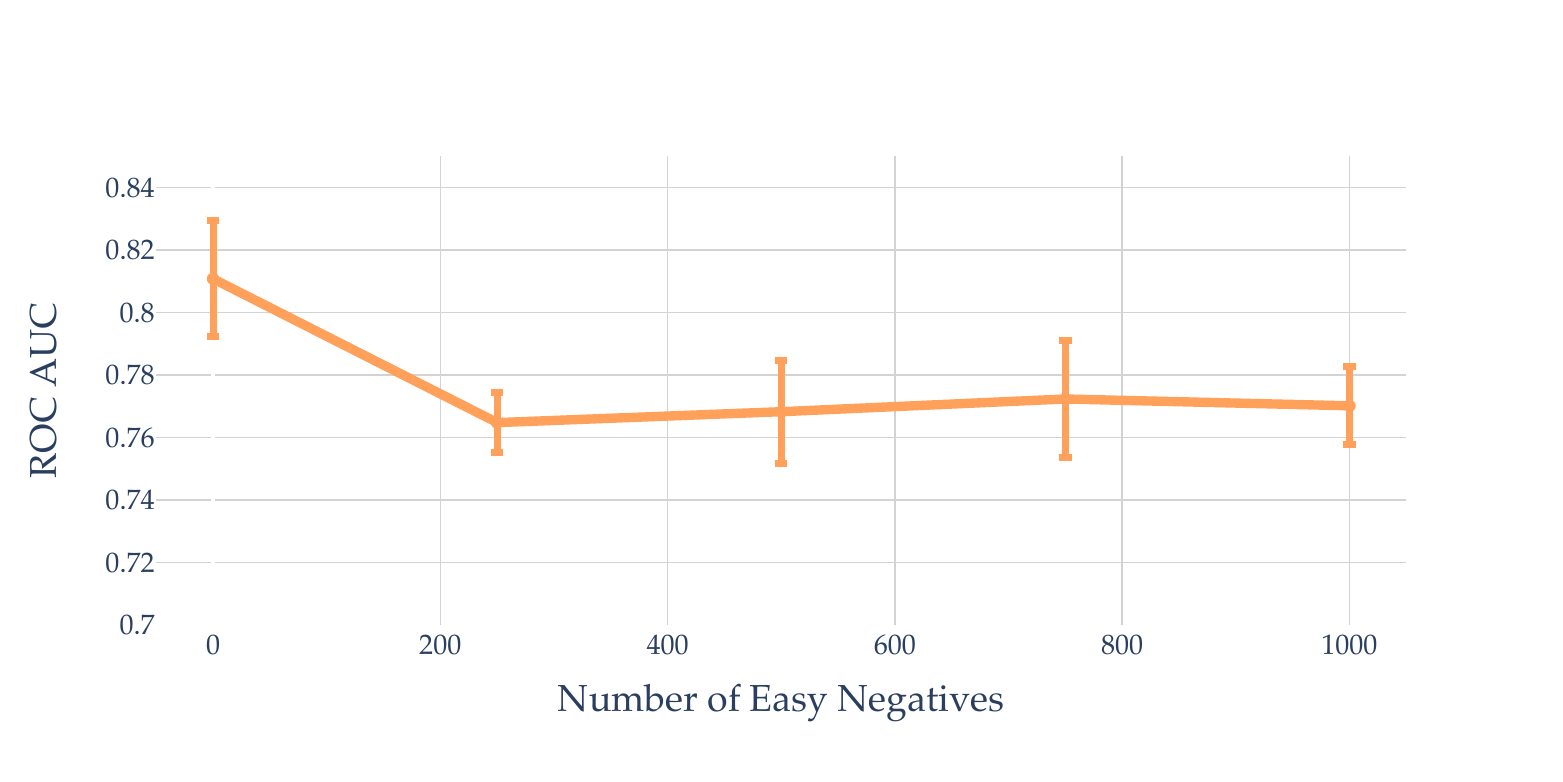}
\caption{Number of easy negatives included in training data for the sparse autoencoder (in addition to the 250 positive and 250 negative examples) and the ROC AUC of the resulting node-level detection. Error bars are shown for 5 training runs at each data point.}
\label{fig:roc_auc_vs_num_easy_negatives}
\end{figure}

However, as seen in Figure \ref{fig:roc_auc_vs_num_easy_negatives}, inclusion of easy negatives actually leads to a decrease in performance on the IOI task. It's possible that the model is forced to assign codes to expressing concepts and behaviours unrelated to the IOI task, and thus cannot as meaningfully distinguish between the semantically-similar positive and negative examples.

\section{Normalisation and design choices}
\label{app:normalisation_choices}

\subsection{Softmax across head or layer} 
A key design choice is whether to take the softmax across the vector of individual head counts or whether to take it across individual layers; that is, first reshape the vector into a matrix of shape $(n_\text{layers} \times n_\text{heads})$. A valid concern is that taking the softmax across layers will make unimportant heads seem important. For instance, if there is a layer with a head that has 1 unique positive code, and all other heads in that layer have 0, this head will have a value of 1.0 and thus be selected no matter what the threshold is. However, it possible that the law of large numbers will cause the number of unique positive codes to be approximately uniform in unimportant layers, so this may not be an issue.

\begin{figure}[htbp]
    \centering
    \begin{minipage}[b]{0.49\textwidth}  
        \centering
        \includegraphics[width=\textwidth]{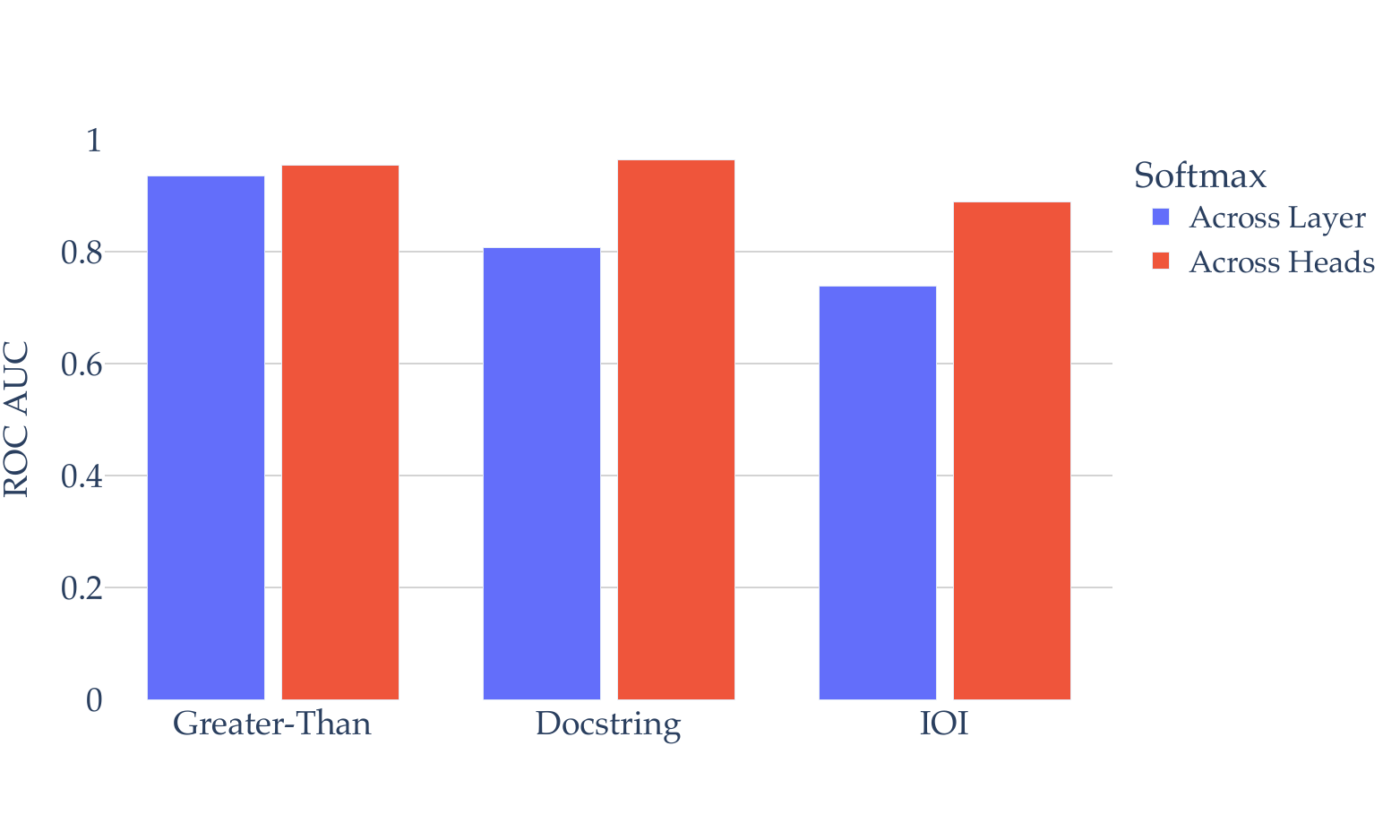}
        \caption{Effect on node-level ROC AUC when applying softmax across individual layers compared with the heads as a single vector. Softmax across heads performs better in all three datasets, with large improvements in IOI and Docstring.}
        \label{fig:softmax_layer_head}
    \end{minipage}
    \hfill  
    \begin{minipage}[b]{0.49\textwidth}  
        \centering
        \includegraphics[width=\textwidth]{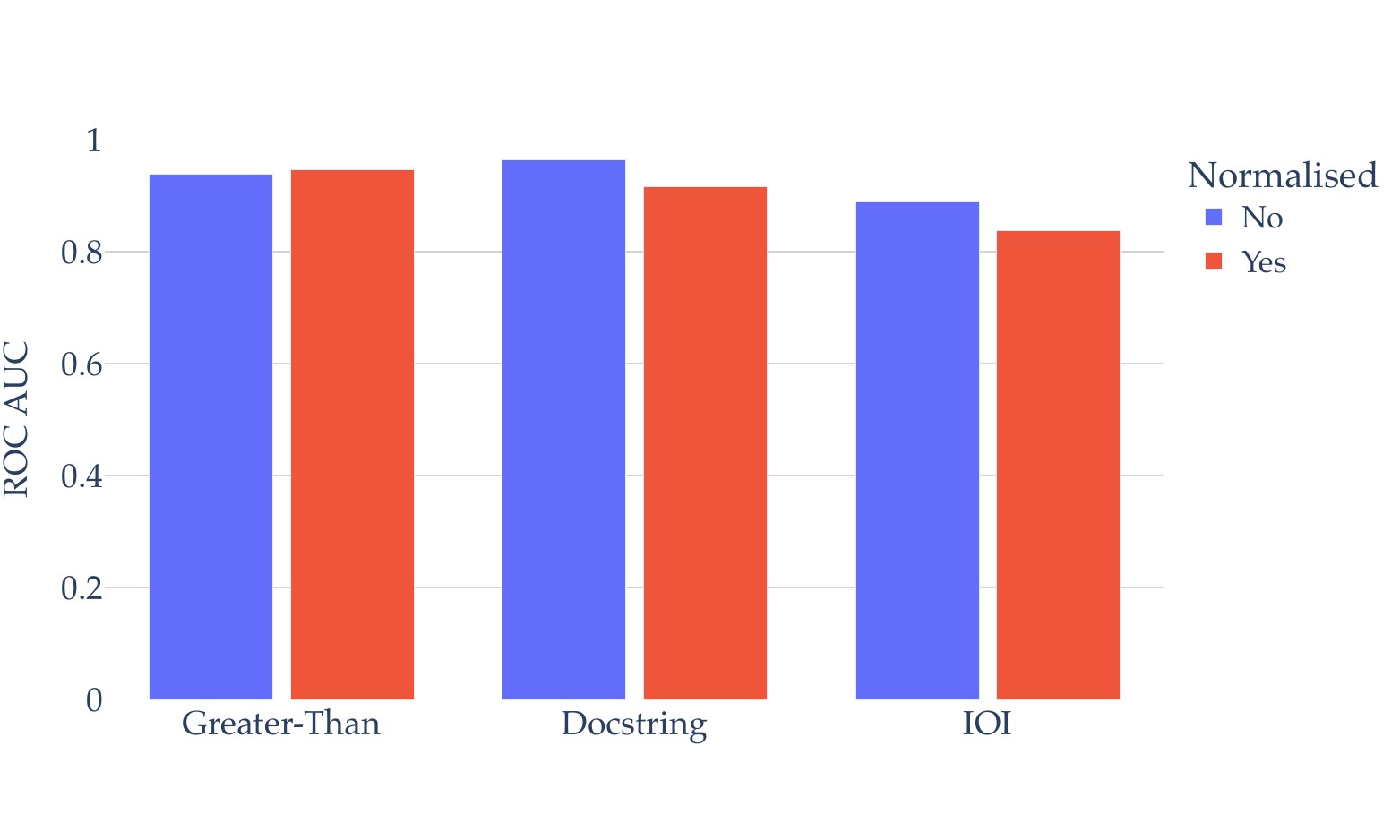}
        \caption{Effect on node-level ROC AUC when normalising the number of positive codes per head (before softmax) by dividing by the overall number of unique codes in that head (positive and negative examples).}
        \label{fig:norm}
    \end{minipage}
\end{figure}

We show the effects of taking the softmax across layer and across individual heads on the node-level ROC AUC in Figure \ref{fig:softmax_layer_head}. Softmax across heads performs best in all three tasks, with significant improvement compared with softmax across layers in the IOI and Docstring tasks.

\subsection{Normalising unique positive codes by overall number of unique codes}

We also hypothesised that we may be able to improve performance by normalising by the overall number of unique codes per head. The reasoning is as follows: if a head has a large number of unique positive codes, our current method is likely to include it in the circuit. However, if the head also has a large number of unique negative codes, then clearly it has a large range of outputs and the autoencoder deemed it necessary to assign many codes to this head, \textit{regardless of whether the example is positive or negative}. In essence, the head is important regardless of if we're in the IOI task or not; including the head in our circuit prediction may be erroneous. Normalising by the overall number of unique codes should correct this.

As shown in Figure \ref{fig:norm}, normalising seems to have a relatively minor effect. It decreases node-level ROC AUC in the Docstring and IOI tasks, and slightly increases performance in the Greater-than task.

\subsection{Number of examples used}

\begin{figure}[ht]
    \centering
    \begin{minipage}{0.48\textwidth}
        \centering
        \includegraphics[width=\textwidth, trim=0 0 0 50pt, clip]{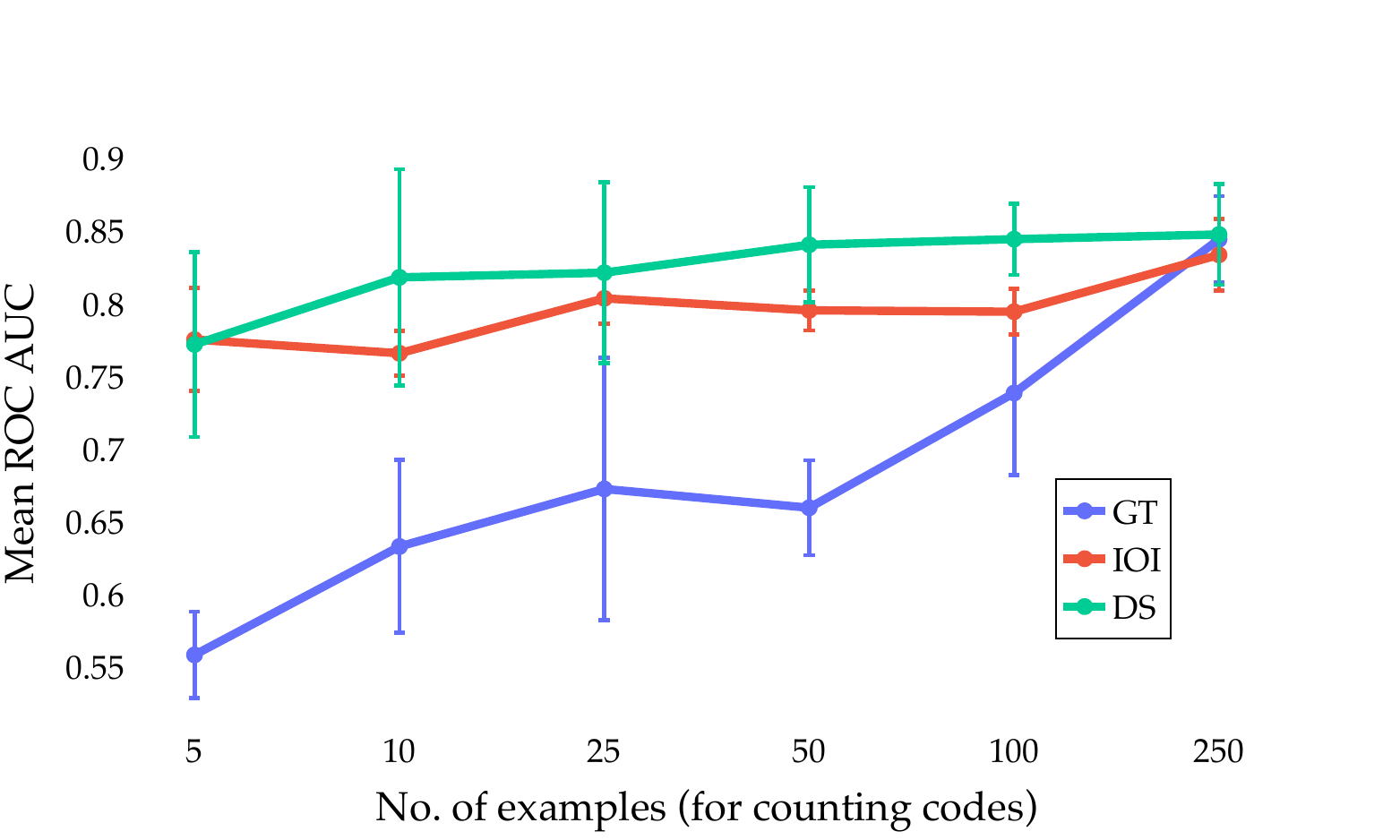}
        \caption{Node-level ROC AUC for varying number of positive and negative examples (each) used for counting the number of unique positive codes per head. Each SAE for each datapoint was trained on 10 examples only. Increasing the number of examples only significantly affects the ROC AUC of the Greater-than task.}
        \label{fig:roc_auc_vs_examples_counting}
    \end{minipage}\hfill
    \begin{minipage}{0.48\textwidth}
        \centering
        \includegraphics[width=\textwidth, trim=0 0 0 50pt, clip]{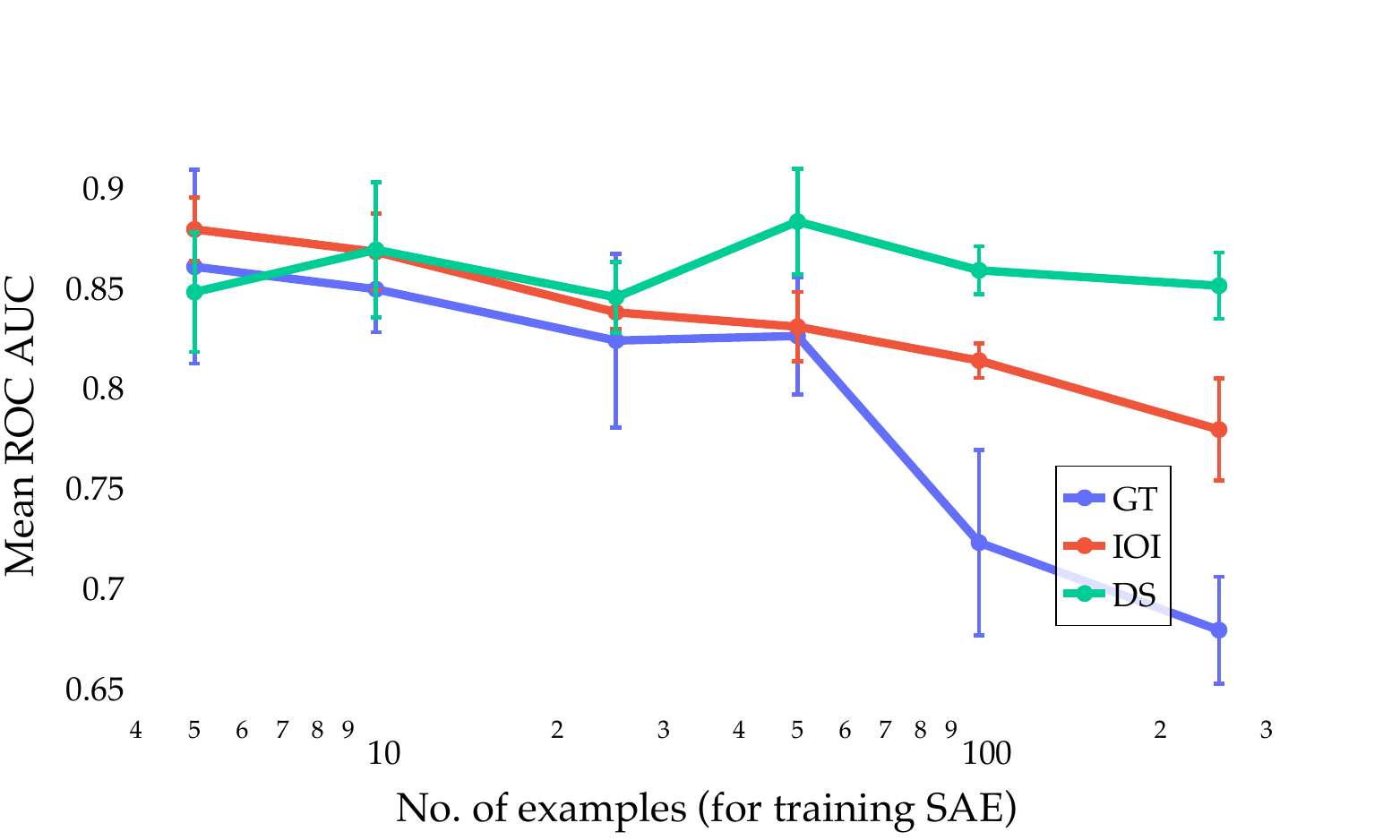}
        \caption{Node-level ROC AUC compared with the number of examples the SAE was trained on. There is a clear decline in performance for the IOI and Greater-Than tasks (where we use GPT-2) as we \textit{increase} the number of examples. Maximum performance for all examples is achieved at around 5-10 examples.}
        \label{fig:roc_auc_vs_examples}
    \end{minipage}
\end{figure}

Finally, we show in Figure \ref{fig:roc_auc_vs_examples_counting} how node-level performance varies with the number of positive and negative examples used to calculate the head importance score (i.e. the softmaxed number of unique positive codes per head). We find that we can use as little as 10 examples for the IOI and Docstring tasks, but require the full 250 positive and 250 negative for the Greater-than task. We suspect this is to do with the numerical nature of the Greater-than task and the fact that there are 100 two digit numbers specifying appropriate completions, and so a larger sample size may be required to represent what each of these different attention head outputs look like.

Additionally, we note that the number of examples the SAE requires during training to learn robust representations is actually only about 5-10. Figure \ref{fig:roc_auc_vs_examples} shows that node-level performance actually \textit{decreases} for IOI and Greater-than (both GPT-2 tasks) as we increase the number of training examples for the sparse autoencoder. Whilst the Docstring task does not see a decrease as we increase the training example set, it still achieves near-maximal performance around 10 examples. For Docstring and IOI, we can also achieve near-maximal performance with just 10 examples for both steps (training the SAE and counting unique positive codes). However, Greater-than requires a significant number of examples for the latter step.

\section{Further comparisons to previous methods}
\label{app:further_comparisons}

In an extension to Figure \ref{fig:edge_node_auc}, we record the actual values for each method with both random and zero ablations in Table \ref{tab:combined_ablation_ours}. This allows us to compare previous methods with ours, with both the optimal hyperparameters for each dataset, and set hyperparameters across datasets. 

\begin{table}[htbp]
\caption{ROC AUCs for circuit-identification in three tasks for GPT-2 Small. Previous methods are shown with both corrupted activations with zero and random ablations in the form \textit{Random} / \textit{Zero}, for both node and edge-level circuit identification, alongside our method. The \textit{Ours} column are the ROC AUCs for a hyperparameter sweep across each individual model. The \textit{Ours (set params)} column shows the results when we set the autoencoder's learned features to 200, $\lambda$ to 0.02, and the threshold for $k$ in edge-selection to be a quarter of the total number of co-occurrences. We bold our results if they exceed the AUC of every other method with both random and zero ablation. The results for previous methods (ACDC, HISP and SP) use logit difference, which is most comparable to our method of only using a single label to assign a difference between positive and negative examples (as opposed to KL-divergence over all token probabilities). Results come from \citet{conmy2024towards}, with the addition of our own results.}
\centering
\begin{tabular}{@{}lccccc@{}}
\toprule
Task & ACDC & HISP & SP & \color{blue}\textit{Ours}\color{black} & \color{blue}\textit{Ours (set params)}\color{black} \\
\midrule
\textit{Node-level} & & & & & \\
Docstring & 0.938 / 0.825 & 0.889 / 0.889 & 0.941 / 0.398 & \color{blue}\textbf{0.945}\color{black} & \color{blue}0.915 $(\pm 0.014)$ \color{black} \\
Greater-than & 0.766 / 0.783 & 0.631 / 0.631 & 0.811 / 0.522 & \color{blue}\textbf{0.821}\color{black} & \color{blue}\textbf{0.832 $(\pm 0.058)$}\color{black} \\
IOI & 0.777 / 0.424 & 0.728 / 0.728 & 0.797 / 0.479 & \color{blue}\textbf{0.854}\color{black} & \color{blue}\textbf{0.853 $(\pm 0.016)$}\color{black} \\
\midrule
\textit{Edge-level} & & & & & \\
Docstring & 0.972 / 0.929 & 0.821 / 0.821 & 0.942 / 0.482 & \color{blue}\textbf{0.974}\color{black} & \color{blue}0.914 $(\pm 0.020)$\color{black} \\
Greater-than & 0.461 / 0.491 & 0.706 / 0.706 & 0.812 / 0.639 & \color{blue}\textbf{0.963}\color{black} & \color{blue}\textbf{0.856 $(\pm 0.021)$} \color{black} \\
IOI & 0.589 / 0.447 & 0.836 / 0.836 & 0.707 / 0.393 & \color{blue}\textbf{0.863}\color{black} & \color{blue}\textbf{0.840} $(\pm 0.016)$\color{black} \\
\bottomrule
\end{tabular}
\label{tab:combined_ablation_ours}
\end{table}

\section{\texttt{tracr}-tasks}
\label{app:tracr}

\texttt{tracr} is a compiler that converts human-readable programs into transformer weights \citep{weiss2021thinking, lindner2024tracr}. This allows us to automatically determine the attention heads responsible for implementing a certain behaviour, as we have access to the underlying assignation of components to layers. 

\subsection{Compiled models}

We compile \texttt{tracr} transformers for four different tasks: reversing a list (\texttt{tracr-reverse}), counting the fraction of previous tokens in a position equal to a certain token (\texttt{tracr-fracprev}), sorting a list (\texttt{tracr-sort}), and sorting a list by the frequency of the tokens in the list (\texttt{tracr-sortfreq}). All code required to compile these models is available in the \href{https://github.com/google-deepmind/tracr/blob/main/tracr/examples/Visualize_Tracr_Models.ipynb}{Tracr repository}. Note that all of these transformers output a vector of tokens rather than a single token, and each of the compiled transformers has a maximum sequence length, which we set to 6 for all examples.

We then simulate our circuit-discovery methodology as follows. For each vocabulary of inputs to each task, we generate 250 permutations with replacement for our positive examples. Because \texttt{tracr} transformers have compiled weights that only implement a single task, there is no way to ``turn off'' a circuit with negative examples. Due to this, we corrupt the residual stream directly to create our negative examples. To do this for each example, we add Gaussian noise to attention head vectors $q, k$ or $v$ if the respective actual weight matrices $Q$, $K$ or $V$ in the transformer contain all zeros; simultaneously, we zero out the attention head vectors $q, k$ and $v$ if the respective weight matrices $Q, K$ or $V$ contain a non-zero element. We define an attention head component ($Q$, $K$ or $V$) as being in the ground-truth circuit if it makes a non-trivial write the to residual stream (i.e. the output to the residual stream is non-zero). Finally, we train our SAE on 10 randomly sampled positive and negative examples and use the full 500 examples to calculate the number of unique positive codes.

\begin{figure}
    \centering
    \begin{subfigure}[b]{0.49\textwidth}
        \includegraphics[width=\textwidth]{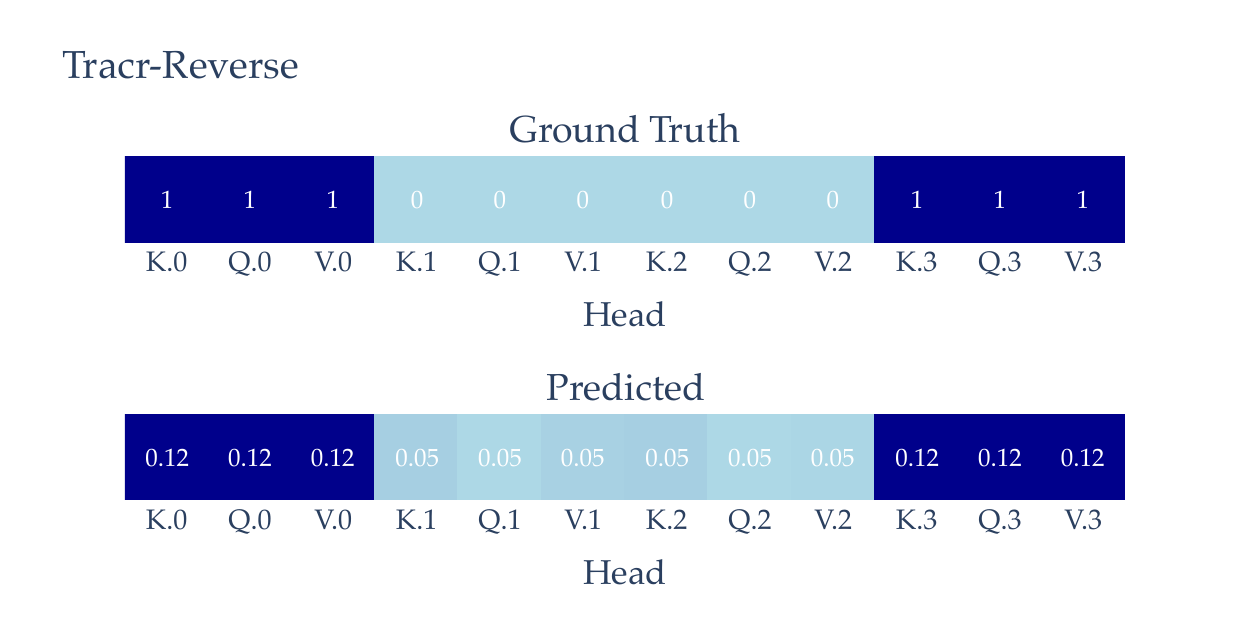}
    \end{subfigure}
    ~ 
    \begin{subfigure}[b]{0.49\textwidth}
        \includegraphics[width=\textwidth]{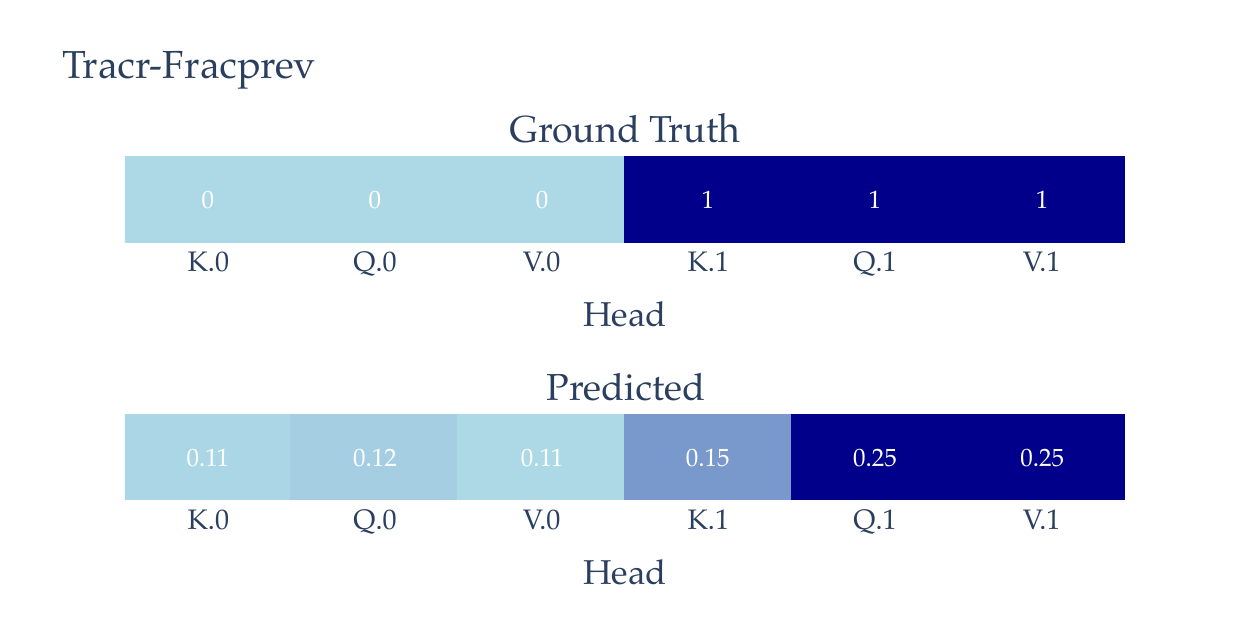}
    \end{subfigure}
    \\
    \begin{subfigure}[b]{0.49\textwidth}
        \includegraphics[width=\textwidth]{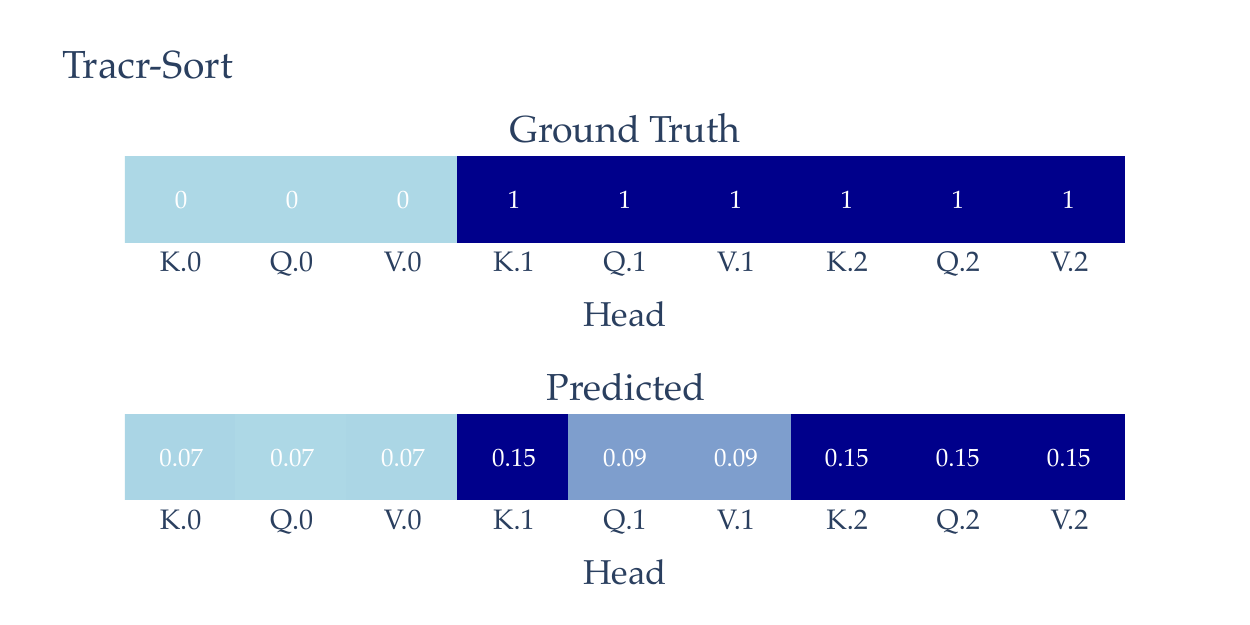}
    \end{subfigure}
    ~ 
    \begin{subfigure}[b]{0.49\textwidth}
        \includegraphics[width=\textwidth]{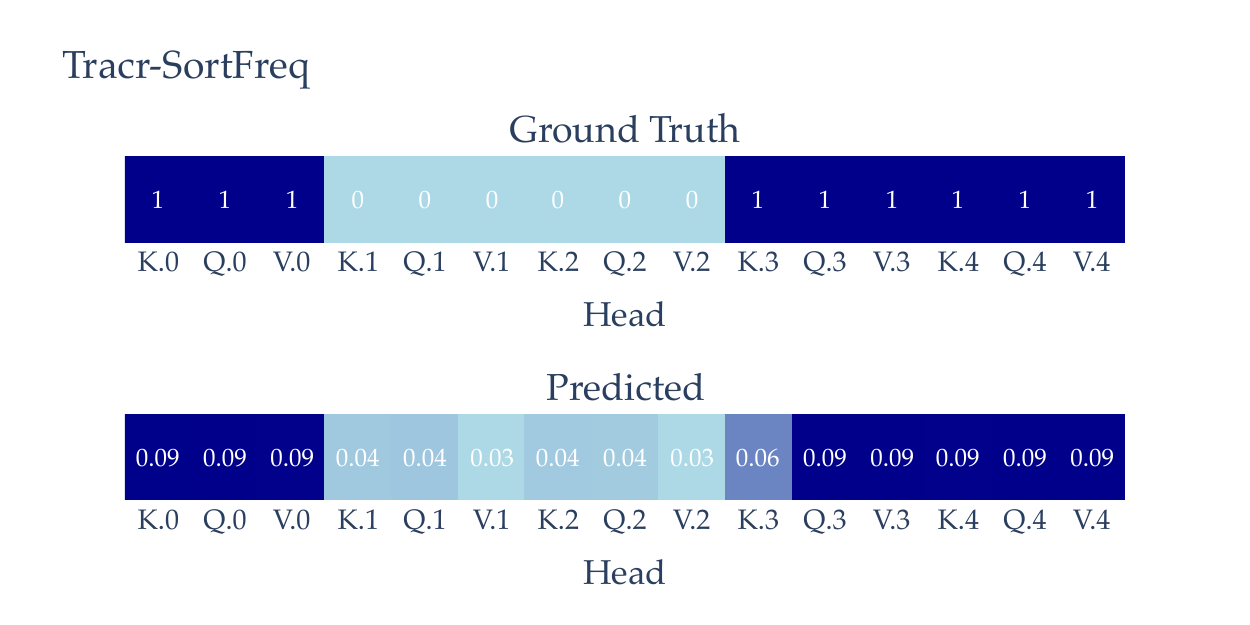}
    \end{subfigure}
    \caption{The ground-truth circuits for attention-head vectors $K$, $Q$ and $V$ from \texttt{tracr}-compiled transformers, alongside the softmaxed number of unique positive codes from our SAE representations. For each of the four tasks, the softmaxed values achieve an ROC AUC of 1.0 across thresholds.}
    \label{fig:tracr_models}
\end{figure}

Our method achieves a perfect ROC AUC of 1.0 on all \texttt{tracr}-tasks (Figure \ref{fig:tracr_models}). Whilst we should expect good performance on what amounts to a relatively simple task (since the residual streams are corrupted directly), this does provide further evidence of the mechanism that makes our approach successful. Clearly, circuit identification really does boil down to heads that are active on positive examples but inactive on negative. It appears that the SAE then \textit{assigns a high number of codes to active attention heads}. Active heads are then obviously more likely to be involved in the circuit. We are currently investigating how to collect further evidence in support of this hypothesis.

\section{Induction circuit}


Another ``circuit'' commonly studied in the literature is \textit{induction}, which is implemented by heads that complete token sequences in the form \texttt{[A][B] ... [A] -> [B]} \citep{olsson2022context}. We include induction here as a study in how our method may be evaluated without a ground-truth circuit, and how these evaluations may demonstrate model agnosticity.

We generate our positive and negative examples (25 of each) for the task by randomly sampling examples of 10 integers, representing tokens, and then repeating this pattern of 10 integers for each example for the positive examples, and generating 10 more non-repeating tokens for the negative examples. The process for identifying the circuit is then exactly the same as above. We use 10 tokenised examples for training the SAE. 

\begin{figure}[ht]
    \centering
    \begin{subfigure}[b]{0.49\textwidth}
        \includegraphics[width=\textwidth]{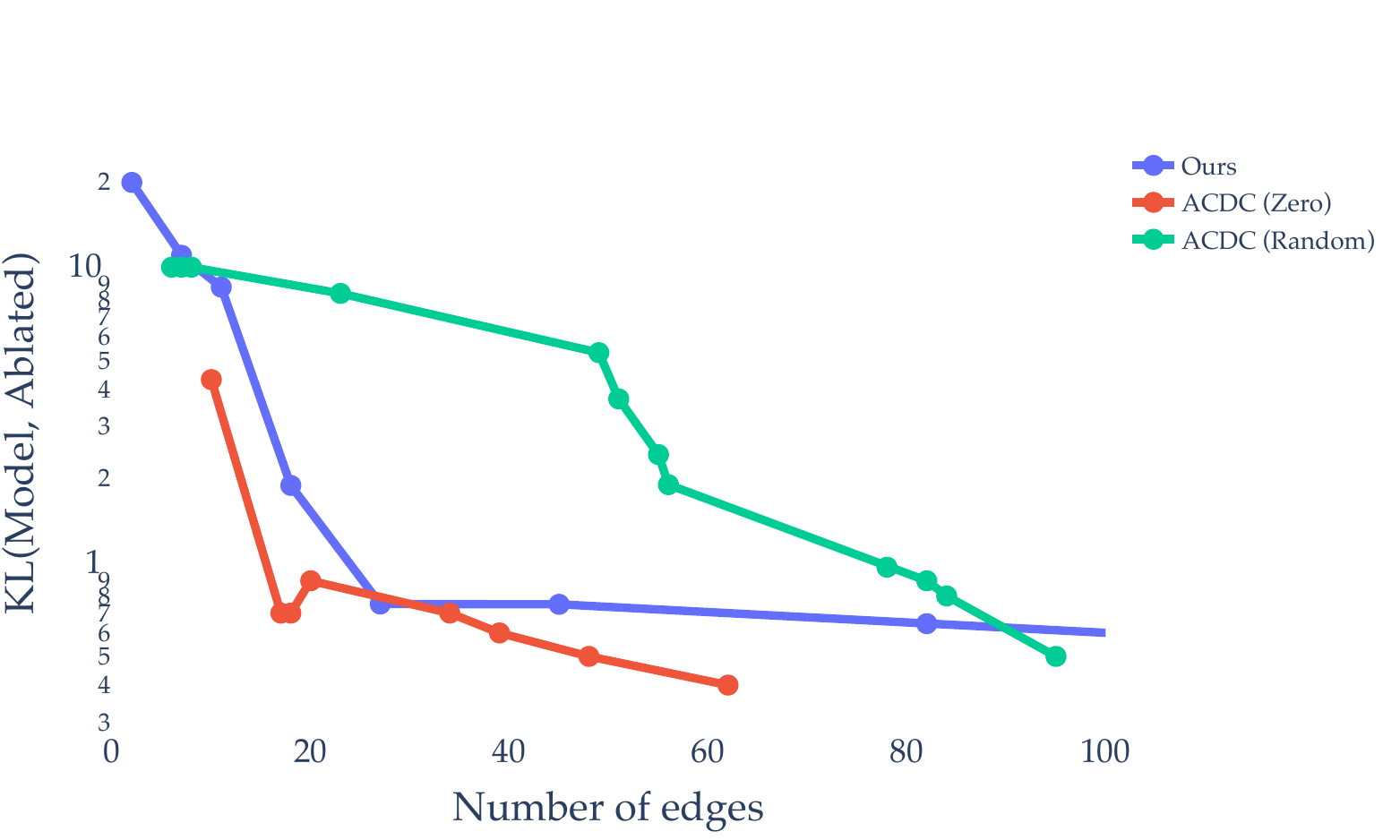}
        \caption{Edge-level circuit identification}
        \label{fig:edge_node_kl_induction}
    \end{subfigure}
    \hfill 
    \begin{subfigure}[b]{0.49\textwidth}
        \includegraphics[width=\textwidth]{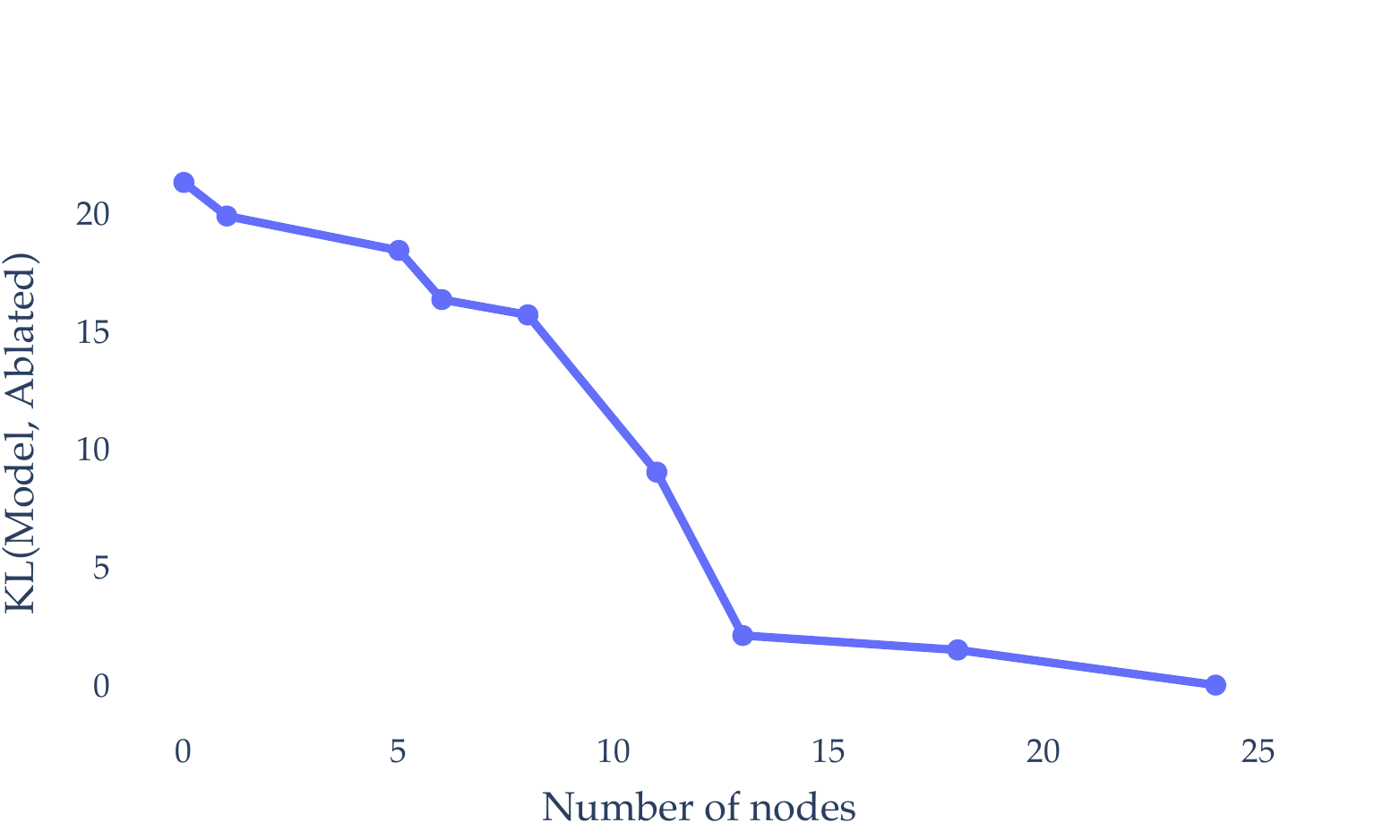}
        \caption{Node-level circuit identification}
        \label{fig:kl_nodes}
    \end{subfigure}
    \caption{KL divergence between the clean model and the circuit (with the remainder of the model ablated) for various thresholds: $\theta$ for ours, and $\tau$ for ACDC.}
    \label{fig:combined_figures}
\end{figure}

Our initial induction experiment was performed on a pre-trained 2-layer, 8-head attention only transformer from \citet{goldowsky2023localizing}. We compared how the KL divergence between the clean model and our circuit (with all other components ablated to zero) changed as we altered our threshold $\theta$. For edge-level circuit identification, we compared the same KL on the same model with ACDC, with both zero and random ablation. Figure \ref{fig:edge_node_kl_induction} shows that the KL divergence drops at approximately the same rate for our method as that of ACDC. This is promising and perhaps surprising, as our method is not set up to minimise KL divergence between clean model and circuit, whereas ACDC is. For node-level, we reach near-zero KL divergence around 12-15 nodes (i.e. 12-15 attention heads), showing our method is good at not identifying unimportant heads in the circuit (Figure \ref{fig:kl_nodes}).

We also experimented with how our method performed in terms of faithfulness across different language models, this time using the mean loss on the repeated tokens as our metric. We repeated the above process for node and edge-level circuit identification on GPT2-Small, Pythia-160M \citep{biderman2023pythia}, and Opt-125M \citep{zhang2022opt}. It appears that, at least for the induction task, recovery of performance occurs at the same rate across different models. Note that we can directly compare the number of nodes and edges across these models here as they each have 12 layers with 12 attention heads in each layer.

\begin{figure}[htbp]
    \centering
    \begin{subfigure}{0.49\textwidth}  
        \includegraphics[width=\textwidth]{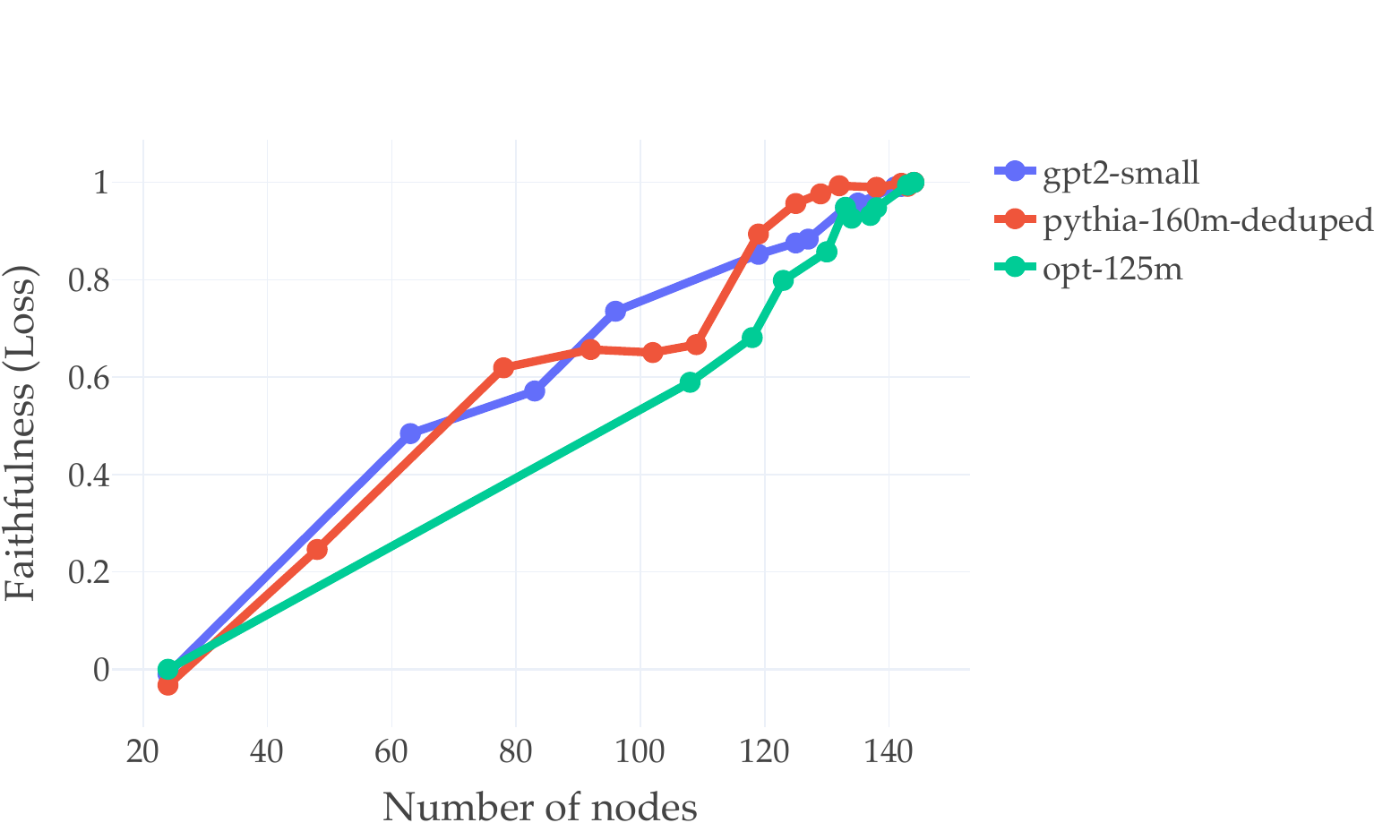}
        \caption{Node faithfulness}
        \label{fig:faithfulness_node_all_models_induction}
    \end{subfigure}
    \hfill  
    \begin{subfigure}{0.49\textwidth}
        \includegraphics[width=\textwidth]{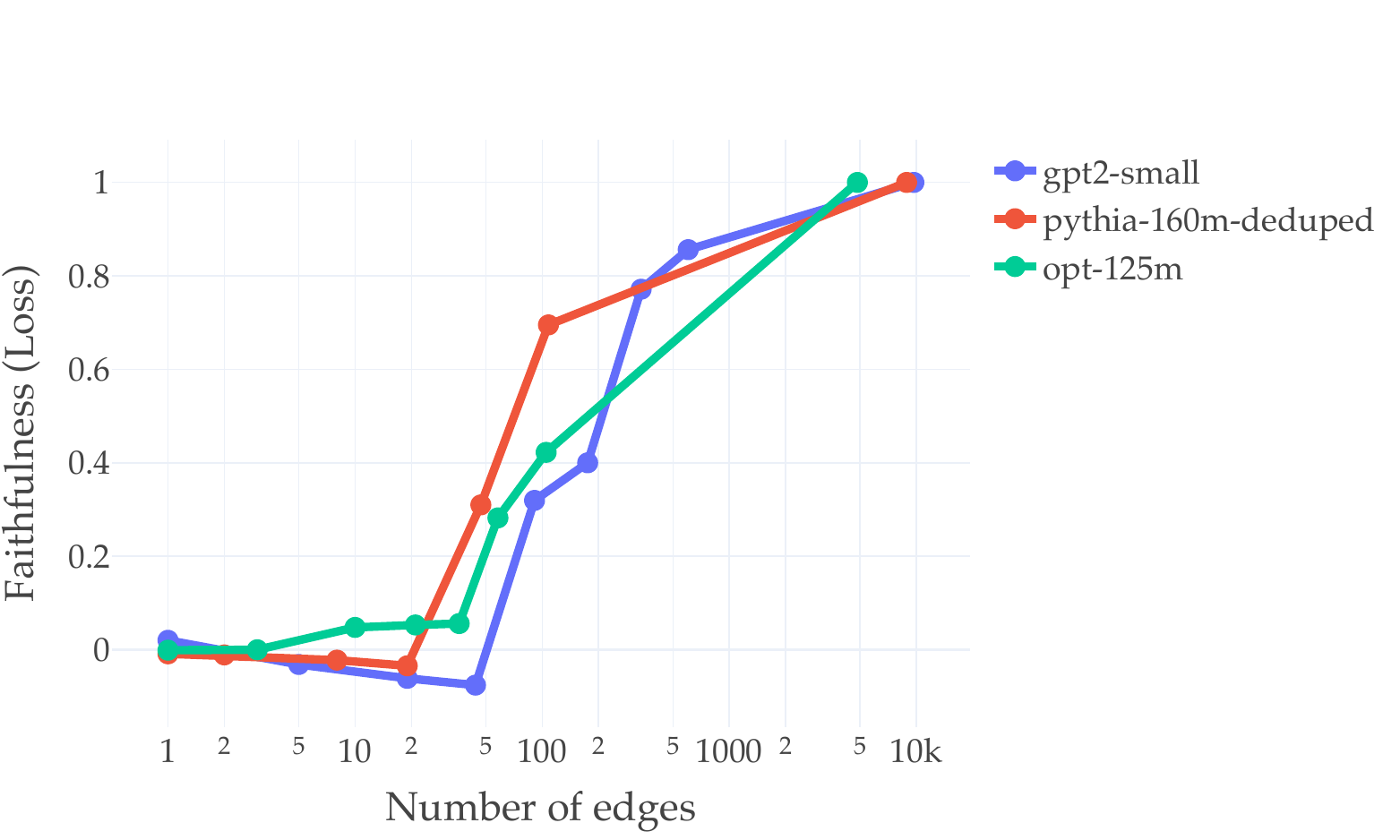}
        \caption{Edge faithfulness}
        \label{fig:faithfulness_edge_all_models_induction}
    \end{subfigure}
    \caption{Node-level faithfulness and edge-level faithfulness for three different pre-trained models, with loss on the repeated tokens in the induction examples as the metric. All models can be loaded in the TransformerLens library \citep{githubGitHubNeelnandaioTransformerLens}.}
\end{figure}

\section{Visualising the distribution of codes and their relation to the circuit}

\subsection{Correlation between occurrences/co-occurrences and presence in the circuit}

We provide two visualisations of the correlation between the number of unique positive codes and the presence of a head in the circuit. First, we show how the number of unique positive codes per head alongside the ground-truth circuit, arranged by layer and head, in Figure \ref{fig:comparison_unique_to_positive_vs_ground_truth}. Clearly, heads with a high number of positive layers are much more likely to be in the circuit. 

We show the high correlation between number of unique positive codes per head and its presence in the circuit in Figure \ref{fig:comparison_unique_to_positive_vs_ground_truth}. This corresponds to the node-level circuit identification, where (after softmax) we predict a head's presence in the circuit if it exceeds the pre-defined threshold $\theta$.

\begin{figure}[htbp]
    \centering
    \begin{subfigure}[t]{\textwidth}
        \includegraphics[width=\textwidth]{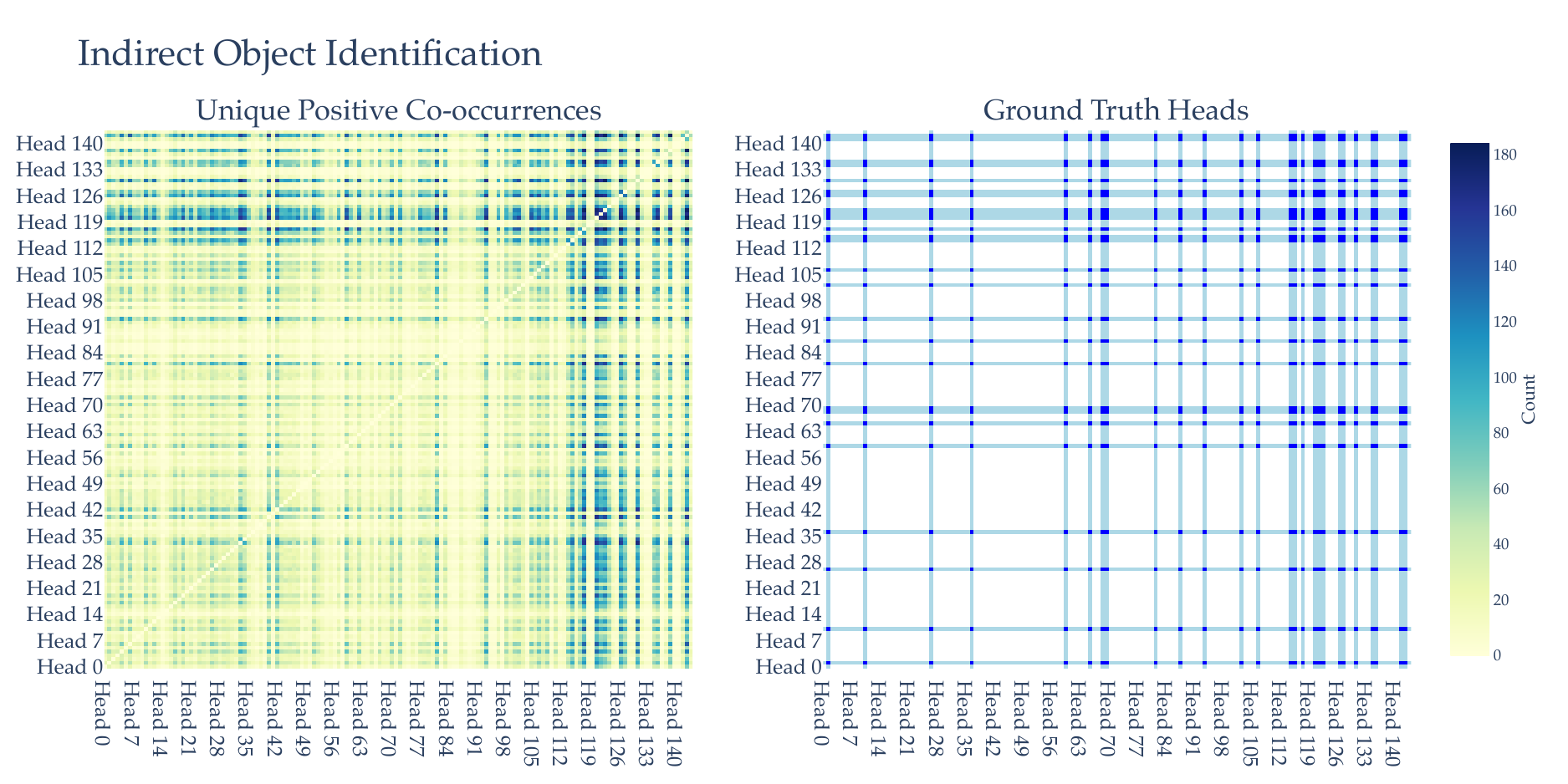}
        \label{fig:ioi_unique_to_positive_vs_ground_truth_edges}
    \end{subfigure}
    \\
    \begin{subfigure}[t]{\textwidth}
        \includegraphics[width=\textwidth]{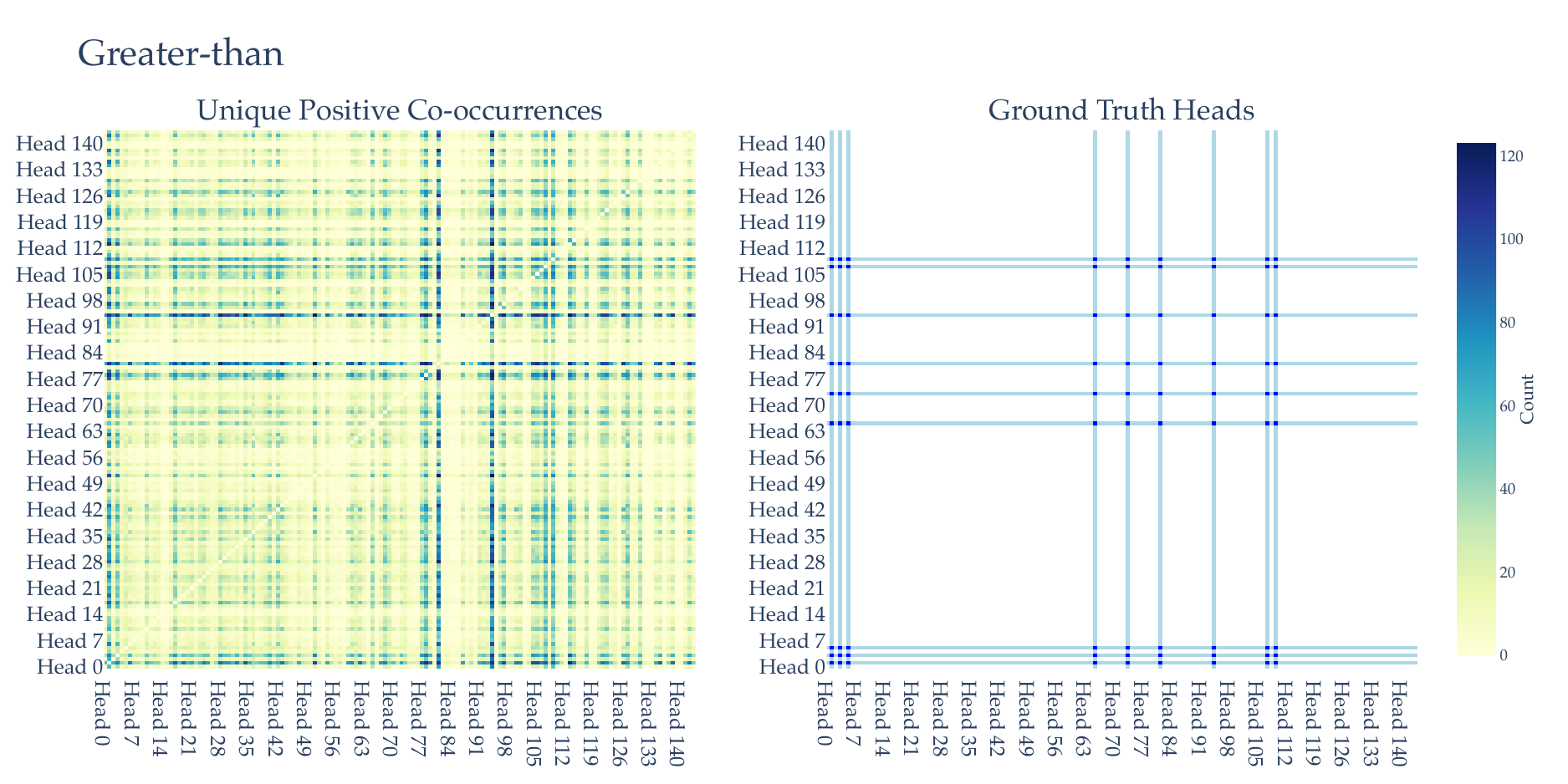}
        \label{fig:gt_unique_to_positive_vs_ground_truth_edges}
    \end{subfigure}
    \caption{Number of unique co-occurrences of codes by head in positive examples (left) and the ground-truth circuit (right). The entry at head $i$, head $j$ in the ground-truth matrix is dark blue if both heads are in the ground-truth circuit, light-blue if only one is, and white if neither are. For both IOI and Greater-than circuits, there are extremely similar patterns between both matrices.}
    \label{fig:combined_figures_cooc}
\end{figure}

Similarly, we show matrices of the co-occurrence of codes in Figure \ref{fig:combined_figures_cooc}, alongside the ground-truth circuit. However, because we have an array of size $n_\text{heads} \times n_\text{heads}$, this time we colour the ground-truth circuit as follows: a (head, head) array entry is dark blue if both heads are in the circuit, light blue if only one is, and white if neither are. Again, we see a very strong similarity between the number of co-occurrences of codes in (head, head) pairs and the presence of one or both heads in the circuit.

\subsection{Distribution and sparsity of codes}
\begin{figure}
    \centering
    \includegraphics[width=\textwidth]{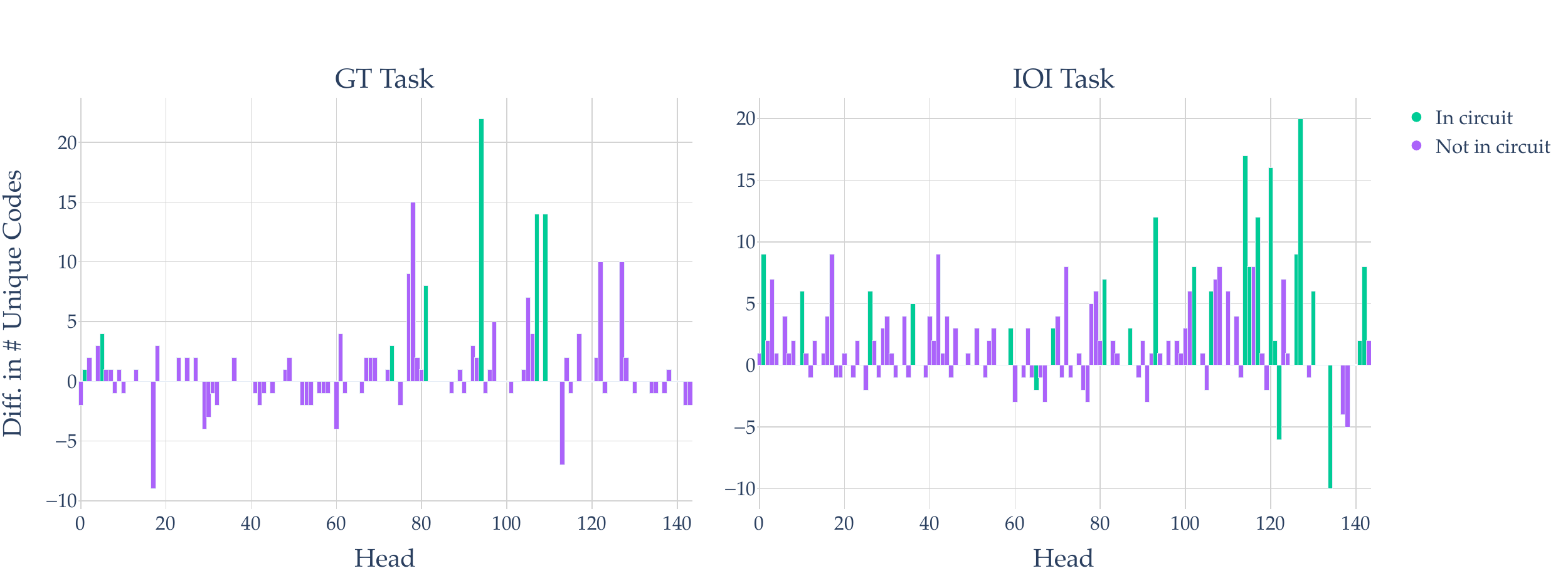}
    \caption{Plotting the difference between the number of unique codes per head in positive examples and negative examples. We then colour the bar by whether the head is in the ground-truth circuit or not. Interestingly, heads with a much larger number of unique codes in positive examples (as opposed to negative examples) are much more likely to be in the circuit.}
    \label{fig:unique_features_per_head_diff_combined}
\end{figure}

We also visualise the \textit{difference} between the number of unique positive codes and unique negative codes per head (Figure \ref{fig:unique_features_per_head_diff_combined}). Again, we see a strong pattern where heads with a high difference (many more unique positive codes than unique negative) are very likely to be in the ground-truth circuit.

\begin{figure}
    \centering
    \includegraphics[width=0.75\textwidth]{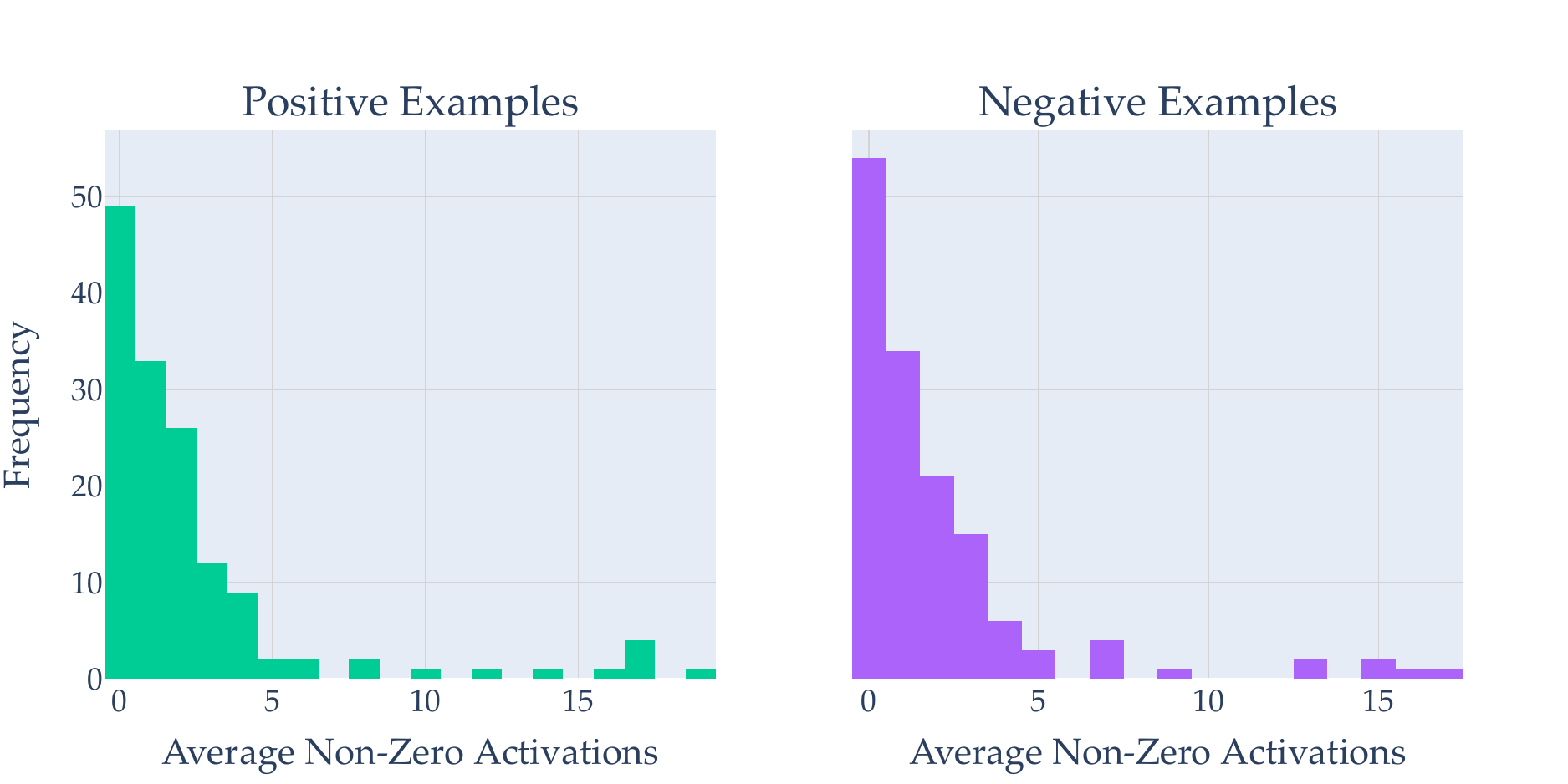}
    \caption{The distribution of non-zero activations by head across positive and negative examples. Positive examples tend to elicit slightly more non-zero activations.}
    \label{fig:average_non_zero_activations_pos_neg}
\end{figure}

We also examine the sparsity of our learned representations, and whether there is any different between positive and negative representations. We plot the histogram of average non-zero activations across all heads for the positive and negative examples in Figure \ref{fig:average_non_zero_activations_pos_neg}. Whilst there doesn't appear to be any significant difference, the average non-zero activations were slightly higher per positive example (0.56) compared to negative examples (0.42).

\begin{figure}
\centering
\includegraphics[width=0.9\textwidth]{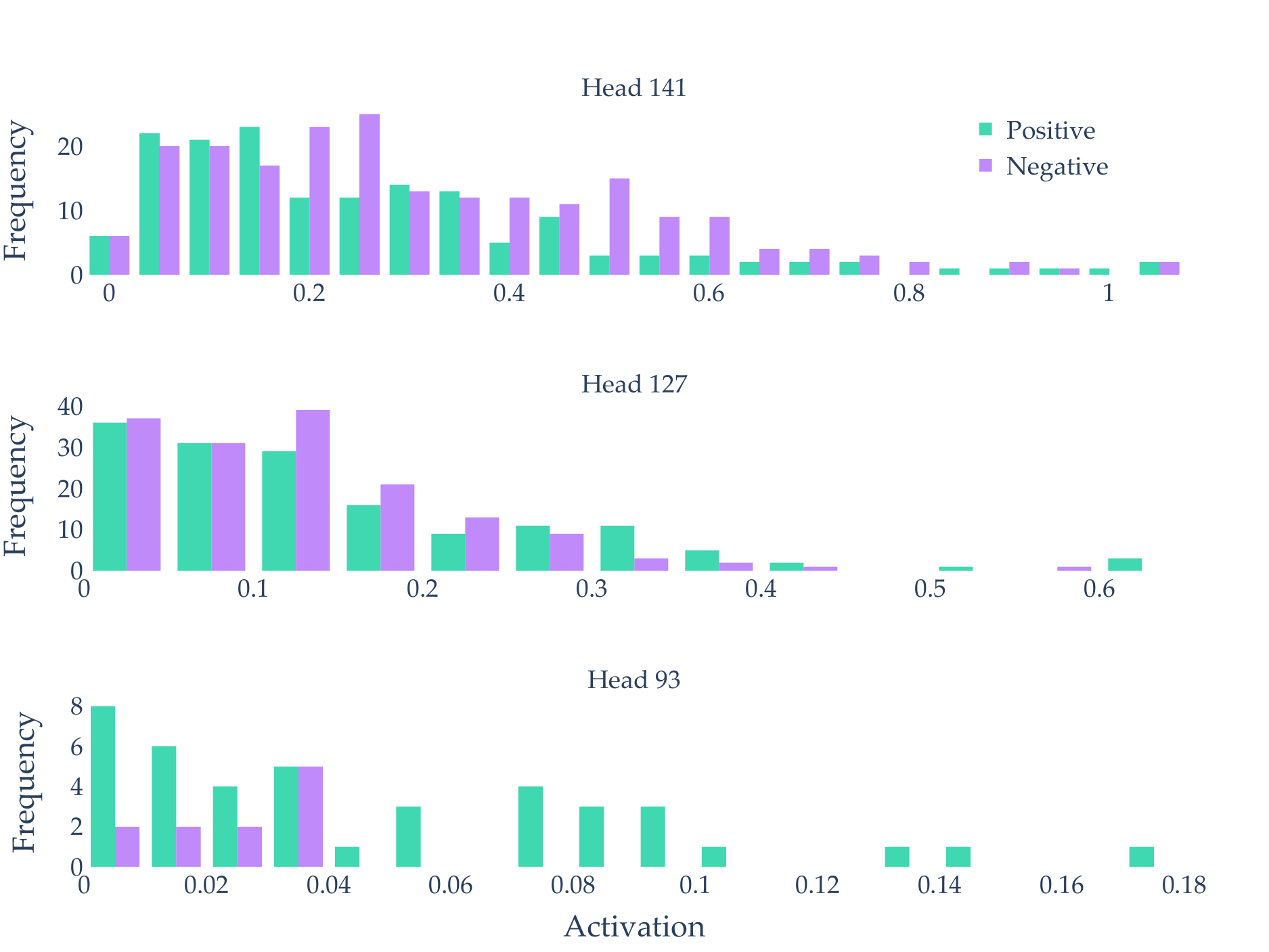}
\caption{Activation histograms for the most common positive code in each of the three ground truth heads (141, 127, and 93) of the IOI task. The activations for the positive examples are consistently higher than those for the negative examples, suggesting the relevance of these codes to the task-specific behavior.}
\label{fig:activation_histograms_most_common_code}
\end{figure}

Finally, we investigate the relationship between the most common positive codes and their activations in the ground truth heads. Figure \ref{fig:activation_histograms_most_common_code} presents activation histograms for the most frequently occurring positive code in each of the three identified heads (141, 127, and 93) for the IOI task. By comparing the activation distributions of these codes for positive and negative examples, we observe a consistent pattern of separated activations between positive and negative examples. The relatively clear separation between the positive and negative activation distributions further highlights the importance of these codes in the functioning of the identified circuit. This analysis provides valuable insights into the fine-grained workings of the learned representations and their role in capturing task-specific patterns.

\section{What didn't work: contrastive loss experiments}

In addition to the completely unsupervised setting discussed above, we hypothesised that injecting some information about the task into the autoencoder might assist it in constructing representations of residual streams that better allowed us to distinguish between positive and negative examples. As such, we trained the sparse autoencoder with an additional loss component that penalised the cosine similarity between positive and negative vectors at the same sequence position, whilst penalising dissimilarity between vectors with the same label at the same sequence position.

In the abstract setting, suppose we wish to compute the average cosine similarity between vectors at the same sequence position \(i\) in both tensors \(A\) and \(B\), while maintaining the sequence length \(S\) as the same for both. Given \(A \in \mathbb{R}^{D \times S \times N}\) and \(B \in \mathbb{R}^{D\times S \times M}\), we calculate the average cosine similarity for vectors at each sequence position \(i\) across all vectors at the same position in the other tensor. Specifically, for every vector at sequence position \(i\) in \(A\), compute its average cosine similarity with all vectors at position \(i\) in \(B\), and vice versa.

The process is as follows:
\begin{enumerate}
\item \textit{Cosine Similarity Calculation for Position \(i\)}: For each position \(i\) in the sequence \(S\), and for each dimension \(d\) \((d \in \{1, \dots, D\})\), calculate the cosine similarity between vector \(a_{dis} \in A\) and every vector \(b_{dis} \in B\), where \(s\) is fixed for this operation indicating the sequence position. Specifically, for vectors at sequence position \(i\), the similarity between \(a_{dis}\) and all \(b_{dis} \in B\) is calculated as:
\[
\text{sim}_{dis} = \frac{1}{M} \sum_{m=1}^{M} \frac{\sum_{d=1}^{D} a_{dis} \cdot b_{dim}}{\sqrt{\sum_{d=1}^{D} a_{dis}^2} \cdot \sqrt{\sum_{d=1}^{D} b_{dim}^2}}
\]

\item \textit{Average Across All Sequence Positions}: After computing \(\text{sim}_{dis}\) for every vector at sequence position \(i\) in \(A\) against all vectors at position \(i\) in \(B\), average these similarities across all sequence positions \(S\):
\[
\text{CS}(A, B) = \frac{1}{S} \sum_{s=1}^{S} \left( \frac{1}{N} \sum_{n=1}^{N} \text{sim}_{dsn} + \frac{1}{M} \sum_{m=1}^{M} \text{sim}_{dsm} \right)
\]

\item \textit{Final loss}: Finally, we take our tensor of positive examples $P \in \mathbb{R}^{D\times S \times N}$, where $D$ is the number of learned features in the sparse autoencoder, $S$ is our sequence length (for the individual attention heads data $S=144$, 12 heads in 12 layers) and $N$ is the number of positive examples in the batch. Similarly, we take our tensor of negative examples $N\in \mathbb{R}^{D\times S \times M}$, where $M$ is the number of negative examples in the batch. We then calculate the additional loss component using a typical contrastive loss structure:
\[
\mathcal{L}_\text{cont}(\mathbf{P}, \mathbf{N}) = CS(\mathbf{P},\mathbf{N}) + \text{max} (0, \epsilon - CS(\mathbf{P},\mathbf{P})/2 - CS(\mathbf{N},\mathbf{N})/2 )
\]
This is simply added to the overall loss, in addition to the reconstruction loss and sparsity loss (note the concatenation of $\mathbf{P}$ and $\mathbf{N}$ along the batch dimension yields $\mathbf{x}$):
$$
\mathcal{L}(\mathbf{x})=\underbrace{\|\mathbf{x}-\hat{\mathbf{x}}\|_2^2}_{\text {Reconstruction loss }}
+
\underbrace{\lambda\|\mathbf{c}\|_1}_{\text {Sparsity loss }} 
+
\underbrace{\alpha \mathcal{L}_\text{cont}(\mathbf{P}, \mathbf{N})}_{\text {Contrastive loss }}
$$

\end{enumerate}

\begin{figure}
    \centering
    \includegraphics[width=0.8\textwidth]{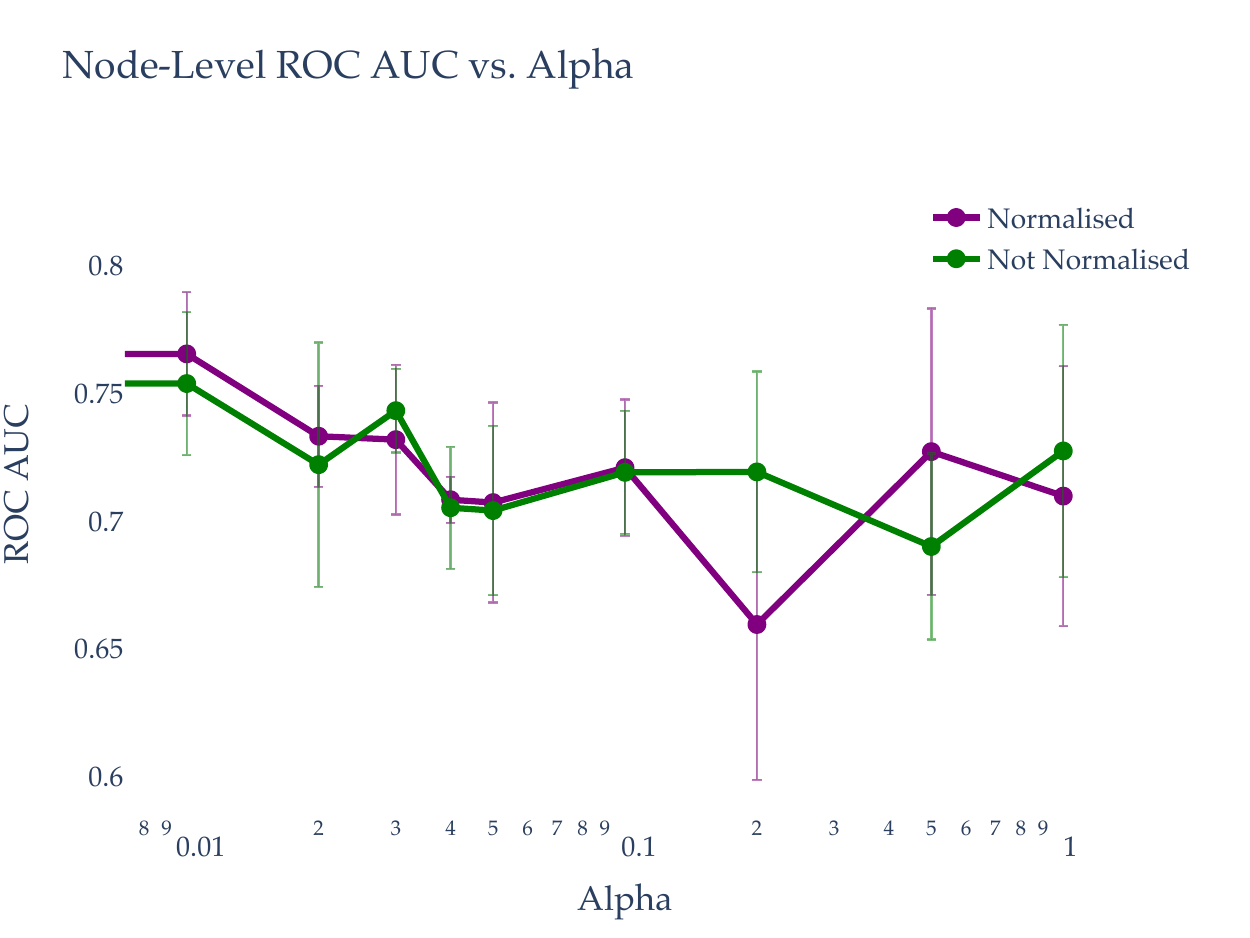}
    \caption{Node-level ROC AUC compared to the contrastive loss strength, measured by the hyperparameter $\alpha$. Contrastive loss seems to hinder performance.}
    \label{fig:roc_auc_vs_alpha}
\end{figure}

We do 10 runs at varying $\alpha$ levels for the IOI task and show the results in Figure \ref{fig:roc_auc_vs_alpha}. Unfortunately, contrastive loss seemed to decrease the quality of the final predictions. This holds both with and without normalisation.

\section{Entropy vs. Co-occurrence for Edge-Level Circuit Identification}

In addition to the edge-level circuit identification methodology described in the main text, we explored an alternative approach based on the entropy of positive code co-occurrences in head pairs. This method again involves constructing a co-occurrence matrix $\mathbf{C} \in \mathbb{R}^{n_\text{heads} \times n_\text{heads} \times d_\text{bottleneck} \times d_\text{bottleneck}}$, where each entry $\mathbf{C}_{h1,h2,f1,f2}$ represents the frequency of co-occurrence of feature $f1$ in head $h1$ with feature $f2$ in head $h2$ across all examples. The matrix is populated by analysing the model's activations for each input example, identifying the ``active'' feature (argmax across the feature dimension) for each head in every example, and incrementally building the co-occurrence counts for each observed pair of active features across all head pairs.

To distill meaningful relationships from the co-occurrence matrix, we calculate the entropy $H_{h1,h2}$ for each head pair $(h1, h2)$ as follows:
\begin{align*}
H_{h1,h2} = -\sum_{f1=1}^{d_\text{bottleneck}} \sum_{f2=1}^{d_\text{bottleneck}} p_{h1,h2,f1,f2} \log_2(p_{h1,h2,f1,f2})
\end{align*}
where $p_{h1,h2,f1,f2}$ represents the normalised probability of co-occurrence of features $f1$ and $f2$ between heads $h1$ and $h2$, derived from $\mathbf{C}$. Prior to entropy calculation, $\mathbf{C}$ is normalised such that for each head pair, the sum of all co-occurrence probabilities equals one.

We apply a softmax function row-wise across the entropy matrix to normalise the entropy values and calculate the entropy matrices for the positive example set, denoted as $\mathbf{H}^+$. From $\mathbf{H}^+$, we select the top $k$ head pairs based on their normalised entropy difference, indicating the most significant edges in the circuit. These head pairs are then mapped back to their corresponding layers and heads within the model architecture.

\begin{figure}
    \centering
    \includegraphics[width=0.8\textwidth]{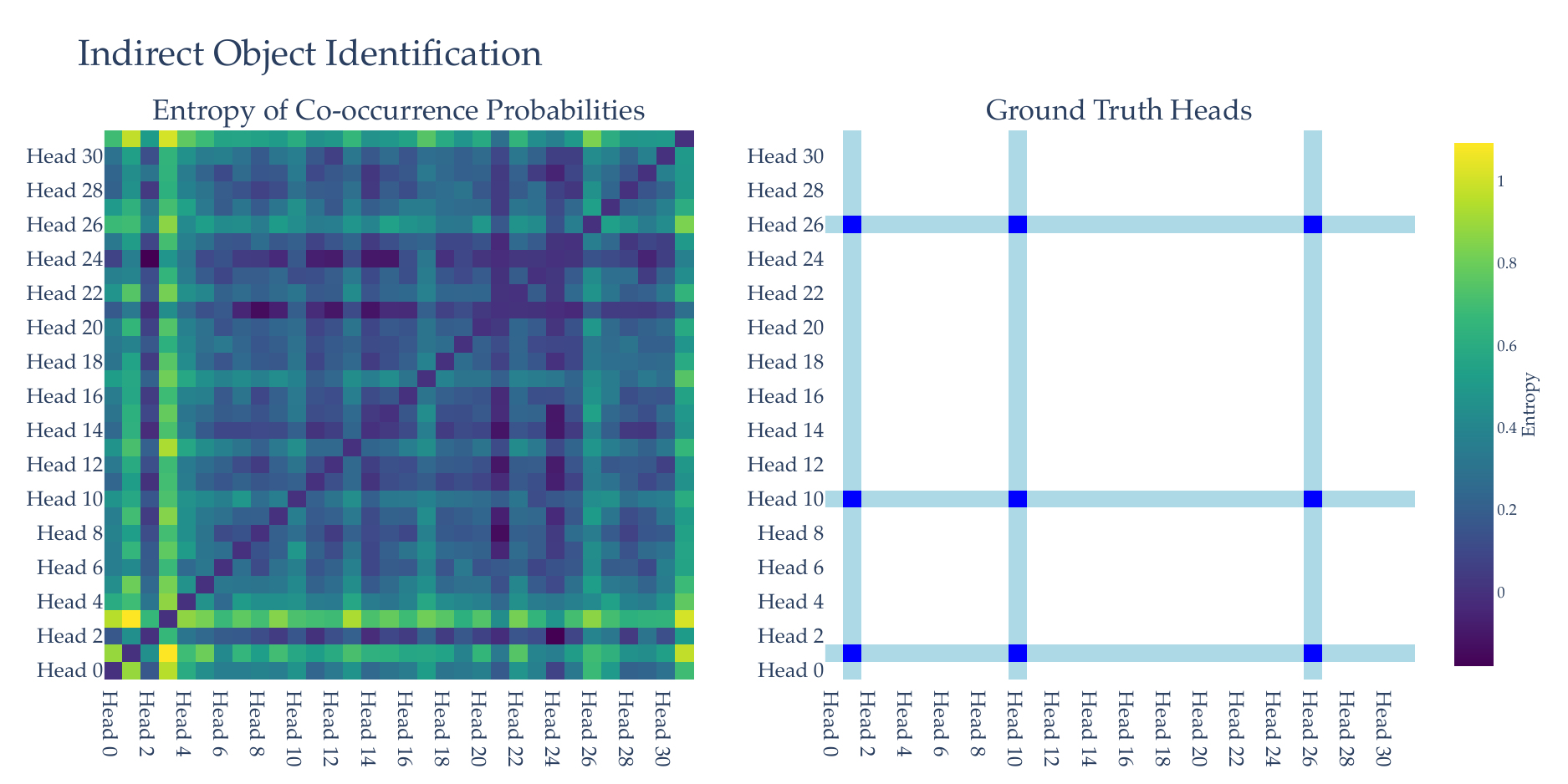}
    \caption{The entropy of co-occurrence probabilities mapped against the ground-truth heads in the circuit. Compare to just examining the co-occurrences as shown in Figure \ref{fig:combined_figures_cooc}.}
    \label{fig:ioi_entropy_vs_ground_truth_edges}
\end{figure}

\begin{figure}
    \centering
    \includegraphics[width=0.8\textwidth]{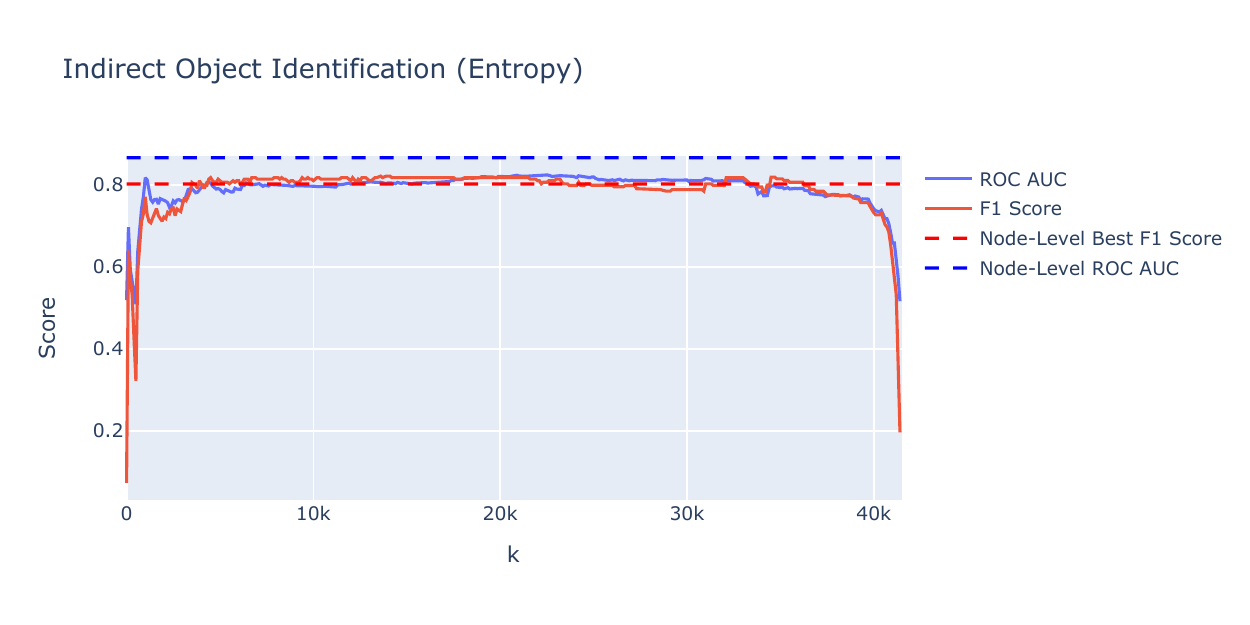}
    \caption{Use of entropy instead of softmaxed co-occurrence to identify heads belonging to the circuit. Whilst performance was still reasonable, it did not reach the ROC AUC of the original method.}
    \label{fig:ioi_unique_co_occurrences_entropy}
\end{figure}

To evaluate the robustness and predictive power of the identified circuit components (edges), we generate a binary prediction array for the ground truth head pairs, applying a threshold to the softmax-normalised values from $\mathbf{H}^+$. However, we found entropy to be a worse signal for predicting the presence of a head in a circuit. As shown in Figure \ref{fig:ioi_entropy_vs_ground_truth_edges}, the entropy values do not align well with the ground truth edges in the IOI task. Furthermore, even when selecting the best $k$ hyperparameter, the ROC AUC does not exceed 0.50 (see Figure \ref{fig:ioi_unique_co_occurrences_entropy}), indicating that the entropy-based approach does not reach the performance level of the original co-occurrence-based method described in the main text. 

\section{Full hyperparameter sweeps}

\subsection{Best hyperparameters for each task}

The best hyperparameters for each task are given in Table \ref{tab:hyperparams}. Importantly, all optimal hyperparameters are approximately the same between the Greater-than and IOI tasks, with some differences to Docstring. We hypothesise that this makes sense because Docstring has a smaller number of heads in the model and thus may need higher regularisation i.e. a higher sparsity penalty $\lambda$.

\begin{table}[htbp]
\caption{Best hyperparameters found using an Optuna search for 100 iterations over number of learned features and $\lambda$ (the sparsity penalty). The number of learned features was chosen between 128 and 2048. The value for lambda was chosen between 0.01 and 0.1. Additionally, we show the threshold (after softmax) that maximises the F1 score for each dataset.}
\centering
\begin{tabular}{@{}lccc@{}}
\toprule
\textit{Task} & Learned Features & \( \lambda \) & Threshold \\
\midrule
Docstring & 270 & 0.067 & 1.12e-07 \\
Greater-than & 246 & 0.011 & 1.71e-15 \\
IOI & 379 & 0.022 & 6.26e-16 \\
\bottomrule
\end{tabular}
\label{tab:hyperparams}
\end{table}

\subsection{Effect of $k$ in edge-selection}
\label{app:k_effect}

\begin{figure}[htbp]
    \centering
    \begin{minipage}[b]{0.32\textwidth}
        \centering
        \includegraphics[width=\textwidth]{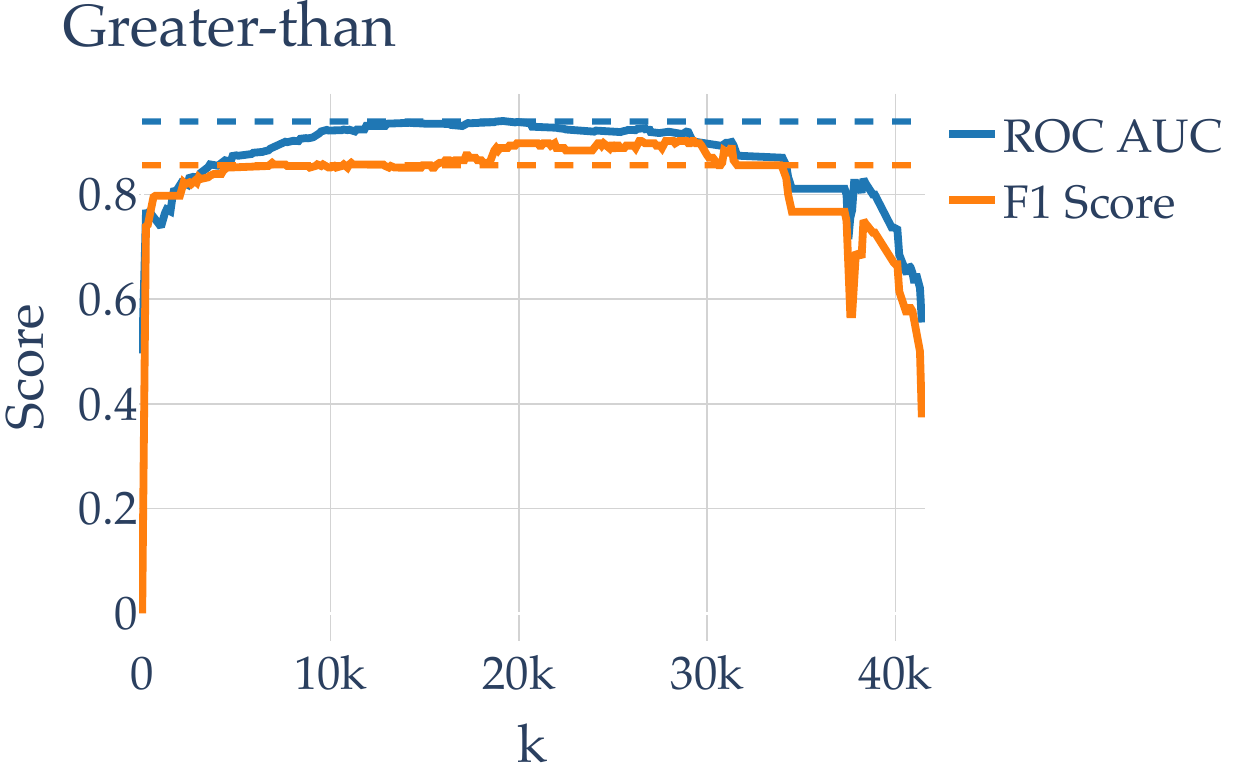}
    \end{minipage}
    \hfill
    \begin{minipage}[b]{0.32\textwidth}
        \centering
        \includegraphics[width=\textwidth]{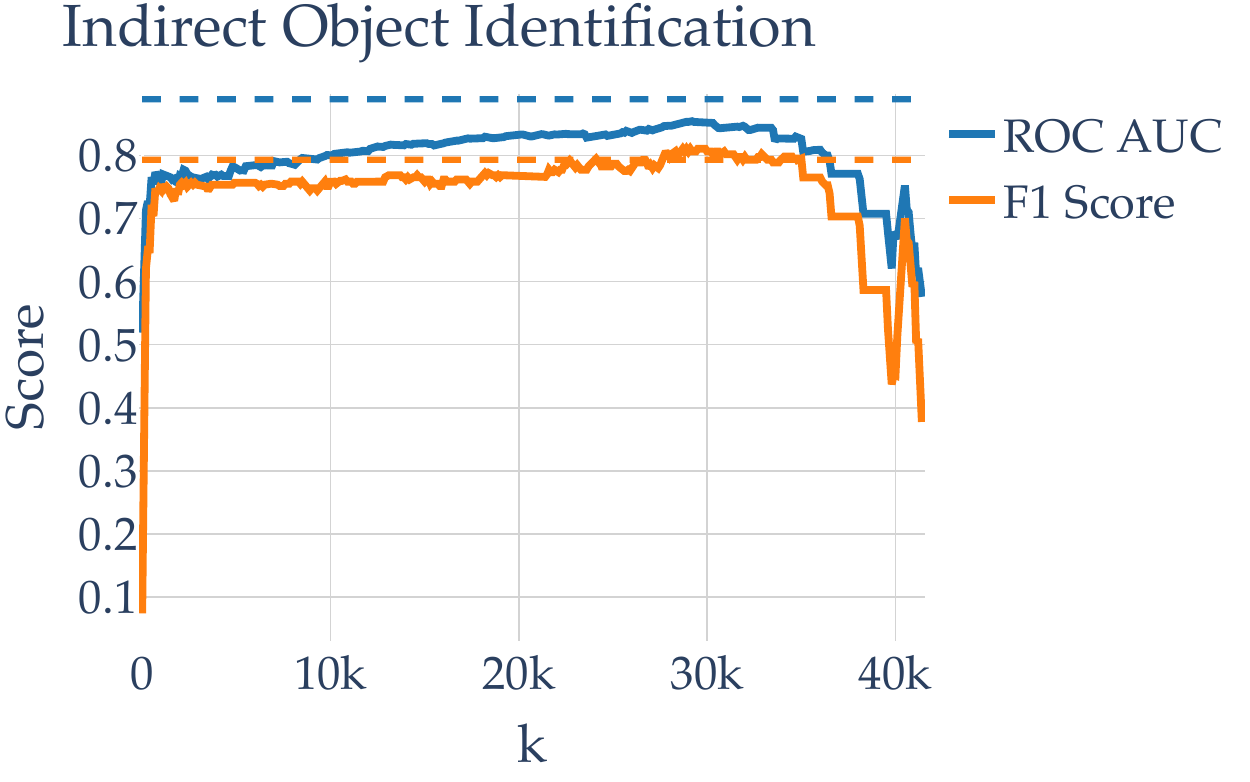}
    \end{minipage}
    \hfill
    \begin{minipage}[b]{0.32\textwidth}
        \centering
        \includegraphics[width=\textwidth]{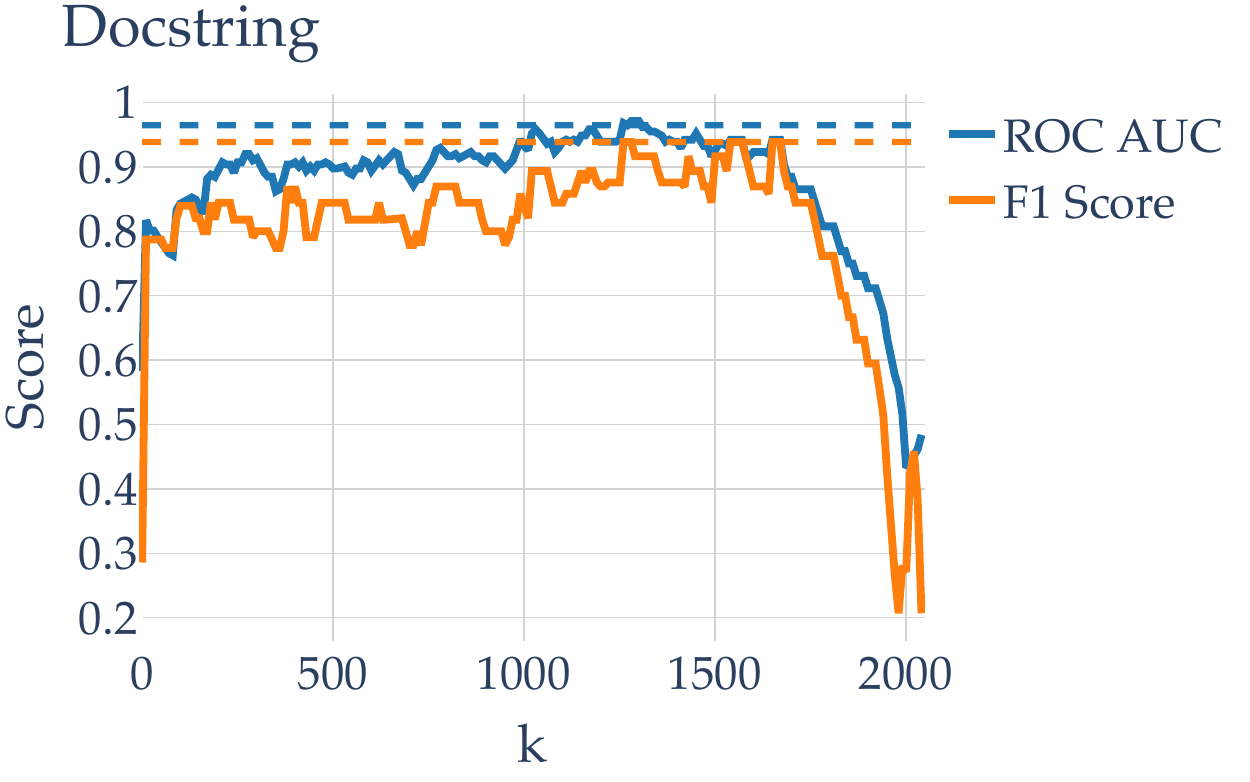}
    \end{minipage}
    
    \caption{Comparison of keeping the set of top-$k$ co-occurrences in edge-level circuit identification for the three tasks. Choosing $k$ to be approximately half of the total number of co-occurrences in codes seems to be a good heuristic to maximise the ROC AUC \textit{a priori}.}
    \label{fig:k_effect_comparison}
\end{figure}

Another hyperparameter introduced by edge-level detection is the number of (head, head) co-occurrence pairs to add to keep from the sorted list of positive co-occurrence counts before taking the set and only keeping heads existing in the remaining tuples in that set. Figure \ref{fig:k_effect_comparison} illustrates that ROC AUC is relatively robust to the selection of $k$; most of the values of $k$ lead to essentially the same performance, except for at the very start and very end. In fact, it seems that selecting $k$ to be half of the number of overall co-occurrence pairs of codes in heads is a robust heuristic for optimising performance across datasets.

\subsection{Contour plots}

\begin{figure}[htbp]
    \centering
    \begin{subfigure}[b]{0.49\textwidth} 
        \centering
        \includegraphics[width=\textwidth]{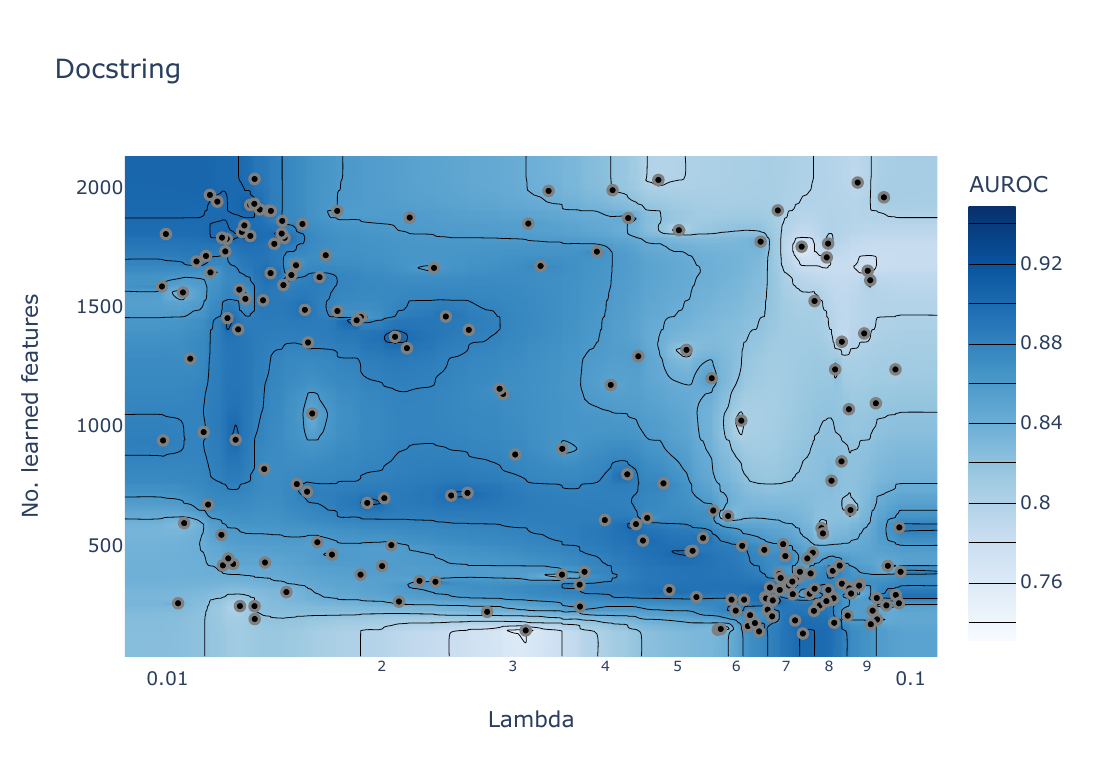}
        \label{fig:ds_contour_plot}
    \end{subfigure}
    \hfill 
    \begin{subfigure}[b]{0.49\textwidth} 
        \centering
        \includegraphics[width=\textwidth]{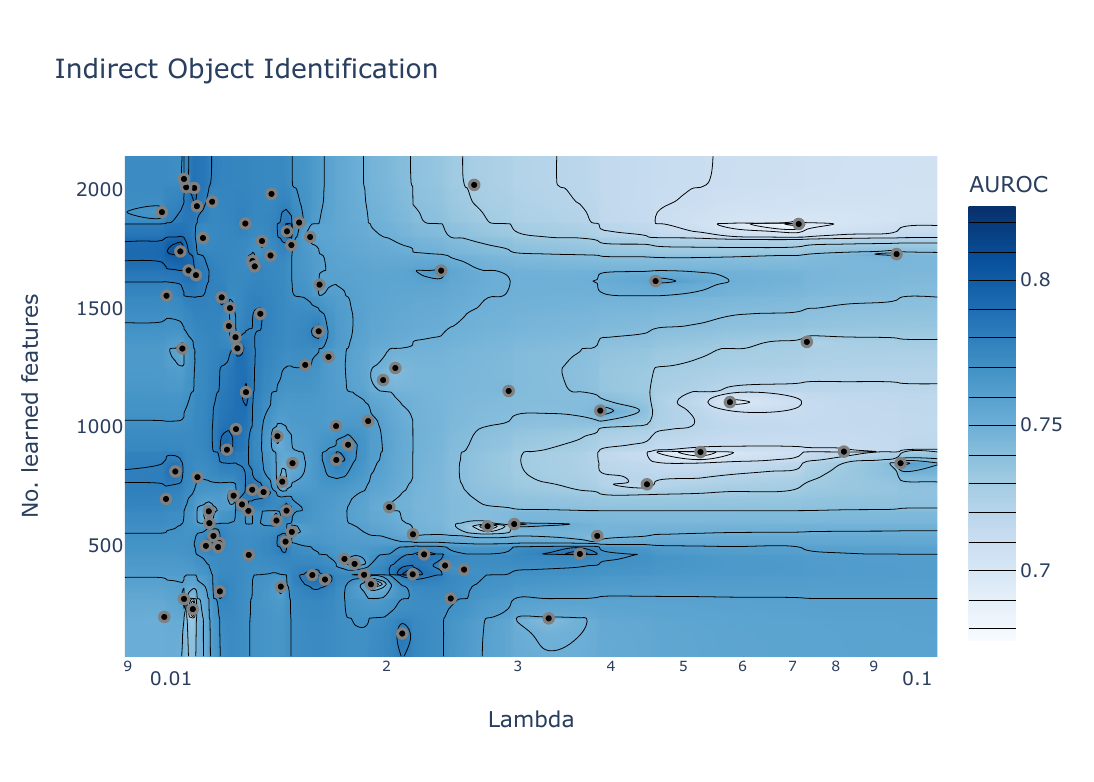}
        \label{fig:ioi_contour_plot}
    \end{subfigure}
    
    \begin{subfigure}[b]{0.49\textwidth} 
        \centering
        \includegraphics[width=\textwidth]{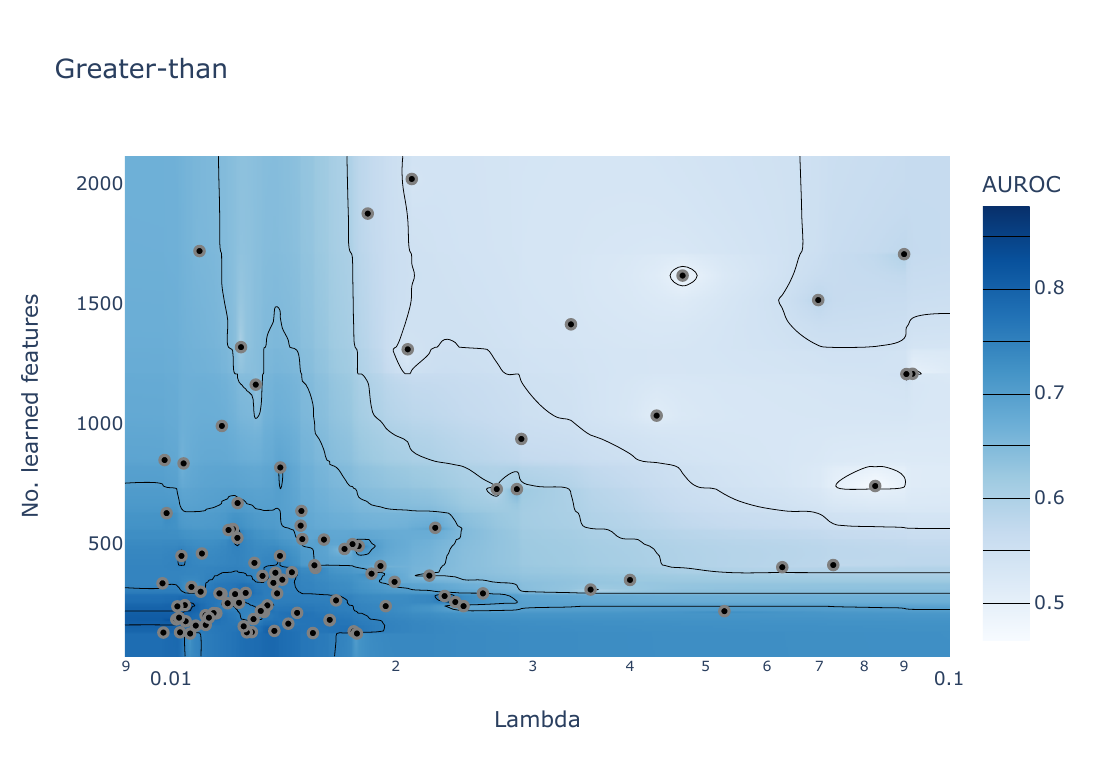}
        \label{fig:gt_contour_plot}
    \end{subfigure}
    
    \caption{Optuna hyperparameter searches over 100 autoencoder training runs for the Greater-than and IOI tasks and 200 runs for the Docstring task, optimising the node-level ROC AUC of the predicted circuit using our method.}
    \label{fig:combined_plots}
\end{figure}

We show contour plots from the Optuna optimisation in Figure \ref{fig:combined_plots}. It seems that a lower $\lambda$ across datasets is beneficial. However, the number of learned features does not seem to be that important for Docstring or IOI, whilst a low number of learned features leads to lower AUC for the Greater-than task.

\section{Comparison to Vector-Quantised Variational Autoencoders (VQ-VAEs)}
\label{app:vqvae}

Our work on using sparse autoencoders for circuit discovery shares some key similarities with Vector-Quantized Variational Autoencoders (VQ-VAEs) \citep{van2017neural}. Both methods aim to learn discrete representations of input data that capture the most salient information while discarding noise and irrelevant details.

In a VQ-VAE, the encoder network maps the input data to a continuous latent space, which is then quantised using a learned codebook of discrete vectors. The quantisation step involves finding the nearest codebook vector to each latent vector and replacing it with the corresponding discrete code. This process is analogous to our approach of taking the argmax of the learned features to obtain discrete codes representing the most activated feature for each attention head.

However, there are some important differences between our method and VQ-VAEs. First, our approach uses a sparse autoencoder with a sparsity penalty in the loss function to encourage the learning of a sparse representation. This sparsity constraint helps to identify the most important features and reduces the influence of noise and irrelevant information. In contrast, VQ-VAEs essentially enforce the maximum amount of sparsity by default, as they assign a single vector to each bottleneck representation, which is bijective to a set of integer codes. This is equivalent to having a sparse autoencoder with only one activating feature at any time, and this feature can only take on a single value (i.e. 1).

Second, in a VQ-VAE, the codebook vectors are learned jointly with the encoder and decoder networks using a vector quantisation objective. The codebook is updated during training to better represent the latent space. In our approach, we do not explicitly learn a codebook; instead, we rely on the sparsity constraint to encourage the autoencoder to learn a set of meaningful features that can be discretised using the argmax operation. However, the hidden dimension of our autoencoder is analogous to the codebook size when we apply our discretisation.

Third, VQ-VAEs use a straight-through gradient estimator to propagate gradients through the quantisation step, allowing for end-to-end training of the encoder, codebook, and decoder \citep{van2017neural}. In our method, we train the sparse autoencoder using standard backpropagation without the need for a specialised gradient estimator, because the discretisation comes after.

Despite these differences, both our method and VQ-VAEs share the goal of learning a meaningful discrete representation of the input data. It would be interesting to substitue VQ-VAEs into our method and determine if a similar or better performance can be achieved.


\newpage

\end{document}